%% file: main.tex
\documentclass{article}

\usepackage[main, final]{neurips_2026}

\usepackage[utf8]{inputenc} %
\usepackage[T1]{fontenc}    %
\usepackage{amsfonts}       %
\usepackage{amsmath}        %
\usepackage{amssymb}        %
\usepackage{url}            %
\usepackage{booktabs}       %
\usepackage{nicefrac}       %
\usepackage{microtype}      %
\usepackage{xcolor}         %

\usepackage{subcaption}
\usepackage{calc}
\usepackage{tabularx}
\usepackage{float}
\usepackage{multirow}
\usepackage{multicol}
\usepackage{comment}
\usepackage{makecell}
\usepackage{siunitx}
\usepackage{tikz}
\usepackage{pgfplots}
\pgfplotsset{compat=1.18}
\usepackage{xspace}

\usepackage{algorithm}
\usepackage{algpseudocodex}

\usepackage[bottom]{footmisc}

\definecolor{linkblue}{rgb}{0.1, 0.45, 0.8}
\usepackage[colorlinks=true, citecolor=linkblue, linkcolor=linkblue, urlcolor=linkblue]{hyperref}       %
\usepackage[capitalize]{cleveref}       %

\usepackage{etoc}  %

\input{custom_commands}

\title{On the Robustness of Watermarking for Autoregressive Image Generation}

\author{%
  Andreas~Müller \\
  Ruhr University Bochum\\
  \texttt{andreas.mueller-t1x@rub.de} \\
  \And
  Denis Lukovnikov \\
  Ruhr University Bochum \\
  \texttt{denis.lukovnikov@rub.de} \\
  \And
  Shingo Kodama\\
  Middlebury College \\
  \texttt{skodama@middlebury.edu} \\
  \And
  Minh Pham\\
  New York University\\
  \texttt{mp5847@nyu.edu} \\
  \And
  Anubhav Jain\\
  Independent Researcher\\
  \texttt{aj3281@nyu.edu} \\
  \And
  Jonathan Petit\\
  Qualcomm\\
  petit@qti.qualcomm.com\\
  \And
  Niv Cohen\\
  Technion -\\Israel Institute of Technology\\
  \texttt{nivc@technion.ac.il} \\
  \\
  \And
  Asja Fischer\\
  Ruhr University Bochum \\
  \texttt{asja.fischer@rub.de} \\
}

\usepackage{titlesec}

\begin{document}

\maketitle

\begin{abstract}
The proliferation of realistic autoregressive (AR) image generators necessitates robust methods for mitigating misinformation and filtering synthetic image content.
To this end, watermarking schemes for image models embed a subtle signal at generation time, enabling downstream verification with a corresponding detector.
In this work, we study the vulnerability of such watermarks to removal and forgery attacks.
We evaluate existing attacks and introduce three new ones: (i) a vector-quantized regeneration attack for removal, (ii) an adversarial optimization-based attack, and (iii) a frequency injection attack.
Our evaluation reveals that, depending on the scheme, either removal or forgery can be effective in black-box settings.
Consequently, we find that current AR watermarking schemes may not reliably support content filtering.
Specifically, we show that BitMark can be susceptible to \emph{Watermark Mimicry}, a form of forgery in which authentic images are modified to imitate a generator's watermark to avoid being harvested for model training.\\
Code: \href{https://github.com/and-mill/ar-watermarking-robustness}{github.com/and-mill/ar-watermarking-robustness}
\end{abstract}

\section{Introduction}
\label{sec:introduction}
Autoregressive (AR) image generation has emerged as a powerful paradigm for visual synthesis, with recent models such as VAR~\citep{tian2024nspvar}, HMAR~\citep{kumbong2025hmar},  Infinity~\citep{han2025infinity}, and many closed-source models demonstrating remarkable capabilities. Unlike diffusion-based approaches, these models generate images by predicting discrete visual tokens in an autoregressive manner, enabling efficient sampling and fine-grained control over the generation process.
With the growing adoption of autoregressive image generators, provenance and misuse have become pressing concerns, leading to several in-generation watermarking methods tailored to these models~\citep{jovanovic2025wmar, kerner2025BitMark, lukovnikov2025clustermark, tong2025indexmark}.
These watermarking techniques embed a signal during the generation process, making them inherently more resistant to post-hoc attacks compared to post-processing watermarking schemes~\citep{Wen2023TreeRing,ZhaZhaSu2024invisibleimagewatermarksprovably}. Typically, they embed a signal by slightly biasing %
the sampling probabilities of specific %
tokens selected during generation. The resulting watermarks exhibit high robustness against common transformations, as well as regeneration attacks using diffusion models~\citep{ZhaZhaSu2024invisibleimagewatermarksprovably}.

\begin{figure}[t]
    \centering
    \includegraphics[width=0.75\columnwidth]{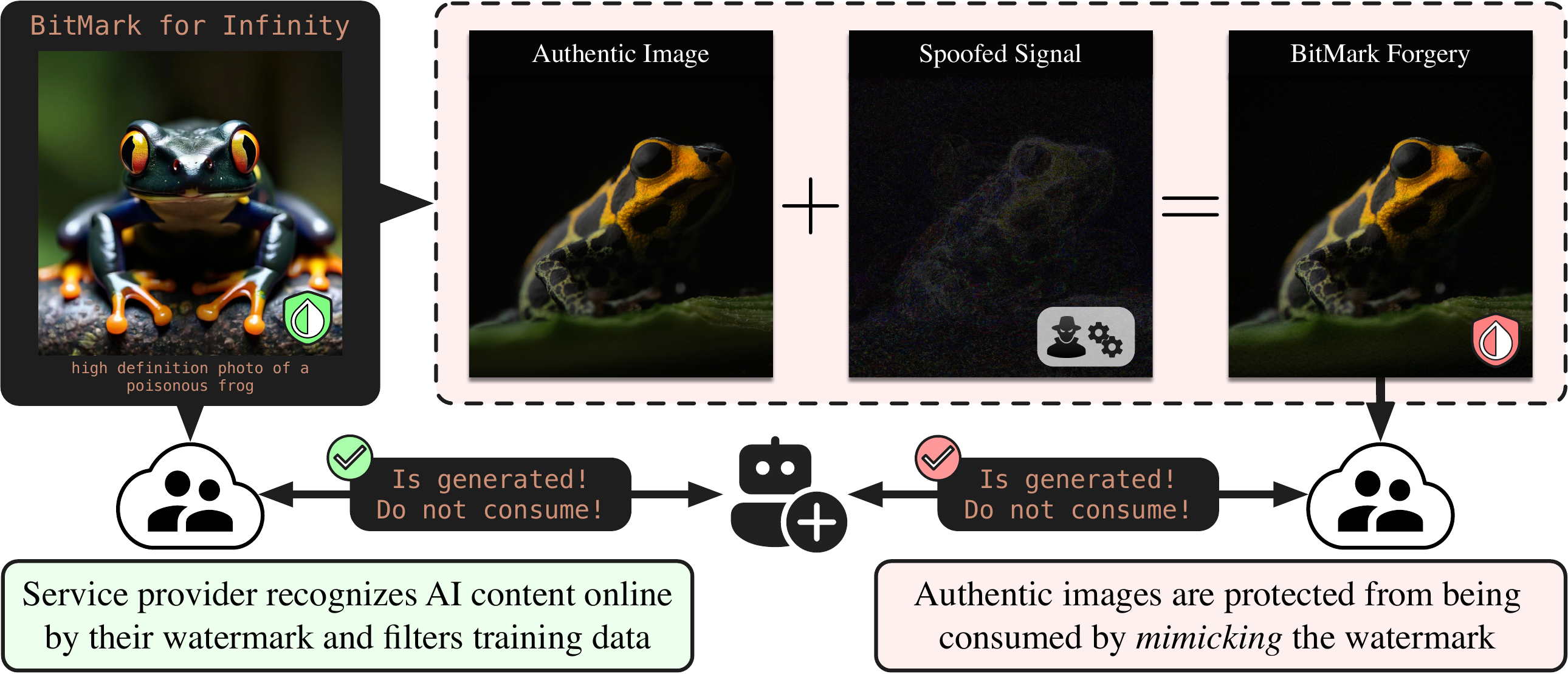}
    \caption{\textbf{Watermark Mimicry subverts synthetic content filtering}. An image generated with the radioactive BitMark watermark (left) carries an embedded signal that allows service providers to identify synthetic content and exclude it from future training to prevent model collapse. However, other parties can transfer this watermark onto unrelated images,
    causing the detector to mistake authentic images (here, \emph{Ranitomeya imitator - the mimic poison frog}) as AI-generated (right). This adversarial strategy inverts the protective intent of radioactive watermarking: rather than preventing real data from being mislabeled, one may \textit{exploit} the detector's decision boundary to shield genuine content from being harvested for training.
    {\tiny
    Photo by \href{https://www.inaturalist.org/people/andes2amazonexpeditions}{andes2amazonexpeditions}
    is licensed under \href{https://creativecommons.org/licenses/by-nc/4.0/}{CC BY-NC 4.0}.
    Sourced from \href{https://www.inaturalist.org/photos/500754765}{iNaturalist}.
    }}
    \label{fig:frontpage}
\end{figure}

However, targeted attacks specifically designed to exploit the structure of autoregressive image generators and their watermarking mechanisms remain largely %
unexplored.
This gap in understanding poses significant risks, as sophisticated adversaries could potentially circumvent watermark detection or falsely claim a non-watermarked image to be watermarked~\citep{Muller2025semanticforgery,jain2025forgingremoving}.
The latter point is of particular interest in the context of recent radioactive watermarking techniques, specifically \emph{BitMark} for bitwise autoregressive image generation, which are intended as a means of filtering synthetic content from training data to prevent model degradation~\citep{shumailov2024modelcollapse,alemohammad2024selfconsuming}: Triggering false detection via forgery attacks may enable third parties to perform \emph{Watermark Mimicry}, where authentic images can be protected from being harvested for model training (see~\cref{fig:frontpage}).
These risks and opportunities motivate our investigation of forgery attacks on watermarks for AR image generators.

\paragraph{Contributions.}
In this work, (i) we demonstrate that recent in-generation watermarks for autoregressive image generation can, in most cases, be successfully removed with limited impact on image quality, requiring neither access to the original generator model nor knowledge of the secret watermarking parameters. (ii) We propose three new watermark attacks. (iii) We show that while BitMark is robust against removal attempts, it cannot admit radioactive images while rejecting forgery attacks.

\section{Background}
\label{sec:background}

\subsection{Autoregressive Image Generation}

Recent works~\citep{han2026nextstep, wang2025tokenbridge,fan2025fluid, kumbong2025hmar, han2025infinity, tang2025hart, li2024continuous, wang2024emu3, yu2025rar, chang2023muse, chang2022maskgit,han2025infinity, tian2024nspvar} have proposed AR methods for generating images, using different sampling approaches.
When generating an image, recent AR methods first decode a low-resolution latent $z \in \R^{d \times h \times w}$, and subsequently map $z$ to a high-resolution pixel image $x \in \R^{3 \times H \times W}$ using a decoder \decoder.
In order to produce vectors in latent space, earlier AR image generators vector-quantize the latent space, building a vocabulary \vocabulary of visual tokens $q$, each corresponding to a $d$-dimensional vector in the latent space.
The mapping $\codebook: \vocabulary \rightarrow \mathbb{R}^d$ is called the codebook \codebook.

An AR image generator decomposes the joint distribution of visual tokens $p(q_1, q_2, \dots, q_{h\cdot w})$ as the product $\prod_i^{h\cdot w} \armodel(q_i | q_{<i}, c)$. %
Here, the distribution $p_{\theta}$ over the vocabulary \vocabulary is obtained from the logits computed by a neural network $f_{\theta}$ using a softmax layer: $\armodel (\cdot) = \softmax \circ f_{\theta}(\cdot)$.
Generation then proceeds by sampling a token from this categorical distribution, conditioned on previously produced tokens~\citep{yu2025rar,sun2024llamagen,chameleonteam2025chameleon}: $\hat{q}_t \sim \softmax(f_{\theta}(\hat{q}_{<t}))$.

In this work, we also study attacks on the recent watermark scheme \textit{BitMark}~\citep{kerner2025BitMark} that was developed for the multi-scale bitwise autoregressive Infinity generator~\citep{han2025infinity}.
Rather than generating latent tokens one-by-one, Infinity samples the latent $z$ by generating a sequence of residuals at different scales, using the next scale prediction approach (VAR~\citep{tian2024nspvar}).
In contrast to regular left-to-right token-level autoregressive decoding, all elements in one scale are sampled independently from each other, conditioned on all previous scales.
Where VAR predicts tokens from a fixed vocabulary, Infinity predicts residuals for every channel of the latent independently, and the residuals are quantized to two levels (one bit) per channel.

\subsection{Watermarking for Autoregressive Image Generation}
\label{sec:kgw}
The existing watermarking schemes for AR image generators are derived from \emph{KGW watermarking} originally proposed for large language models~\citep{kgw,kirchenbauer2024reliability}.
First, at every generation step, the previously decoded tokens %
are used to %
partition the vocabulary of the generator into two sets, a ``green'' set $G_i$, and a ``red'' set $R_i$. %
It is done by first computing a hash $o_i$ based on the previous $l$ tokens and a secret key $\kappa$: $o_i = \text{hash}(y_{i-1}, \dots, y_{i-l}, \kappa )$.
The hash $o_i$ is then used to 
randomly select a fraction $\gamma$ of tokens from the vocabulary \vocabulary to form the set of green tokens $G_i$ of size $\lfloor \gamma * |\vocabulary| \rfloor$.
Finally, the model's distribution over \vocabulary is biased to favor tokens from the green set by adding~$\wmstrength$ to their logits:
\begin{align}
    m_i[v] &= \mathbf{1}_{G_i}(v)  \enspace , \\
    p'_{\theta}(y_i | y_{<i}) &= \text{softmax} (f_{\theta}(y_{<i}) + m_i * \wmstrength) \enspace . \label{eq:wm_boost}
\end{align}

In order to verify the watermark for the given sequence $(y_1, \dots, y_T)$, the green sets $G_i$ are recomputed for every token given the preceding context and the number of green tokens is counted.
Then, a one-sided right-tailed statistical test is used to assess how unlikely the observed number of green tokens ($N_g = \#_{\text{green}}$) is, if the selection was truly random.
The null hypothesis is that $N_g$ is distributed according to an unbiased binomial distribution: $H_0: N_g \sim \text{Binomial}(T, \gamma)$.
The right-tailed p-value $p = \text{Pr} (X \geq N_g)$ is computed and $H_0$ is rejected for a sequence if %
$p$ is lower than a threshold chosen for a specified false positive rate (FPR).

In this work, we study removal and forgery attacks on a representative selection of two distinct types of recently proposed watermarking schemes for autoregressive image generation models.

\textbf{Token-based watermarking} includes \textbf{IndexMark}~\citep{tong2025indexmark}, \textbf{WMAR}~\citep{jovanovic2025wmar}, and \textbf{ClusterMark}~\citep{lukovnikov2025clustermark}. They closely follow KGW watermarking and use the preceding tokens to partition the vocabulary into green and red sets, where green tokens' logits are boosted by $\delta$ during sampling according to ~\Cref{eq:wm_boost}. Importantly, all of them \emph{fine-tune the VQ-VAE encoder} to improve the retrieval accuracy of original token indexes from generated images.

\textbf{Bitwise-autoregressive watermarking} includes \textbf{BitMark}~\citep{kerner2025BitMark}. BitMark treats sampled bits in one scale as a sequence, and biases individual bits based on $l$ previous bits and a fixed green set n-gram $G$ of size $l+1$. For example, the default setting $l=1$ and $G_2 = \{01, 10\}$ leads to a higher prevalence of alternating bits in generated image latents. Importantly, BitMark serves as a detection mechanism to filter synthetic images out of future training pipelines, which is further facilitated by its \emph{radioactive} property, enabling the watermark to propagate across model training.

Appendix~\ref{sec:supp:scheme_details} provides detailed descriptions of each scheme. Other schemes~\citep{hui2025autoregressivelexical, wu2025creweight, dai2025safebvar} are not included in our evaluation due to their code being unavailable.

\section{Attacks on Autoregressive Image Watermarks}
\label{sec:attacks}

In this work, we evaluate existing removal and forgery attacks on AR model watermarking and propose three new attack methods that specifically target AR generative models. The first is \emph{Vector-Quantized Regeneration (VQ-Regen)}, a removal attack that aims at disrupting the precise token index recovery by regenerating a watermarked image from close-by tokens. The second is \emph{Latent Encoder Optimization (\advoptname)}, which can be applied as both a removal and a forgery attack. The third is \emph{Frequency Injection Forgery}, %
which performs watermark forgery by injecting localized peaks into a target image's Fourier representation. 

\subsection{Vector-Quantized Regeneration Attack (VQ-Regen)}
\label{sec:qregen}
In the Vector-Quantized Regeneration Attack (VQ-Regen) for watermark removal, a VQ-VAE codebook is used to generate a perturbed reconstruction $x'$ as follows.
First, the image is encoded: $z=\encoder(x)$, where $z \in \R^{d \times h \times w}$.
 For each spatial location $(i, j)$ %
 all codebook vectors in \codebook are sorted by their Euclidean distance from $z_{\cdot, i, j}$.
This produces a ranking $s_{i,j}$ of the vocabulary \vocabulary, such that $s_{i, j}[1]$ is the closest and $s_{i,j}[k]$ is the k-th closest token index.
Next, the attacked token map $t'\in \vocabulary^{h \times w}$ is constructed by replacing the standard quantization step with the selection of the $k$-th nearest token index: $t'_{i,j} = s_{i,j}[k]$.
Finally, the attacked image is produced by mapping $t'$ back to latent space using codebook lookup $z' = \codebook[t']$, and decoding: $x' = \decoder(z')$.
This is explained in detail in \cref{alg:supp:vqregen} in the Appendix. Note that setting $k=1$ recovers the (unattacked) standard nearest-neighbor reconstruction, while $k > 1$ induces a controlled structural deviation in the generated image.

\subsection{Latent Encoder Optimization (\advoptname)}
\label{sec:encopt}
This attack adds a subtle pixel-level perturbation $\pert$ to an image $x$, which propagates through the autoencoder to shift the latent representation in a targeted, gradient-guided manner, while minimally affecting the image appearance.
For removal, the goal is to move the latents away from the current latents, such that different tokens are decoded, and the watermark signal is lost.
For forgery, the latents of the authentic cover image are moved towards those of a watermarked reference image. 
This is similar to the attacks proposed in~\citep{Muller2025semanticforgery,jain2025forgingremoving}, except we only use the VAE.

\paragraph{\advoptname-Removal.}
A pixel-delta $\pert$ is optimized under a $p$-norm constraint,
such that $\encoder(x+\pert)$ moves away from the initial
$\encoder(x)$:
\begin{equation}
    \pert^*
    = \argmax_{\pert}
      \left|\encoder(x+\pert)-\encoder(x)\right|^2
    \qquad \text{s.t.}\quad |\pert|^p < c.
\end{equation}

\noindent
Here, $\encoder$ can be any encoder that maps pixels to a latent
space, either from the model's own VAE with the same weights
(leading to a white-box setting), from another, unrelated VAE
(black-box), or a VAE of similar architecture (grey-box).

\paragraph{\advoptname-Forgery.}
Here, $\pert$ is optimized such that it moves the latent of an
\textit{authentic cover} image $x_c$ towards the latent
$\encoder(x_w)$ of an (unrelated) \textit{watermarked} image $x_w$
that was obtained beforehand:
\begin{equation}
    \pert^*
    = \argmin_{\pert}
      \left|\encoder(x_c+\pert)-\encoder(x_w)\right|^2
    \qquad \text{s.t.}\quad |\pert|^p < c.
\end{equation}

\paragraph{\bitadvoptname-Removal for BitMark.} BitMark~\citep{kerner2025BitMark} proposed the (unsuccessful) \emph{BitFlipper} attack to highlight the robustness of BitMark in a white-box setting which assumes knowledge of \encoder, scale resolutions, and green set $G$ (we refer to this as the \emph{white-box{}$+$} setting).
Here, we present the \bitadvoptname-Removal attack under the same assumptions.
In our attack, 
(1) the target image $x$ is converted to a sequence of (quantized) residuals similarly to the regular encoding procedure.
Then, (2) the bits that would reduce the green count most are identified. With the knowledge that BitMark by default uses $G = \{\text{0\textrightarrow1}, \text{1\textrightarrow0}\}$, the positions of trigrams $010$ and $101$ are identified as flipping targets, since flipping the $1$ in $010$ would decrease the green token count by 2. 
With the target bits identified, (3) an adversarial optimization similar to \advoptname-Removal is performed on $x$ with the objective to flip these bits using an L1 loss.
This method is formally elaborated in~\cref{supp:alg:bitopt} in the Appendix.

\subsection{Frequency Injection Forgery for BitMark}
\label{sec:fftforgery}
We conducted a steganographic analysis of BitMark-generated images by averaging 5,000 samples and found that the residual signal verifies against the watermark detector with p-values on the order of $10^{-33}$ (see Fig.~\ref{fig:fftforgery}).
Yet, adding the averaged signal to authentic cover images (for forgery) or subtracting it from watermarked target images (for removal), similarly to the averaging attack~\citep{yang2024steganalysisdigitalwatermarkingdefense}, resulted in poor image quality while not sufficiently affecting detection metrics.

Inspired by the presence of frequency peaks in the average, we develop a simple forgery attack in which a pattern of magnitude peaks is injected into the frequency representation of an authentic cover image at regular intervals along the diagonals to introduce a regular pixel pattern. We set random phases for each color channel to reduce the saliency of the injected pattern and reduce visibility. By repeatedly querying the target detector with a separate dataset of 100 attack instances with varying peak locations and magnitudes, we find three viable settings (denoted as A, B, and C), each yielding a desirable trade-off between image quality and attack success.
A detailed description of this tuning phase, the injection procedure, and exact settings are provided in Section~\ref{sec:supp:experimental_settings}, as well as~\cref{alg:supp:fftforgery} in the Appendix.

\begin{figure}[t]
    \centering
    \includegraphics[width=1.0\columnwidth]{figures/fft_concept_setting2.pdf}
    \caption{\textbf{Frequency Injection Forgery for BitMark.}
    \textit{(Left)} Averaging 5,000 BitMark-watermarked images reveals a structured spatial artifact  detectable by the watermarking scheme ($p = 3\times10^{-33}$), and its Fourier transform on the right.
    \textit{(Center)} An authentic cover image and its Fourier transform are not detected as watermarked ($p = 0.02$, $\text{PSNR} = \infty$).
    \textit{(Right)} Injecting the frequency peaks at positions indicated by red circles into the authentic cover image achieves high visual fidelity ($\text{PSNR} = 35.9$~dB) 
    while successfully triggering BitMark's detector ($p = 1\times10^{-19}$).}
    \label{fig:fftforgery}
\end{figure}

\vspace{-2pt}

\section{Evaluation}
\label{sec:evaluation}

\label{sec:evaluation:setup}

\subsection{Target Watermarking Schemes}
\label{sec:evaluation:tokenindex-based:deployed_targetmodels}
We investigate attacks on three discrete token watermarking schemes (IndexMark, WMAR, ClusterMark) and one for bitwise autoregressive models (BitMark).  Each scheme is tested with multiple generative models and different settings, as listed below. 
Unless otherwise specified, we use the default settings in the code or paper provided with each watermarking scheme.
    
    \textbf{IndexMark}~\citep{tong2025indexmark} is deployed with the LlamaGen~\citep{sun2024llamagen} model in both the GPT-B (generating images of size $256\times256$) and GPT-L ($384\times384$) variants.
    Both models are class-conditional, using 1k ImageNet~\citep{russakovsky2015imagenet} classes.

    \textbf{WMAR}~\citep{jovanovic2025wmar} is tested with different models: RAR-XL~\citep{yu2025rar} (class-conditional, $256\times256$), Taming Transformers~\citep{esser2021taming} (class-conditional, $256\times256$) and Anole~\citep{chern2024anole} (text-conditional, $512\times512$). 
    WMAR is used with the fine-tuned encoder and decoder, but without the synchronization layer in order to study the core method's robustness in isolation. In Appendix~\ref{sec:supp:wmar_synchronization}, we provide alternative results, and show that the layer degrades WMAR's overall performance while only slightly mitigating geometric transformations.

    \textbf{ClusterMark}~\citep{lukovnikov2025clustermark} is tested with LlamaGen both with GPT-B and GPT-L, as well as RAR-XL in the default settings with 64 clusters. We also use the fine-tuned cluster classifier.

    \textbf{BitMark}~\citep{kerner2025BitMark} is tested with Infinity-2B~\citep{han2025infinity} ($\lvert V_{d}\rvert=2^{32}$, text-conditional generation, image size of $1024\times1024$) and default settings (green set $G = \{01, 10\}$ active on all scales, and $\delta=2$).

\subsection{Deployed Attacks}
\label{sec:evaluation:tokenindex-based:deployed_attacks}
We evaluate two categories of attacks.
Full experimental details are provided in Appendix~\ref{sec:supp:experimental_settings}. A full breakdown of computational requirements is provided in Appendix~\ref{sec:supp:compute}.

\noindent\textbf{Our attacks.} We evaluate \textit{\advoptname-Removal} (\advoptname-R) and \textit{\advoptname-Forgery} (\advoptname-F), our optimization-based adversarial attacks, as well as \textit{VQ-Regen}. Specifically for BitMark, we also evaluate \textit{\bitadvoptname-Removal} (\bitadvoptname-R), a variant with additional knowledge of the watermarking settings, as well as our \textit{Frequency Injection Forgery} attack (Freq.Inj.F.).

\noindent\textbf{Existing attacks.} We evaluate the watermarking schemes against strong \textit{Perturbations} (JPEG compression, additive noise, color jitter) and geometric transformations. For visual examples of perturbations, see~\cref{fig:supp:perturbation_examples} in the Appendix. We further add diffusion-based regeneration attacks, \textit{Regen.} and \textit{Rinse}~\citep{ZhaZhaSu2024invisibleimagewatermarksprovably}, and \textit{CtrlRegen+}~\citep{liu2025ctrlregen}. For BitMark, we also include their \textit{BitFlipper} attack~\citep{kerner2025BitMark} as well as \emph{Reprompt-Forgery}~\citep{Muller2025semanticforgery}, which was originally intended to subvert diffusion model watermarking. We found that the latter fails against token-based schemes and hence omitted these results.
Finally, we also include the \emph{Next-Frame Prediction} attack~\citep{qiu2026nextframe} for BitMark. We tried applying it to token-level watermarks as well, but found the attack to severely degrade image quality because those models generate images at a lower resolution (compared to Infinity's $1024 \times 1024$). Adjusting attack parameters (reducing the shift from the default of 40 to only 10 pixels) did not resolve this issue. Hence we omitted these results.

\subsection{Datasets and Metrics}
\label{sec:evaluation:tokenindex-based:datasets_metrics}
For both removal and forgery, each attack is performed against 1,000 watermarked target images generated with each of the multiple target models deployed per watermarking scheme. \advoptname-Removal and -Forgery, as well as \bitadvoptname-Removal attacks are an exception and are evaluated on 100 images, due to the large number of settings and higher runtime requirements for a full optimization run.
For forgery attacks, we use images from the COCO dataset~\citep{lin2014coco} as covers.
In line with previous work~\citep{tong2025indexmark,jovanovic2025wmar,lukovnikov2025clustermark,kerner2025BitMark}, we report average TPR@FPR=1\% as an accuracy metric, as well as median p-values as a threshold-agnostic alternative to TPR. We also report quality degradation between attacked images and watermarked images (removal), as well as forged images against the original ones (forgery), in terms of PSNR$\uparrow$ and LPIPS$\downarrow$~\citep{zhang2018lpips} scores respectively. Unless otherwise stated, all results are reported as aggregates across multiple deployed models for each watermarking scheme, as well as different proxy models used by the attacker.

For BitMark, we additionally include accuracy metrics on radioactive data in different settings: We fine-tune Infinity-2B and Stable Diffusion v2.1~\citep{RomBlaLor2022stablediffusion} on 1,000 (watermarked/unwatermarked) images generated with Infinity-2B using captions from COCO as text-conditioning.
During this fine-tuning, different ratios of watermarked to unwatermarked images are used.

\subsection{White-box, Grey-box, and Black-box Settings}
We categorize attacks using proxy models or encoders by their access to the verifier's encoder.

In black-box settings (\blackbox), the attacker uses an entirely different encoder or model (or none) from the one deployed by the verifier. We aggregate results across these attacker encoders to evaluate attack success using readily available models, reflecting realistic attack scenarios.

In grey-box settings (\greybox), the attacker uses an encoder closely related to the verifier's encoder. For token-based watermarks, this is the non-fine-tuned version of the verifier's encoder. For bit-autoregressive watermarks, we use models with the same architecture and training procedure but different vocabulary sizes ($\lvert V_d\rvert=2^{16}$, $2^{24}$, or $2^{64}$), while the verifier uses the default $\lvert V_d\rvert=2^{32}$. We aggregate results across these vocabulary sizes.

In white-box settings (\whitebox{}), the attacker uses the exact same fine-tuned encoder as the verifier. These settings provide insight into each attack's capabilities and serve as an upper bound on attack success. The BitFlipper and \bitadvoptname-Removal against BitMark are denoted as having white-box{}$+$ access (\whitebox{}$+$), meaning that they require additional knowledge of watermarking parameters.

\vspace{-6pt}
\paragraph{Query Access to the Target Detector.}
\label{sec:query_access}
Unless otherwise stated, all attacks are evaluated without query access to the target detector. Frequency Injection Forgery requires implicit detector access during tuning to identify suitable injection parameters, which can then be reused to generate an unlimited number of additional attack instances without further queries. \advoptname attacks are evaluated with explicit query access: We optimize for 300 steps, query the detector every 10 steps, and retain the most successful iterate. Further details are provided in Section~\ref{supp:sec:evaluation:deployed_attacks}.

\begin{figure*}[tp]
  \centering

  \begin{minipage}{\linewidth}
    \centering
    \captionof{table}{%
      Results for \textbf{token-based methods}, aggregated over multiple
      target models deployed with each watermarking scheme.
      \blackbox is black-box, \greybox is grey-box, \whitebox is white-box.
      P-values are reported as medians, and all other metrics as means.
      For removal, attackers target lower TPR (higher p-values);
      for forgery, higher TPR (lower p-values).}
    \label{tab:token-based}
    \scalebox{0.71}{%
      \input{tables/tokenbased_main}%
    }
  \end{minipage}
  \par\medskip

  \includegraphics[width=0.98\linewidth]{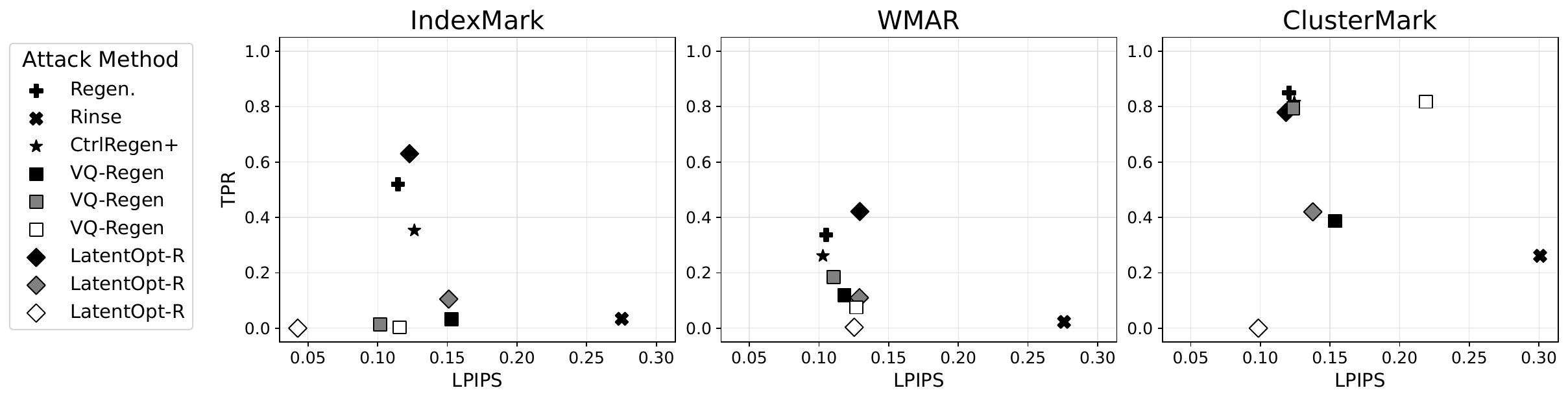}
  \caption{\textbf{Trade-off between watermark removal success (TPR) and perceptual distortion (LPIPS)}
for token-based methods.
Marker shapes distinguish attack methods, while black, gray, and white
fills indicate black-box, gray-box, and white-box access, respectively.
The attacker favours outcomes toward the lower-left corner.}
  \label{fig:frontierplot}
  \vspace{-9pt}
\end{figure*}

\subsection{Results for Token-Based Watermarks}
\begin{figure}[t]
    \centering
    \includegraphics[width=1.0\columnwidth]{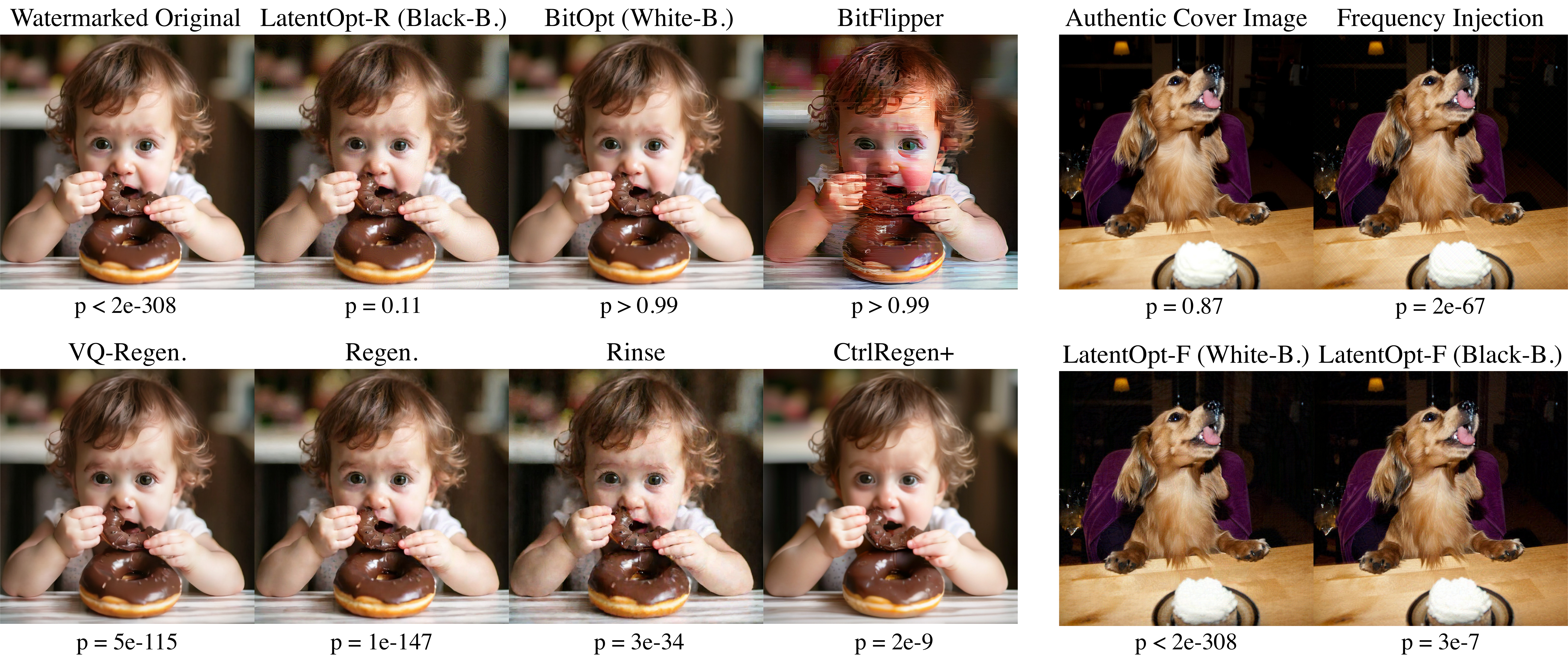}
    \caption{\textbf{Qualitative examples} of removal attacks (left) and forgery attacks (right) on BitMark, deployed with Infinity-2B. \advoptname-R and \advoptname-F attacks are capped at $\|\pert\|^{\infty}=\frac{8}{255}$.\\More visual examples are provided in Appendix~\ref{sec:supp:full_results}.
    }
    \label{fig:main_qualitative_examples}
\end{figure}

The results for both removal and forgery attacks on token-based watermarks are shown in \cref{tab:token-based}.
\cref{fig:frontierplot} shows the trade-off between removal attacks.
\cref{fig:main_qualitative_examples} shows qualitative examples for a subset of evaluated attacks.
A model-wise breakdown of all reported values is shown in Appendix~\ref{sec:supp:full_results}. 
Overall, the results indicate susceptibility to removal attacks even at the permissive threshold of FPR=1\%. However, forgery fails without access to the verifier's exact latent encoder, i.e. the white-box setting.%

Firstly, we observe that baseline image perturbations and geometric transformations can disturb the watermark detection for all schemes.
Note that, while discrete token watermarking schemes have already been shown to be vulnerable to geometric transformations~\citep{jovanovic2025wmar}, this can be addressed by adding an additional synchronization layer like SyncSeal~\citep{fernandez2025syncseal}.

Secondly, we observe that Regen. and CtrlRegen+, while generally effective with a success rate of 48\%-74\% for IndexMark and WMAR, display high pixel distortion with ${\sim}24$ dB PSNR, while maintaining relatively high perceptual quality (low LPIPS of ${\sim}0.11$). Rinse, while effective, shows considerable quality loss in terms of both PSNR and LPIPS.
ClusterMark generally exhibits more robustness against these attacks.

Similarly, VQ-Regen causes severe pixel-wise distortion (low PSNR), while maintaining relatively high perceptual quality (low LPIPS). Generally, VQ-Regen in the black-box setting achieves slightly higher removal success with comparable quality degradation. Interestingly, for ClusterMark, using similar or even the exact same proxy VQ-VAE as the deployed target model renders the attack less effective: The close alignment of the proxy VQ-VAE to the target in the grey- and white-box setting actually hinders the attacker from removing the watermark, as the substituted tokens are still likely to fall into correct clusters, maintaining the integrity of green set assignments with consecutive tokens.

VQ-Regen provides the most consistently favourable trade-off between image quality (LPIPS) and watermark removal success (TPR), while our \textit{\advoptname-Removal} attack achieves a comparable trade-off only in some settings, as shown in \cref{fig:frontierplot}.

We find that \emph{forgery} is significantly more difficult, failing in all but some white-box settings. Here, forgery is most effective when the LlamaGen model is used by both the verifier and attacker, as is the case for part of the setup for IndexMark and ClusterMark. Please refer to \cref{tab:supp:full_results:IndexMark:GPT-B,tab:supp:full_results:IndexMark:GPT-L,tab:supp:full_results:ClusterMark:GPT-B,tab:supp:full_results:ClusterMark:GPT-L} in the Appendix for a breakdown of LlamaGen-specific results. A possible explanation is the relatively small embedding dimension (8) used in LlamaGen's VQ-VAE, in contrast to the $256$ dimensions in all other models. Entering correct Voronoi cells for each token is easier with fewer embedding dimensions because the decision boundaries depend on fewer orthogonal directions, making the target region geometrically less fragmented and reducing the number of independent constraints that the forgery attack instance must simultaneously satisfy.

Finally, we conducted a systematic study of the transferability between target and proxy encoders under the \advoptname-Removal attack, which is provided in Appendix~\ref{sec:pertubation_budget_effect}.

\begin{table}[tbp]
    \centering
    
    \caption{Results for removal attack (left), as well as watermarked, radioactive data baseline, and forgery (right) attacks for \textbf{BitMark}.
    \blackbox is black-box, \greybox is grey-box, \whitebox is white-box. \whitebox{}$+$ indicates having additional access to watermarking parameters.
    \advoptname and VQ-Regen (\blackbox \& \greybox) were performed with several attacker models, or specifically with LlamaGen (indicated by $^\clubsuit$).
    P-values are reported as medians, and all other metrics as means.
    For removal, attackers target lower TPR (higher p-values); for forgery, higher TPR (lower p-values).}
  
    \begin{minipage}[t]{0.50\textwidth}
        \centering
        \resizebox{\linewidth}{!}{%
            \input{tables/bitmark_main_removal}
        }
    \end{minipage}
    \hfill
    \begin{minipage}[t]{0.49\textwidth}
        \centering
        \resizebox{\linewidth}{!}{%
            \input{tables/bitmark_main_forgery}
        }
    \end{minipage}
    \label{tab:bitmark}
\end{table}

\vspace{-1pt}

\subsection{Results for BitMark}
\cref{tab:bitmark} reports the results of the studied attacks on BitMark. \cref{fig:main_qualitative_examples} shows qualitative examples for a subset of evaluated attacks.

We observe that \textit{BitMark is extremely robust to removal attempts}, likely due to the large number of biased bits, which results in higher statistical significance of the embedded signal during verification. 
Diffusion regeneration attacks (Regen., Rinse, CtrlRegen+) are largely ineffective, only slightly increasing the measured p-values, remaining below the detection threshold.
Our VQ-Regen is similarly ineffective.
For \advoptname-Removal, using LlamaGen's VAE encoder as the attacker's proxy model (indicated by $^\clubsuit$) reduces TPR below 90\% while maintaining PSNR above 32~dB. This represents a slight improvement in removal success over the aggregated results for the other proxy encoders.
While the white-box \advoptname-Removal fails to significantly disturb BitMark within the perturbation budget $c=\frac{8}{255}$, our BitMark-specific \textit{\bitadvoptname}-R attack is able to completely erase the watermark while maintaining a PSNR of over $46$ dB.

Conversely, \textit{forgery of BitMark is more effective than for token-based watermarks}. First, we find that Reprompt-Forgery, originally proposed against diffusion model watermarks, is highly effective against BitMark. Furthermore, \advoptname-Forgery is able to push the TPR above 50\% in the black-box setting, while maintaining PSNR > 31.
Again, LlamaGen's encoder is significantly more successful than the average proxy encoder in black-box settings with a TPR of $88\%$, outperforming even the average grey-box encoder. This indicates that an attacker can benefit from evaluating multiple unrelated proxies, as they can differ significantly in terms of attack success.
In the white-box setting, forged images yield low p-values that approach those of original, unmodified watermarked images. Frequency injection is able to achieve low p-values without requiring a proxy encoder.

\paragraph{Forgery against BitMark enables the Watermark Mimicry scenario.}
Our \advoptname and Frequency Injection attacks can embed the watermark signal into arbitrary cover images, enabling the \emph{Watermark Mimicry} scenario described in Section~\ref{sec:introduction}.
While Reprompt-Forgery can reliably trigger false acceptance in BitMark's verifier, it does not enable the same scenario: the attack can only produce synthetic images and therefore cannot shield authentic content from incorporation into training data. However, it does enable the false attribution scenario for which it was originally proposed~\citep{Muller2025semanticforgery}.

\begin{figure}[t]
    \centering
    \includegraphics[width=0.85\columnwidth]{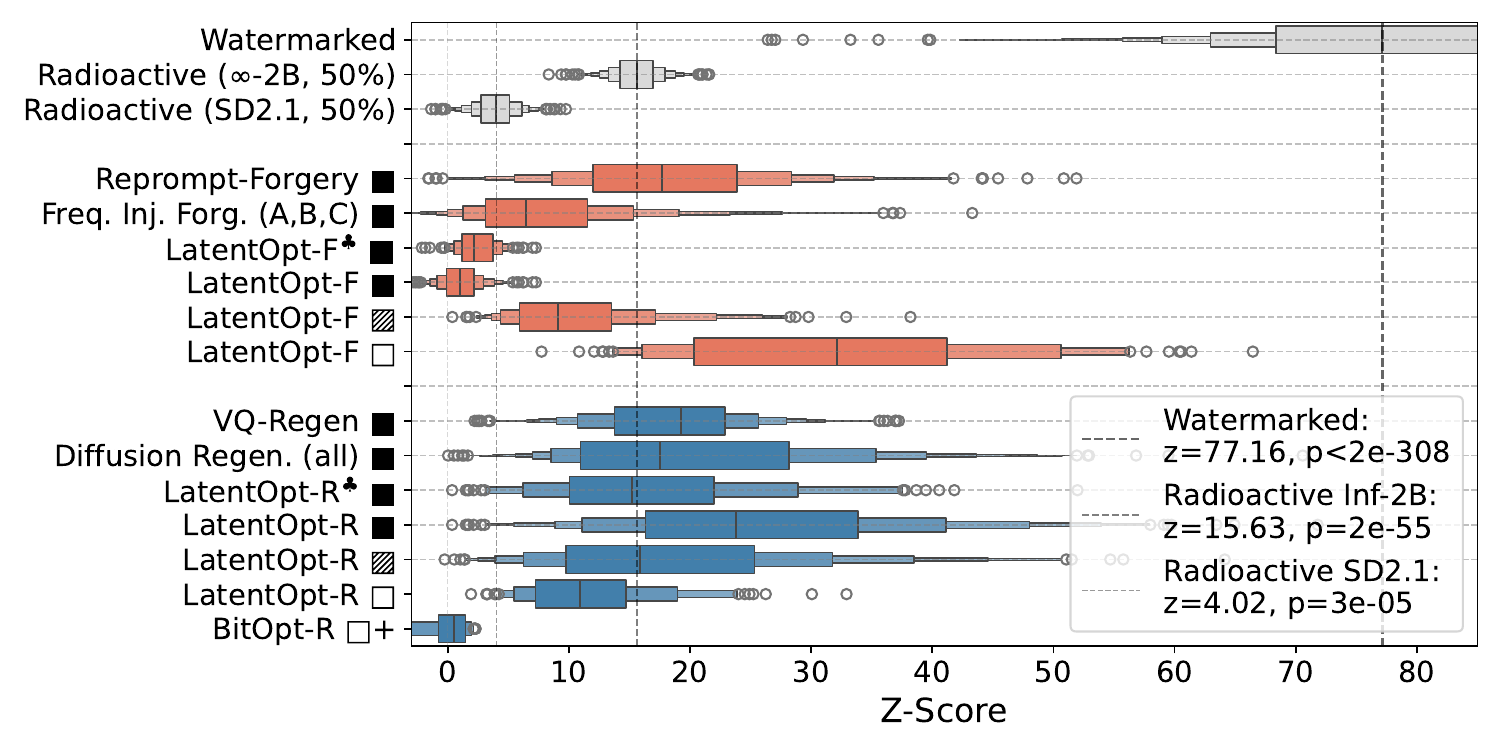}
    \caption{\textbf{Distributions of z-scores for different images as verified by BitMark}, including original watermarked and radioactive images (\textcolor{mygray}{grey}), forgery (\textcolor{myred}{orange}) and removal (\textcolor{myblue}{blue}) attack instances.}
    \label{fig:main_boxplot}
\end{figure}

\paragraph{Can attacks against BitMark be mitigated by adjusting the detection threshold?}
The results for token-based schemes are clear-cut in that removal is generally feasible even at the permissive operating point of FPR $=1\%$. BitMark presents a more complex trade-off between removal, forgery, and radioactivity. We therefore examine whether adjusting its detection threshold can mitigate removal and forgery attacks while preserving detection of radioactively generated images.

The box-plot in \cref{fig:main_boxplot} shows the spread of z-scores for the different forgery and removal attacks from \cref{tab:bitmark}, compared to the z-scores of images generated by generators (Infinity-2B, SD2.1) affected by radioactivity, i.e. generators that have been fine-tuned on 50\% watermarked images. The median z-scores for the radioactive model data, as well as the original watermarked images, are shown as dashed lines. 
We observe that removal attacks in certain black-box settings cannot be clearly separated from Frequency Injection forgery and a significant portion of their distribution mass overlaps. Hence, perfectly protecting against both removal and forgery by adjusting the detection threshold is not possible, even for a black-box attacker.
Furthermore, in order to enable the detection of radioactively generated images, the detection threshold has to be set below the bulk of the corresponding distributions.
However, in doing so, the verifier also enables false positive detection of forgery attack instances. For example, frequency injection forgery well surpasses any reasonable threshold that includes detection of images generated by the radioactive SD2.1 model trained on 50\% watermarked training data.
Note that assuming that a model would have been fine-tuned on 50\% watermarked data is already a strong assumption. Enabling detection of even lower fractions increases the opportunity for forgery attempts. For example, the more realistic scenario of an Infinity-2B model being fine-tuned on 10\% watermarked data ($z=3.48$, $p=2.4\times10^{-4}$) completely rules out being able to both detect radioactive images and prevent forgery attacks.
Overall, although watermark owners may be less concerned about forgery in the data filtering setting, Watermark Mimicry nonetheless enables third parties to leverage it for protecting their intellectual property.

\clearpage

\section{Related Work}
\label{sec:conclusion:relatedwork}

Prior to the recent watermarking schemes for autoregressive (AR) image generation models, several in-generation watermarking schemes were proposed for latent diffusion models (LDMs)~\citep{Wen2023TreeRing,Yang2024GaussianShading,CiYanSon24RingID,Gunn2024Undetectable,Fernandez2023Stable}, which rely on DDIM inversion~\citep{song2021denoising,hong2024exactinversion} for watermark verification. A subsequent line of work has studied removal~\citep{ZhaZhaSu2024invisibleimagewatermarksprovably,liu2025ctrlregen,jain2025forgingremoving,kassis2025unmarker,AnDinRab2024Benchmarking,shamshad2026raven,yang2024steganalysisdigitalwatermarkingdefense,lukDiaFen2024leveraging,saberi2023robustness,qiu2026nextframe} and forgery~\citep{Muller2025semanticforgery,jain2025forgingremoving,yang2024steganalysisdigitalwatermarkingdefense} attacks against these schemes.
Of these, we include regeneration-based removal attacks~\citep{ZhaZhaSu2024invisibleimagewatermarksprovably,liu2025ctrlregen} in our comparison, and as discussed in~\cref{sec:fftforgery}, we also evaluate the averaging attack~\citep{yang2024steganalysisdigitalwatermarkingdefense}. We further tried UnMarker~\citep{kassis2025unmarker} for removal, but its prohibitive memory requirements precluded its use on images larger than 384$\times$384, ruling out much of our evaluation setup. We were unable to obtain source code for the recent RAVEN~\citep{shamshad2026raven} attack.

While the attacks of~\citep{Muller2025semanticforgery,lukDiaFen2024leveraging} could in principle target in-generation watermarks for AR models, they are tailored to LDM-specific mechanics - most notably their reliance on DDIM inversion for forgery. Finally, the optimization-based attacks of~\citep{AnDinRab2024Benchmarking,jain2025forgingremoving} on LDM watermarks are conceptually similar to our \advoptname attack, but rely on LDM-specific VAE encoders and use different optimization objectives.

\paragraph{Protecting Content from Generative AI.}
\label{par:extended_related_works}
A growing line of work aims to prevent authentic content from being incorporated into generative model training. These methods fall broadly into two groups: proactive perturbation of the data and post-hoc removal from the model. On the perturbation side, Glaze~\citep{shan2023glaze} aims to block style mimicry by adding an adversarial perturbation that shifts the image in a style-feature space toward a different artist's style, so that fine-tuning learns the wrong style.
  AdvDM~\citep{liang2023advdm} and its extension Mist~\citep{liang2023mist}  aim to block diffusion models from using the content by adversarially perturbing the image to maximize the diffusion training loss and thereby make it a misleading
  training example. PhotoGuard~\citep{salman2023photoguard} and Anti-DreamBooth~\citep{vanle2023antidreambooth} take a complementary approach,
   protecting images against malicious diffusion-based editing and personalized subject-driven generation. Relatedly, data poisoning perturbs images so they degrade models trained on them~\citep{shan2024nightshade,huang2021unlearnable,fowl2021adversarial}. Our Watermark Mimicry of radioactive watermarks extends these attempts by potentially enabling a given image to be excluded from the training dataset in the first place.
A second line of work intervenes after training, erasing already-memorized content. Fine-tuning-based methods modify
  diffusion weights to forget specific styles, objects, or
identities~\citep{gandikota2023erasing,kumari2023ablating,zhang2023forgetmenot}. Closed-form and continual-learning
  variants offer more structured erasure with formal stability
guarantees~\citep{heng2023selective,belrose2023leace,gandikota2024uce}. A more recent thread targets robust erasure
  that resists adversarial prompts and re-learning attempts~\citep{srivatsan2025stereo,gong2024rece,huang2024receler}.

\section{Limitations}
\label{sec:limitations}

\textbf{White-Box Setting.} Some versions of our attacks target the white-box setting. While rarely applicable in practice, we believe it offers meaningful analysis, highlighting possible attack vectors that adversaries could further develop to work in more realistic scenarios. In addition, they provide an upper bound on the attackability of a watermarking method.

\textbf{Frequency Injection Parameters.} Although this attack relies on hand-tuned parameters, it is the first demonstration of BitMark forgery without access to the generator or its VQ-VAE encoder, motivating further work on image-agnostic attacks.

\section{Conclusion}
\label{sec:conclusion}
In this work, we present the first study of watermark forgery and removal attacks against autoregressive (AR) image generators. Our evaluation reveals distinct vulnerabilities across the investigated schemes. Token-based AR watermarks~\citep{tong2025indexmark,jovanovic2025wmar,lukovnikov2025clustermark} are susceptible to removal even in black-box settings, while forgery remains challenging. BitMark~\citep{kerner2025BitMark}, in contrast, exhibits exceptional resistance to removal attacks. However, it is vulnerable to black-box forgery at detection thresholds that also identify radioactively generated images. Such thresholds support synthetic content filtering but also enable \emph{Watermark Mimicry}, allowing authentic content to be shielded from inclusion in future model training.

We hope that our work draws attention to bitwise autoregressive watermarking, which proved exceptionally robust to removal attempts, and provides a foundation for future research on watermarking for autoregressive image generation, including methods resistant to forgery.

\clearpage

\begin{ack}
This work was funded by the Deutsche Forschungsgemeinschaft (DFG, German Research Foundation) under Germany’s Excellence Strategy - EXC 2092CASA - 390781972 and by the Ministry of Culture and Science of North Rhine-Westphalia as part of the Lamarr Fellow Network.
Minh Pham would like to acknowledge support from the AI Research Institutes program supported by NSF and USDA-NIFA under AI Institute: for Resilient Agriculture, Award No. 2021-67021-35329.
Niv Cohen is a Jacques Lewiner Career Advancement Chair Fellow at the Technion.
\end{ack}

\bibliographystyle{plain}
\bibliography{main}

\clearpage

\appendix

\etocdepthtag.toc{appendix}

\etocsettagdepth{appendix}{subsubsection}
\etocsettagdepth{none}{none}

\renewcommand{\contentsname}{Contents}
\tableofcontents

\newpage
\raggedbottom

\section{Overiew and Reference of Target Watermarking Schemes}
\label{sec:supp:scheme_details}

We provide detailed description of each targeted watermarking scheme for reference:

\begin{figure}[t]
    \centering
    \hspace{0.016\columnwidth}
    \includegraphics[width=0.98\columnwidth]{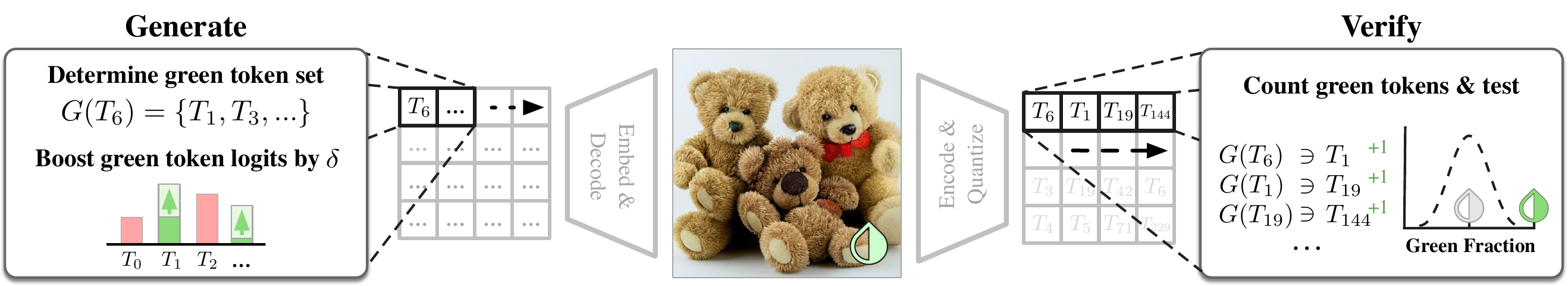}
    \caption{\textbf{Overview of token-based watermarking (IndexMark, WMAR, ClusterMark).}
    During \textit{generation} (left), green tokens (e.g., $T_1$) are boosted. During \textit{verification} (right), a statistical test is performed based on the fraction of present green tokens to determine the presence of a watermark.}
    \label{supp:fig:kgw_overview}
\end{figure}

\subsection{Token-based Watermarking Schemes}
\label{sec:supp:scheme_details_token-based}
\Cref{supp:fig:kgw_overview} shows a schematic overview of the token-based watermarking schemes. 

\textbf{IndexMark}~\citep{tong2025indexmark} divides the vocabulary into a set of pairs of tokens, maximizing the total token similarity within the pairs.
The pairing is formulated as a maximum-weight perfect matching problem using cosine similarity between codebook embeddings.
One of the tokens in each pair is assigned to the green set and the other to the red.
In order to embed the watermark in the generated token sequence, selected red tokens are replaced with their green counterparts.
The selection is guided by the relative probabilities of the paired tokens, prioritizing replacements for which the model does not strongly prefer the red token.
A replacement threshold controls the trade-off between watermark strength and image quality.
This minimizes image degradation and introduces a detectable watermark signal.
A separate Index Encoder is trained to improve token reconstruction accuracy, while the original encoder, codebook, and decoder remain frozen.
During detection, the image is re-encoded, and the proportion of recovered green tokens is compared against a statistical threshold based on an expected green-token proportion of $0.5$ for unwatermarked images.

\textbf{WMAR}{$_\mathrm{{(NeurIPS'25)}}$}~\citep{jovanovic2025wmar}, following KGW, uses the preceding tokens and a secret key to partition the vocabulary into green and red sets, where green tokens' logits are boosted during sampling.
Detection re-encodes the image, reconstructs these partitions, and tests whether the number of green tokens is significantly larger than expected under the binomial null model.
In order to improve robustness, both a copy of the pixel-latent encoder and the decoder are fine-tuned with perturbations.
Specifically, it aims to improve \emph{reverse cycle consistency}, i.e. the reconstruction accuracy of originally generated tokens from watermarked images.
The training objective encourages agreement between the original codebook embeddings and the re-encoded latents, while pixel and perceptual losses keep the decoder outputs close to those of the original decoder.
Lastly, WMAR proposes to use an image synchronization layer watermark to recover original image orientations, improving robustness against removal attempts by geometric transformations such as rotation.
The synchronization signal is used to estimate and invert these transformations before token reconstruction and watermark detection.
The follow-up method SyncSeal~\citep{fernandez2025syncseal} can be used as a replacement for the original synchronization layer.

\textbf{ClusterMark}{$_\mathrm{{(CVPR'26)}}$}~\citep{lukovnikov2025clustermark} also uses the preceding token to partition the vocabulary and studies token clustering as a possible means to improve robustness against removal attempts (since similar tokens are clustered and assigned the same color, switching between tokens within the same cluster does not destroy the watermark).
Specifically, the vocabulary is partitioned using $k$-means clustering with Euclidean distances between codebook embeddings.
The cluster identity of the preceding token, together with a secret key, determines the green and red clusters, and the logits of tokens belonging to green clusters are boosted during sampling.
Furthermore, a copy of the encoder is augmented with a classification head and trained with a perturbation-augmented dataset to further boost the reconstruction accuracy of original tokens or clusters from watermarked images.
The two variants either predict individual tokens and map them to clusters, or predict cluster identities directly, using cross-entropy supervision from unwatermarked generated images.
Detection then evaluates the recovered sequence for an unusually high proportion of green tokens.

\clearpage

\begin{figure}[t]
    \centering
    \includegraphics[width=1.0\columnwidth]{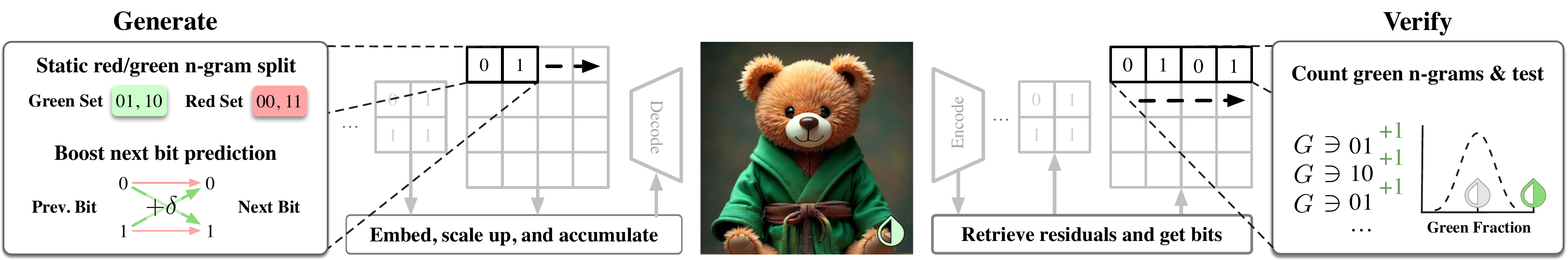}
    \caption{\textbf{Overview of BitMark}. Here, shown for the multi-scale generation model Infinity. \textit{Generation} (left): A fixed green set of n-grams (here, bigrams $G=\{01,10\}$) is used to bias sampling of bits on a given scale by~$\delta$.
    \textit{Verification} (right): Bits are recovered, the fraction of green n-grams is determined, and a statistical hypothesis test is performed to check if it is significantly above chance level, indicating the presence of a watermark.}
    \label{supp:fig:bitmark_overview}
\end{figure}

\subsection{Watermarking for Bitwise Autoregressive Models}
\label{sec:supp:scheme_details_bitwise}

\textbf{BitMark}{$_\mathrm{{(NeurIPS'25)}}$}~\citep{kerner2025BitMark} has been developed specifically for the Infinity~\citep{han2025infinity} generator.
Its bitwise design is motivated by the observation that individual bits are more consistently recovered than complete tokens when generated images are decoded and re-encoded.
Recall that during image generation, Infinity~\citep{han2025infinity} computes logits over bits at scales of increasing resolution, conditioning the parallel prediction of all bits in one scale on bits in all previously predicted scales.
With this in mind, BitMark follows an approach similar to KGW, where first the distributions over the bits in one scale are viewed as a sequence (unfolded over the channel dimension first).
Given the sequence of binary distributions, the watermark is embedded by biasing the sampling of a bit based on the $l$ previously sampled bits.
Concretely, a bias $\delta$ is added to the logit of the bit that completes a green-listed sequence, controlling the trade-off between watermark strength and image quality.
Given that the vocabulary size is only 2, there are only a few possible conditional partitionings of the vocabulary (e.g. $G_2 = \{01, 10\}$ for $l=1$), where each one is empirically evaluated and the best ones are used as the final method.
During detection, the image is re-encoded into bit representations across scales, and a $z$-test assesses whether the fraction of green-listed sequences significantly exceeds the assumed unwatermarked baseline of $0.5$.
BitMark was specifically proposed to enable service providers to re-identify their generated content and exclude it from subsequent training, thereby mitigating model collapse~\citep{kerner2025BitMark,shumailov2024modelcollapse,alemohammad2024selfconsuming}.
This goal is reinforced by its radioactive property: the watermark signal propagates to models trained on watermarked data, even transferring across architectures (e.g., to diffusion models).

\cref{supp:fig:bitmark_overview} shows a schematic overview of BitMark and~\cref{alg:supp:bitmark_detection} shows the detection procedure in detail.

\begin{figure}[H]
\centering
\scalebox{1.0}{%
\begin{minipage}{1.0\linewidth}
\begin{algorithm}[H]
\caption{Watermark Detection (Alg. 2 from BitMark~\citep{kerner2025BitMark})}
\label{alg:supp:bitmark_detection}
\begin{algorithmic}[1]
\Statex \textbf{Inputs:} raw image $x$, green list $G$, red list $R$, encoder $\encoder$, quantizer $\mathcal{Q}$
\Statex \textbf{Hyperparameters:} steps $K$ (number of resolutions), resolutions $(h_i, w_i)_{i=1}^K$, the number of tokens for resolution $i$ is $r_i$, number of latent channels $m$.

\State $e = \encoder(x)$
    \Comment{Encode image into latent features}
\State $C = 0$
\For{$i = 1, \ldots, K$}
    \State $u_i = \mathcal{Q}(\text{Interpolate}(e, h_i, w_i))$
        \Comment{Quantize at scale $i$}
    \State $u_i = (b_1, \ldots, b_{r_i \cdot m})$
        \Comment{Flatten bits, channels first}
    \State $C = \text{Count}((b_1, \ldots, b_{r_i \cdot m}), G)$
        \Comment{Count green-listed bit patterns}
    \State $z_i = \text{Lookup}(u_i)$
    \State $z_i = \text{Interpolate}(z_i, h_K, w_K)$
        \Comment{Resize to the finest resolution}
    \State $e = e - \phi_i(z_i)$
        \Comment{Remove this scale's reconstructed contribution}
\EndFor
\State \textbf{Return:} $\text{StatisticalTest}(\mathcal{H}_0, (C))$
    \Comment{Test for an excess of green patterns}
\end{algorithmic}
\end{algorithm}
\end{minipage}%
}
\end{figure}

\clearpage

\section{Full Experimental Settings}
\label{sec:supp:experimental_settings}
We provide full details on our experimental setup and implementation details, namely details on the watermarking schemes and generators (\cref{sec:supp:wm_schemes_and_generators}) and deployed attacks (\cref{supp:sec:evaluation:deployed_attacks}), 
For reference, Table~\ref{tab:supp:vqvae_overview} summarizes details of the VQ-VAEs used in our evaluation, while Table~\ref{tab:supp:model_combinations} lists the models used for each watermarking scheme, proxy model setting, and corresponding box setting, i.e. the level of access and knowledge needed for launching different attacks.

\subsection{Settings for the Watermarking Schemes and Generators}
\label{sec:supp:wm_schemes_and_generators}

\textbf{IndexMark~\citep{tong2025indexmark}.}
We used the pre-constructed token pairs, and experimented with the GPT-B (at $256\times256$ resolution) and GPT-L ($384 \times 384$).
100\% of the red tokens were replaced, which is the strongest watermarking setting.
The rest of the settings are set to default, as specified in their repository\footnote{\tiny{\url{https://github.com/maifoundations/IndexMark/commit/ef603a73c6c0e30651ea993e9c54a0ae97782761}}}. With each model, we generate 1,000 watermarked images (one per ImageNet~\citep{russakovsky2015imagenet} class).

\textbf{WMAR\citep{jovanovic2025wmar}.}
We used the default settings available in the official respository\footnote{\tiny{\url{https://github.com/facebookresearch/wmar/commit/d115650fd547add8444705f9dda2ae5e06f84675}}}. For each of the deployed models (Anole~\citep{chern2024anole} $512\times512$, Taming~\citep{esser2021taming} $256\times256$, and RAR-XL~\citep{yu2025rar} $256\times256$), the watermarking parameters are set to $\gamma=0.25$ (green token fraction) and $\delta=2.0$ (bias strength). Every model has the \emph{stratification strategy} enabled, which excludes dead tokens from green token assignment. For Taming, this leads to 971 alive tokens out of 16,384 tokens in total. For Anole, there are 57,344 alive tokens out of a total of 65,536 tokens. The RAR-XL VQ-VAE does not have dead tokens, so all of the 1,024 tokens in the vocabulary are alive. 
Furthermore, we use WMAR's pretrained finetunes for each model's VQ-VAE decoder (necessary for watermarked generation) and encoder (necessary for watermark verification) for improved \emph{reverse cycle consistency}. 
For RAR-XL experiments with WMAR, we use the optimal generator settings reported in RAR~\citep{yu2025rar} (rather than those in WMAR's codebase) since these led to better FIDs matching the RAR paper.
Finally, we do not enable WMAR's \emph{synchronization layer} for improved robustness against geometric transformations, to study WMAR's robustness to non-geometric attacks in isolation.
For each model, we generate 1,000 watermarked images, using 1,000 ImageNet classes (Taming, RAR-XL), or 1,000 random captions from the COCO dataset~\citep{lin2014coco} (Anole).

\textbf{ClusterMark~\citep{lukovnikov2025clustermark}.}
\label{sec:supp:ClusterMark_details}
We obtained the code from the authors and used the default settings for the image generators LlamaGen~\citep{sun2024llamagen} GPT-B ($256\times256$) and GPT-L $384\times384$, as well as RAR-XL~\citep{yu2025rar}.
We used the default watermarking settings, with green fraction $\gamma = 0.25$, watermark strength $\delta=5$ and $64$ clusters, while using their pretrained cluster predictor for watermark verification.
For all three models, we generate 1,000 watermarked images each, using 1,000 ImageNet classes.

\textbf{BitMark~\citep{kerner2025BitMark}.}
\label{sec:supp:BitMark_details}
For image generation, we used the $\infty$-2B model~\citep{han2025infinity} in the 1 million pixel ($1024\times1024$) setting.
We further used a watermarking strength of $\delta=2$, which is the moderate setting between the three settings $\delta\in\{1,2,3\}$ which have been reported yielding desirable trade-offs between image quality degradation and watermark robustness.
Otherwise, we used the default settings from the official respository\footnote{\tiny{\url{https://github.com/sprintml/BitMark/commit/b1706b128425f7a21996971a973b85943c9b5740}}}, including enabling watermark embedding and detection at every scale.
Watermarked images are generated using 1,000 random captions from the COCO dataset.

\subsection{Attack Dataset and Settings for Deployed Attacks}
\label{supp:sec:evaluation:deployed_attacks}
Removal attacks are performed on 1,000 watermarked images, while forgery attacks use 1,000 authentic cover images from the COCO dataset. \advoptname-Removal and -Forgery, as well as \bitadvoptname-Removal attacks are evaluated on 100 images, due to the large number of settings and higher runtime requirements for a full optimization run.
For \advoptname-Forgery, a single watermarked image from the corresponding watermarking scheme is used as a reference image.
Frequency Injection Forgery does not require reference images after initial tuning.
Since the evaluated watermarking schemes operate on images of different resolutions, the COCO dataset~\citep{lin2014coco} is filtered to include only images whose shortest edge is at least as large as the target resolution. For BitMark, images to with a size of at least 480 are accepted due to limited availability of larger-sized images in COCO. These images are then upscaled and center-cropped accordingly to match the required input size.
Attack setting are listed below.
    
    \textbf{VQ-Regen}: A detailed description is provided in~\cref{sec:supp:vqregen_full_explaination}.
    
    \textbf{\advoptname-Removal and -Forgery}: We run optimization through the encoder of a proxy model's VQ-VAE to obtain pre-quantization latents, i.e. the representation right before nearest neighbor matching to the closest tokens in the vocabulary. This is done for at most 300 steps with a budget of $|\pert|^\infty < \frac{8}{255}$ and verify the resulting attack instances every 10 optimization steps against the target verifier. The Adam optimizer is used, with a learning rate of 0.001 across all setups, with some higher learning rates for \advoptname-Forgery attacks, as reported in~\cref{tab:supp:model_combinations}. During experimentation, we observed that individual proxy encoders used by the attacker display different progressions in terms of attack success and quality degradation, most likely due to differences in latent sizes and value ranges of latent embeddings. Furthermore, individual attack instances may oscillate between detectable and non-detectable states with successive optimization steps, especially in black-box settings. This behaviour could also be addressed by tuning learning rates with a dedicated schedule for individual setups. However, due to the large amount of attacker and verifier model combinations, we opt to streamline our setup by reporting metrics for the step with the highest (removal) or lowest (forgery) p-value per attack instance within a run of 300 optimization steps, measured every 10 iteration. This requires query access to the target detector. We report fixed-step results without such access after 300 optimization steps in Section~\ref{sec:supp:fixed_step_results}.
    
    \textbf{\bitadvoptname-Removal}\bitmarkonly is tested with 100 steps, targeting the finest scales (9-12). Accuracy and quality metrics are reported for the first occurrence of a $p>0.01$, i.e. the first successful removal at the FPR=$1\%$ threshold. \cref{supp:alg:bitopt} and~\ref{alg:supp:bitmark_detection} formally describe the \bitadvoptname-Removal attack and the watermark verification procedure in BitMark for reference, respectively.

    \textbf{BitFlipper}~\citep{kerner2025BitMark}\bitmarkonly assumes full access to the exact VQ-VAE deployed with the watermark verifier, as well as knowledge of watermarking settings. We used the implementation from BitMark’s official repository with minor modifications, which were necessary to bring the code in working condition.
    We used a strength of $\phi=2.2$, which was reported as the best setting.

    \textbf{Next-Frame Prediction}~\citep{qiu2026nextframe}\bitmarkonly is tested with default parameters, including a 40-pixel shift and 10 inference steps. We use their official repository\footnote{\tiny{\url{https://github.com/1249748036/NFPA/commit/7f449f305506f95876fbfde0690395bea720269c}}} and evaluate it only on BitMark, which generates images at $1024\times1024$ resolution. On the smaller images generated by the other target models, the attack severely degrades visual quality, even when its parameters are adjusted, such as reducing the shift to 10 pixels. Hence, they the attack is omitted from the evaluation of these models.

    \textbf{Reprompt-Forgery}\bitmarkonly is tested with the \textit{Inappropriate Image Prompts (I2P)}\footnote{\tiny{\url{https://huggingface.co/datasets/AIML-TUDA/i2p}}} dataset, sorting the "inappropriate\_percentage" column in descending order, to generate harmful attack instances from watermarked reference images with the attacker model SD-v2.1, using the default 50 inversion and denoising steps and the DDIM scheduler. Reprompt attacks on token-based schemes did not show success. Hence, they are reported only for BitMark.
    
    \textbf{Frequency Injection Forgery}\bitmarkonly is reported for the three settings found during the intial tuning phase, with different trade-offs between image quality and attack success. Details are provided in~\cref{sec:supp:frequency_injection_full_details}.

\begin{figure}[t]
    \centering
    \subfloat[Perturbations]{%
        \includegraphics[width=0.65\linewidth]{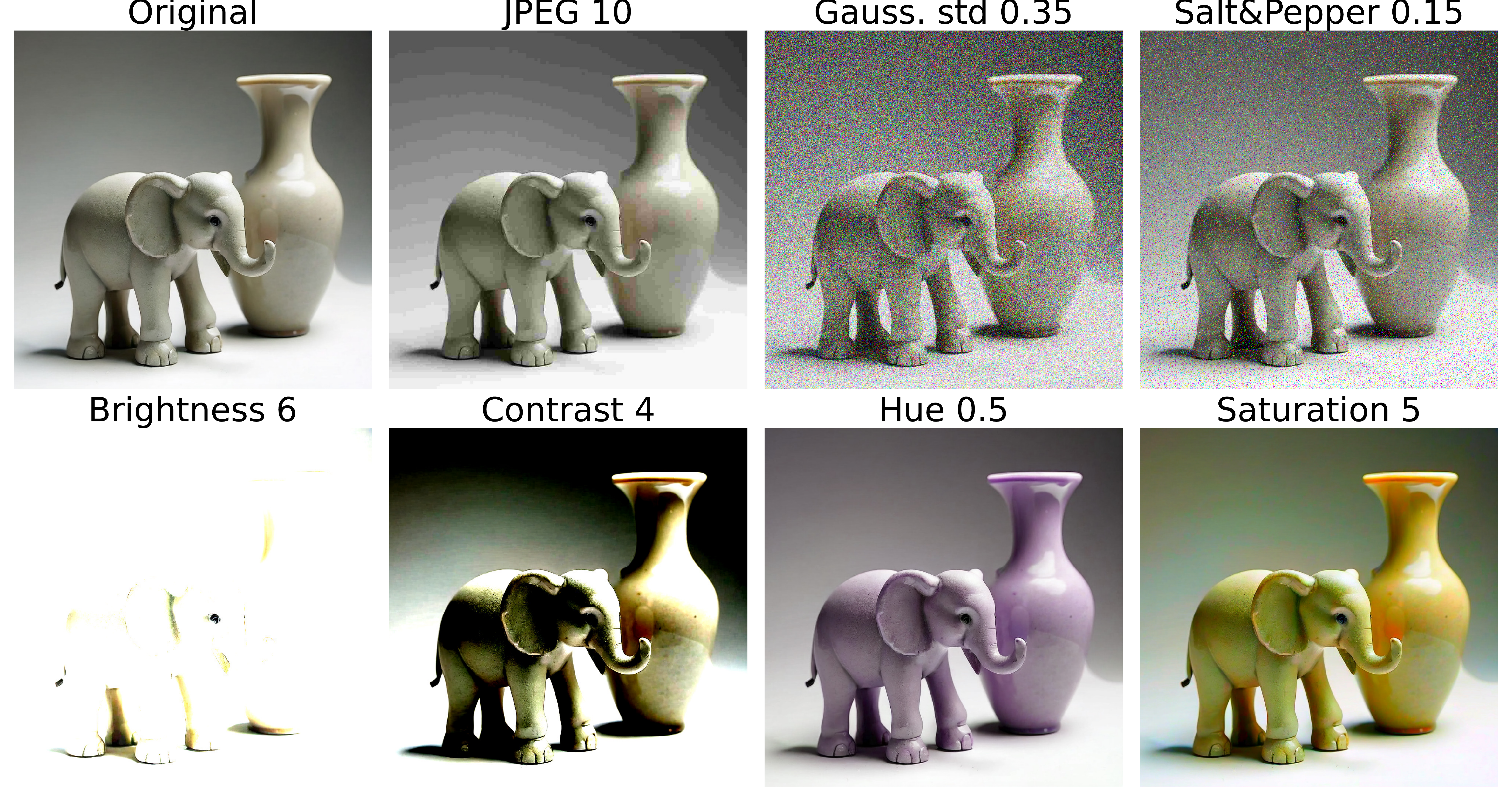}%
        \label{fig:supp:perturbation_examples_strong}%
    }
    \hfill
    \subfloat[Geometric transformations]{%
        \includegraphics[
            width=0.32\linewidth,
            trim={4mm 4mm 4mm 2mm},
            clip
        ]{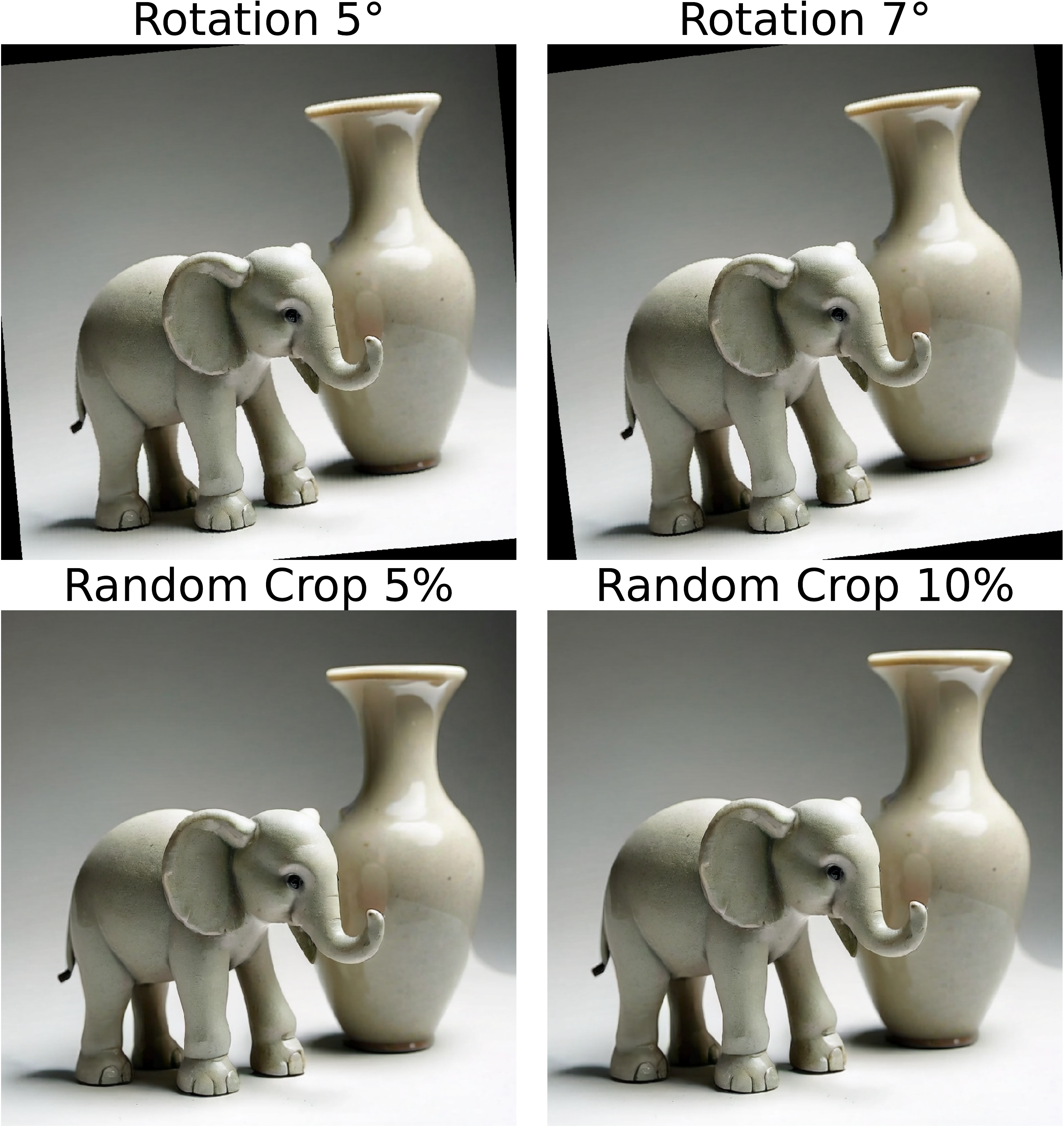}%
        \label{fig:supp:perturbation_examples_geometric}%
    }
    \caption{Visual examples of perturbations and geometric transformations.}
    \label{fig:supp:perturbation_examples}
\end{figure}

    \textbf{Perturbations and Geometric Transformations}:
    The different perturbations are applied individually to a watermarked target image and the average is reported.
    Visual examples and the parameters for each perturbation are provided in~\cref{fig:supp:perturbation_examples}.
    
    \textbf{Regen.}{\footnote{\tiny{\url{https://github.com/XuandongZhao/WatermarkAttacker/commit/2637dd2c3b84a3037dd3940b401090e8bcebe1f6}}}} and \textbf{Rinse} from \citep{ZhaZhaSu2024invisibleimagewatermarksprovably} are tested with the default setting of 60 steps and 3 rounds for Rinse, using Stable Diffusion v2.1~\citep{RomBlaLor2022stablediffusion}.
    
    \textbf{CtrlRegen+}\citep{liu2025ctrlregen} is tested with moderate settings (guidance of 2, strength $0.3$) using their official repository\footnote{\tiny{\url{https://github.com/yepengliu/CtrlRegen/commit/bd121448da96c38cf1254b28fe42652b6b72a8b1}}}, again using SD-v2.1.

\begin{table}[t]
\centering
\caption{
Detailed breakdown of settings used in each attack, excluding perturbations and geometric transformations. For each watermarking scheme and deployed verifier model, the attacker model is shown on the left side. Using an unrelated model on the attacker side (or no model, as in the Frequency Injection Forgery attack) is considered a \emph{black-box setting} and is indicated by \blackbox. Using a model of similar architecture is considered a \emph{grey-box} setting and is indicated by \greybox. For token-based watermarking schemes (IndexMark, WMAR, ClusterMark), this corresponds to using the default version of the deployed verifier model's encoder, without acces the finetuned weights. White-box settings require full access to the exact weights of the verifier's encoder and is indicated by \whitebox. In the case of BitMark, having additional knowledge of the green set $G$ is indicated by \whitebox{}$+$. For the \advoptname-Forgery attack on the Anole and Taming models deployed with WMAR, a higher learning rate (LR=0.01) was chosen. This was done in order to saturate the attack success for experiments with higher perturbation budgets such as $c=\frac{32}{255}$ (see~\cref{sec:pertubation_budget_effect}), since the default LR=0.001 would not show forgery success within 300 steps. This was done to show that forgery is possible for all verifier models, albeit at the cost of extreme quality degradation.
}
\resizebox{0.99\linewidth}{!}{%
  \input{tables/supp/full_setup}
}
\label{tab:supp:model_combinations}
\end{table}

\clearpage

\begin{table}[b]
\centering
\caption{Overview of the visual tokenizers used in each deployed target model. $|\vocabulary|$ and $d$ refer to the vocabulary size and the embedding dimension, respectively.}
\begin{tabular}{l @{\hspace{0.7em}} l @{\hspace{0.9em}} l @{\hspace{0.9em}} l @{\hspace{0.9em}} l @{\hspace{0.9em}} llll}
\toprule
 & \textbf{LlamaGen} & \textbf{Anole} & \textbf{Taming} & \textbf{RAR} & \multicolumn{4}{c}{\textbf{$\infty$-2B}} \\
\midrule
$|\vocabulary|$ & 16384 & 65536 & 16384 & 1024 & $2^{16}$ & $2^{24}$ & $2^{32}$ & $2^{64}$ \\
$d$ & 8 & 256 & 256 & 256 & 16 & 24 & 32 & 64 \\
\bottomrule
\end{tabular}
\label{tab:supp:vqvae_overview}
\end{table}

\begin{figure}[p]
\centering

\centering
\scalebox{1.0}{%
\begin{minipage}{1.0\linewidth}
\begin{algorithm}[H]
\caption{White-Box+ \bitadvoptname-Removal Attack for BitMark}
\label{supp:alg:bitopt}
\begin{algorithmic}[1]
\Require watermarked target image $x$, encoder $\mathcal{E}$, quantizer $\mathcal{Q}$,
         green list $G$, scale resolutions $(h_i, w_i)_{i=1}^{K}$,
         margin $\gamma$, perturbation budget $\epsilon$,
         step size $\alpha$, number of steps $T$
\State $\delta \gets \mathbf{0}$
\For{$t=1, \dots, T$}
    \State $z \gets \mathcal{E}(x + \delta)$
    \State $\hat{z} \gets z$        \Comment{$\hat{z}$ is overall residual}
    \For{$i = 1, \ldots, K$}
        \State $\tilde{e}_i \gets \text{Interpolate}(\hat{z},\, h_i,\, w_i)$
        \Comment{Unquantized residual}
        \State $u_i \gets \mathcal{Q}(\tilde{e}_i)$
        \Comment{Quantized residual}
        \State $\tilde{z}_i \gets \hat{z} - \text{Interpolate}(u_i, h_{K}, w_{K})$
        \Comment{$\tilde{z}_i$ is unquantized residual at scale $i$}
        \State $\hat{z} \gets \hat{z} - u_i$
        \State $(b^{(i)}_1, \ldots, b^{(i)}_{h_i \cdot w_i \cdot m}) \gets \mathbb{B}(u_i > 0)$
        \Comment{Extract bit sequence}
    \EndFor
    \State $\mathcal{T} \gets \emptyset$
    \For{$i=1, \ldots, K$}
        \For{$j \in \{2, \ldots, h_i \cdot w_i \cdot m - 1\}$}
            \If{$(b^{(i)}_{j-1},\, b^{(i)}_j,\, b^{(i)}_{j+1}) \in \{(0,1,0),\; (1,0,1)\}$}
                \State $\mathcal{T} \gets \mathcal{T} \cup \{(i, j)\}$
                \Comment{Flipping $b_j$ reduces green count by 2}
            \EndIf
        \EndFor
    \EndFor
    \State $\mathcal{L} \gets \displaystyle\sum_{(i,\,j)\;\in\;\mathcal{T}}
            |\tilde{e}_i[j] + \operatorname{sign}\!\bigl(\tilde{e}_i[j]\bigr) \cdot \gamma |$   \Comment{L1 Loss}
    \State $\delta \gets \delta - \alpha \cdot \nabla_{\delta}\,\mathcal{L}$
    \State $\delta \gets \operatorname{clip}(\delta,\; -\epsilon,\; \epsilon)$
    \Comment{Project back to $\ell_\infty$ ball}
\EndFor
\State \Return $x + \delta$
\end{algorithmic}
\end{algorithm}
\end{minipage}
}

\end{figure}

\clearpage

\begin{figure}[H]
    \centering
    \includegraphics[width=0.97\columnwidth]{figures/vqregen_progression.pdf}
    \caption{\textbf{Visual examples of the VQ-Regen attack. } The original watermarked image is generated using BitMark deployed with $\infty$-2B.
    }
    \label{fig:supp:vqregen_examples}
\end{figure}

\subsubsection{VQ-Regen Algorithm and Effect VQ-VAEs and Substitution Ranks}
\label{sec:supp:vqregen_full_explaination}
\cref{alg:supp:vqregen} provides a formal description of the VQ-Regen attack and examples of VQ-Regen applied to watermarked images (generated by BitMark) are provided in~\cref{fig:supp:vqregen_examples}.

From the figure, we observe that the choice of VAE and quantizer plays an important role in the VQ-Regen attack and this due to both the embedding dimension size $d$ and the vocabulary size $|\vocabulary|$:
1) The effect of the VAE's vocabulary size is straight forward: With fewer available tokens spanning the embedding space, the next closest token used in the VQ-Regen attack will generally be further away from the original token, resulting in more visual change.
For example, the RAR-XL VQ-VAE with a very small vocabulary size of $|\vocabulary| = 1024$ and a large embedding dimension $d=256$ shows lower image quality. 
2) The effect of the latent dimension $d$ is less obvious. We speculate that Euclidean distances become disproportionately larger with a growing number of embedding dimensions, resulting in larger distances between tokens and thus, more visual change.
For example, the LlamaGen VQ-VAE (first row) with a medium sized vocabulary of $|\vocabulary| = 16384$ and a very small embedding dimension of $d=8$ appears to have a much better reconstruction quality than the other VQ-VAEs with larger embedding dimensions of $d=256$ (bottom three rows).

\begin{figure}[H]
\centering
\scalebox{0.90}{%
\begin{minipage}{\linewidth}
\begin{algorithm}[H]
\caption{Vector-Quantized Regeneration Attack (VQ-Regen)}
\label{alg:supp:vqregen}

\begin{algorithmic}[1]
\Require Image $x$; encoder $\encoder$; decoder $\decoder$;
codebook $\codebook \in \mathbb{R}^{|\vocabulary| \times d}$;
subst. rank $k$ with $1 \le k \le |\vocabulary|$

\State $z \gets \encoder(x)$
    \Comment{$z \in \mathbb{R}^{d \times h \times w}$}
\State Initialize $t' \in \vocabulary^{h \times w}$

\For{$i \gets 1$ to $h$}
    \For{$j \gets 1$ to $w$}
        \State $u \gets z_{:,i,j} \in \mathbb{R}^{d}$
        \State $s \gets \Call{Argsort}{\big(\;\|\;u - \codebook[v,:]\;\|_2\;\big)_{v \in \vocabulary}}$
        \Comment{$s[1]$ is nearest, $s[k]$ is $k$-th nearest}
        \State $t'_{i,j} \gets s[k]$
    \EndFor
\EndFor

\State $z' \gets \codebook[t']$
    \Comment{lookup: $z'_{:,i,j} = \codebook[t'_{i,j},\,:]$}
\State $x' \gets \decoder(z')$
\State \Return $x'$
\end{algorithmic}
\end{algorithm}
\end{minipage}%
}
\end{figure}

\clearpage

\begin{figure}[H]
    \centering
    \includegraphics[width=1.0\linewidth]{figures/fft_concept_supp.pdf}
    \caption{\textbf{Frequency Injection Forgery for BitMark} Settings. We take an authentic cover image, and then compute its FFT. We inject spectral components with \emph{magnitude} $\alpha$ and a channel-wise random phase along the diagonals and spaced 32 pixels apart in both axes, directly overwriting the original coefficients. Finally, we apply inverse FFT to reconstruct the attacked image. \emph{Settings A} and \emph{B} are limited to the first four frequency bins along the diagonals (lower frequencies). \emph{Setting C} modifies all frequency bins along the diagonals. Setting A uses a low magnitude of $ln(\alpha)=7.75$, while settings B and C use a higher magnitude $ln(\alpha)=8.0$.}
    \label{fig:supp:fft_concept}
\end{figure}

\subsubsection{Frequency Injection Forgery for BitMark}
\label{sec:supp:frequency_injection_full_details}
As explained in~\cref{sec:fftforgery}, averaging many BitMarked images reveals a pixel pattern, appearing as a regular grid of high magnitude peaks in the frequency domain, which verifies against the BitMark verifier with a p-value of $3\times10^{-33}$. Based on this observation, we tried to recreate this effect by introducing a similar pattern via frequency injection into authentic cover images. For this, we ran a set of trials where magnitude peak locations are controlled by different step sizes $s\in\{16,32,64,128\}$ by either (i) arranging them in a full lattice with these spacings like in the FFT representation of the averaged image, or (ii) restricting the peaks to the main diagonals to minimize quality degradation. We also varied the magnitude across $ln(\alpha)\in\{7.5, 7.75, 8.0, 8.25, 8.5\}$ and restricted the peaks to the first $n$ occurrences closest to the center. We then evaluated the accuracy metrics (p-value produced by the BitMark verifier) and visual quality metrics (PSNR) on a separate set of 100 images for each setting, filtering out any result above $p=1e-5$ and sorting by PSNR in descending order. Finally, we chose three desirable trade-offs (settings A, B and C). Note that this procedure is not guaranteed to yield optimal results and there may be better trade-offs.

\indent
\cref{fig:supp:fft_concept} shows the settings for our Frequency Injection Forgery attack. The attack is performed by injecting magnitude peaks of strength $\alpha$ along the main diagonals in regular intervals (32 frequency bins in both axes) with random phases for each color channel's FFT. Setting A and B are limited to the first four lower frequency bins, while setting C includes all frequency bins along the main diagonals. Setting A uses a smaller magnitude of $ln(\alpha)=7.75$, while settings B and C use a magnitude of $ln(\alpha)=8.0$. With this, we achieve different trade-offs between the shift towards positive BitMark detection (lower p-values as determined by the BitMark verifier), and the quality degradation in terms of PSNR.

After the intial tuning phase, the injection procedure is then applied on a separate set of 1,000 images to produce the main results in Table~\ref{tab:bitmark}, as well as Table~\ref{tab:supp:full_results:BitMark:1M} in the Appendix.

\indent
We also experimented with frequency injection forgery on token-based methods, but this proved ineffective. Here, averaging a large set of watermarked images does not reveal obvious patterns in the frequency domain.

\clearpage

\begin{figure}[H]
    \centering
    \subfloat[Difference]{\includegraphics[width=0.48\linewidth]{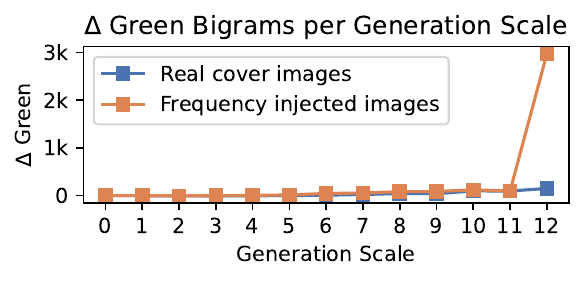}}
    \subfloat[Bits per scale]{\includegraphics[width=0.48\linewidth]{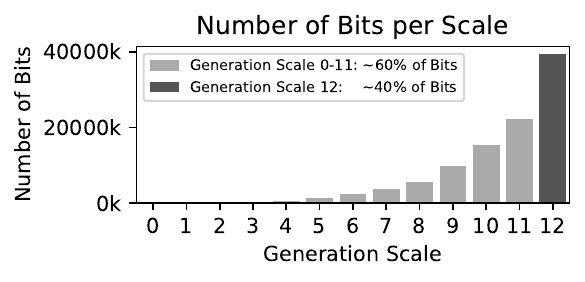}}

    \captionof{figure}{Total difference $\Delta$ between green ($G=\{01,10\}$) and red (($R=\{00,11\}$)) bigrams counts across different scales for 100 authentic cover images (blue) and 100 frequency injection forgery attack instances (setting A, orange) as verified by the BitMark watermarking schemes deployed with $\infty$-2B (\emph{left}). At the finest generation scale (scale 12), the attack introduces a strong surplus of green bigrams $G={01,10}$, causing the verifier to detect the presence of the watermark. The amount of bits across each generation scale of $\infty$-2B is shown to the \emph{right}. The scale most affected by the attack (scale 12) accounts for roughly 40\% of all embedded bits.}
    \label{fig:supp:fft_bits_per_scale}
\end{figure}

\indent
\cref{fig:supp:fft_bits_per_scale} shows the effect of frequency injection on the BitMark verifier: the periodic patterns trigger the $\infty$-2B multi-scale VAE to observe a high count of green bigrams $G=\{01,10\}$ in the last (finest) scale. \cref{fig:supp:fft_bits_per_scale} also shows a breakdown of the number of bits on each scale during generation: approximately 40\% of all bits are located in the last scale. While one could counter our attack by disabling watermark embedding and verification at the finest scale entirely, this would forfeit 40\% of the available embedding capacity, substantially reducing BitMark's robustness. Alternative strategies for detecting scale imbalances may exist, but evaluating their effectiveness in the context of other attacks like removal attempts is beyond the scope of this work.

\cref{alg:supp:fftforgery} provides a formal description of the injection procedure.

\begin{figure}[H]
\centering
\scalebox{1.0}{%
\begin{minipage}{0.9\linewidth}
\begin{algorithm}[H]

\begin{algorithmic}[1]
\Require image $x \in \mathbb{R}^{3 \times H \times W}$, magnitude $\alpha$, spacing $s=32$
\Ensure forged image $x'$

\State $X \gets \text{fftshift}(\mathcal{F}(x))$ \Comment{FFT per channel}
\State $M \gets$ diagonal mask with spacing $d$ in the \textbf{right half} of the spectrum
\Comment{points $(c_y \pm kd,\, c_x + kd)$ only (right-side quadrants)}

\For{channel $c \in \{1,2,3\}$}
\For{$(y,x)$ with $M[y,x]=1$}
\State sample $\phi \sim \mathcal{U}(0,2\pi)$
\State $A \gets \alpha e^{i\phi}$ \Comment{fixed magnitude, random phase}
\State $X_c[y,x] \gets X_c[y,x] + A$
\State $(y',x') \gets ((H-y)\bmod H,(W-x)\bmod W)$
\State $X_c[y',x'] \gets X_c[y',x'] + \overline{A}$
\Comment{write conjugate to opposite quadrant to enforce Hermitian symmetry}
\EndFor
\EndFor

\State $x' \gets \Re\!\big(\mathcal{F}^{-1}(\text{ifftshift}(X))\big)$
\Comment{real output guaranteed by Hermitian symmetry}

\State \Return $x'$
\end{algorithmic}

\caption{Frequency Injection in the FFT Domain}
\label{alg:supp:fftforgery}
\end{algorithm}
\end{minipage}
}

\end{figure}

\clearpage

\section{Effect of WMAR's Synchroniziation Layer}
\label{sec:supp:wmar_synchronization}

In all experiments, we evaluated WMAR with its proposed additional layer for geometric synchronization disabled. The synchronization layer is orthogonal to the autoregressive watermarking method itself, as it constitutes an additional post-hoc watermarking component. Its primary purpose is to reverse geometric transformations by re-aligning a transformed image with its original coordinate system. By ommiting this addition, we are able to compare the underlying autoregressive watermarking mechanism of WMAR against the other token-based schemes in isolation.

In fact, adding a post-hoc watermark can disturb the underlying AR watermarking schemes. This is shown in Table~\ref{tab:supp:wmar_synchronization_comparison}: For WMAR deployed with the Anole model, we compare mean TPR@FPR=1\% and median p-value for a deployment with and without the layer enabled. We use the default synchronization layer included in the original repository\footnote{\tiny{\url{https://github.com/facebookresearch/wmar/commit/d115650fd547add8444705f9dda2ae5e06f84675}}}.

We observe that, while WMAR's robustness against geometric transformation increases slightly from 0.13 to 0.48 TPR, its overall performance drops significantly. For example, TPR for the Regen. attack drops from 0.63 to 0.23.

\begin{table}[H]

\caption{
Comparison of WMAR deployed with Anole, with and without the synchronization layer enabled, under removal and forgery attacks. We report mean TPR at 1\% FPR with 95\% confidence intervals, as well as mean p-values.
}

\resizebox{\linewidth}{!}{%

\begin{tabular}{lll @{\hspace{1.2em}} r @{\hspace{0.7em}} S[table-format=1.1e3, exponent-product=\times,tight-spacing=true] @{\hspace{1.5em}} r @{\hspace{0.7em}} S[table-format=1.1e3, exponent-product=\times,tight-spacing=true]}
\toprule
\multicolumn{3}{l}{WMAR deployed with Anole} & \multicolumn{2}{c}{Without synchronization} & \multicolumn{2}{c}{With synchronization} \\
\cmidrule(lr){4-5} \cmidrule(lr){6-7}
 & & & \multicolumn{1}{c}{TPR} & \multicolumn{1}{c}{P-Value} & \multicolumn{1}{c}{TPR} & \multicolumn{1}{c}{P-Value} \\
\midrule
Watermarked & & & $0.98$ $[0.97,\,0.99]$ & 3.1e-47 & $0.97$ $[0.96,\,0.98]$ & 3.8e-38 \\
\midrule
\multicolumn{7}{c}{Removal} \\
\midrule
Geometric Transf. & & & $0.13$ $[0.12,\,0.14]$ & 1.9e-01 & $0.48$ $[0.46,\,0.50]$ & 1.6e-02 \\
Perturbations & & & $0.42$ $[0.41,\,0.43]$ & 4.2e-02 & $0.40$ $[0.39,\,0.41]$ & 5.9e-02 \\
\midrule
Regen. & & \blackbox & $0.63$ $[0.60,\,0.66]$ & 3.0e-03 & $0.23$ $[0.20,\,0.26]$ & 2.3e-01 \\
Rinse & & \blackbox & $0.04$ $[0.03,\,0.06]$ & 2.3e-01 & $0.01$ $[0.01,\,0.02]$ & 4.3e-01 \\
CtrlRegen{}$+$ & & \blackbox & $0.18$ $[0.16,\,0.21]$ & 8.8e-02 & $0.05$ $[0.04,\,0.07]$ & 3.4e-01 \\
\midrule
VQ-Regen\hspace{0.3cm} (aggregated) & & \blackbox & $0.21$ $[0.19,\,0.22]$ & 1.4e-01 & $0.06$ $[0.04,\,0.09]$ & 4.5e-01 \\
\phantom{VQ-Regen\hspace{0.3cm}} LlamaGen & & \blackbox & $0.53$ $[0.50,\,0.56]$ & 8.0e-03 & $0.17$ $[0.11,\,0.26]$ & 2.2e-01 \\
\phantom{VQ-Regen\hspace{0.3cm}} RAR-XL & & \blackbox & $0.01$ $[0.00,\,0.01]$ & 4.6e-01 & $0.00$ $[0.00,\,0.04]$ & 5.5e-01 \\
\phantom{VQ-Regen\hspace{0.3cm}} Taming & & \blackbox & $0.09$ $[0.07,\,0.11]$ & 1.4e-01 & $0.00$ $[0.00,\,0.04]$ & 4.3e-01 \\
\cmidrule(l{1.95cm}){1-7}
\phantom{VQ-Regen\hspace{0.3cm}} Anole & & \greybox & $0.52$ $[0.49,\,0.55]$ & 1.1e-02 & $0.16$ $[0.10,\,0.24]$ & 2.2e-01 \\
\phantom{VQ-Regen\hspace{0.3cm}} Anole & & \whitebox & $0.20$ $[0.18,\,0.23]$ & 6.9e-02 & $0.20$ $[0.18,\,0.23]$ & 6.9e-02 \\
\midrule
\advoptname-R (aggregated) & & \blackbox & $0.22$ $[0.17,\,0.27]$ & 3.9e-01 & $0.19$ $[0.15,\,0.24]$ & 6.0e-01 \\
\phantom{\advoptname-R} LlamaGen & & \blackbox & $0.03$ $[0.01,\,0.08]$ & 8.0e-01 & $0.04$ $[0.02,\,0.10]$ & 7.7e-01 \\
\phantom{\advoptname-R} RAR-XL & & \blackbox & $0.49$ $[0.39,\,0.59]$ & 1.9e-02 & $0.36$ $[0.27,\,0.46]$ & 1.9e-01 \\
\phantom{\advoptname-R} Taming & & \blackbox & $0.13$ $[0.08,\,0.21]$ & 3.9e-01 & $0.18$ $[0.12,\,0.27]$ & 6.0e-01 \\
\cmidrule(l{1.95cm}){1-7}
\phantom{\advoptname-R} Anole & & \greybox & $0.01$ $[0.00,\,0.05]$ & 9.3e-01 & $0.00$ $[0.00,\,0.04]$ & 9.1e-01 \\
\phantom{\advoptname-R} Anole & & \whitebox & $0.01$ $[0.00,\,0.05]$ & 9.1e-01 & $0.01$ $[0.00,\,0.05]$ & 9.1e-01 \\
\midrule
\multicolumn{7}{c}{Forgery} \\
\midrule
\advoptname-F (aggregated) & & \blackbox & $0.00$ $[0.00,\,0.01]$ & 3.5e-01 & $0.05$ $[0.03,\,0.08]$ & 1.3e-01 \\
\phantom{\advoptname-F} LlamaGen & & \blackbox & $0.00$ $[0.00,\,0.04]$ & 3.4e-01 & $0.03$ $[0.01,\,0.08]$ & 1.2e-01 \\
\phantom{\advoptname-F} RAR-XL & & \blackbox & $0.00$ $[0.00,\,0.04]$ & 3.4e-01 & $0.06$ $[0.03,\,0.12]$ & 1.3e-01 \\
\phantom{\advoptname-F} Taming & & \blackbox & $0.00$ $[0.00,\,0.04]$ & 3.6e-01 & $0.05$ $[0.02,\,0.11]$ & 1.3e-01 \\
\cmidrule(l{1.95cm}){1-7}
\phantom{\advoptname-F} Anole & & \greybox & $0.00$ $[0.00,\,0.04]$ & 4.5e-01 & $0.00$ $[0.00,\,0.04]$ & 1.7e-01 \\
\phantom{\advoptname-F} Anole & & \whitebox & $0.07$ $[0.03,\,0.14]$ & 3.0e-01 & $0.07$ $[0.03,\,0.14]$ & 8.7e-02 \\
\bottomrule
\label{tab:supp:wmar_synchronization_comparison}
\end{tabular}

}
\end{table}

\clearpage

\begin{figure}[H]
    \centering

    \def\cropLeft{0pt}
    \def\cropRight{0pt}
    \def\cropTop{0pt}
    \def\cropBottom{0pt}

    \def\subplotHeight{58pt}

    \edef\transferpanel{%
        \noexpand\includegraphics[
            trim={\cropLeft\space\cropBottom\space\cropRight\space\cropTop},
            clip,
            height=\noexpand\subplotHeight
        ]%
    }

    \resizebox{\linewidth}{!}{%
        \begin{tabular}{@{}c@{}}

            \transferpanel{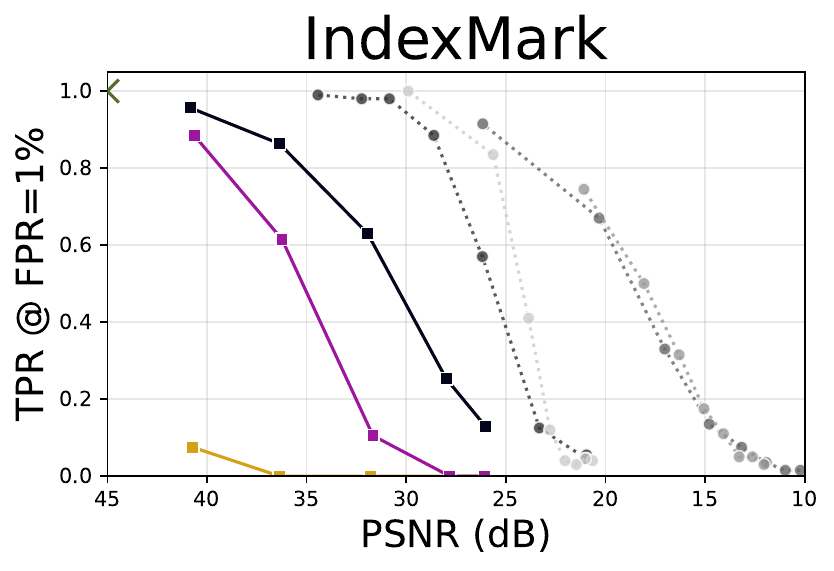}%
            \hspace{2pt}%
            \transferpanel{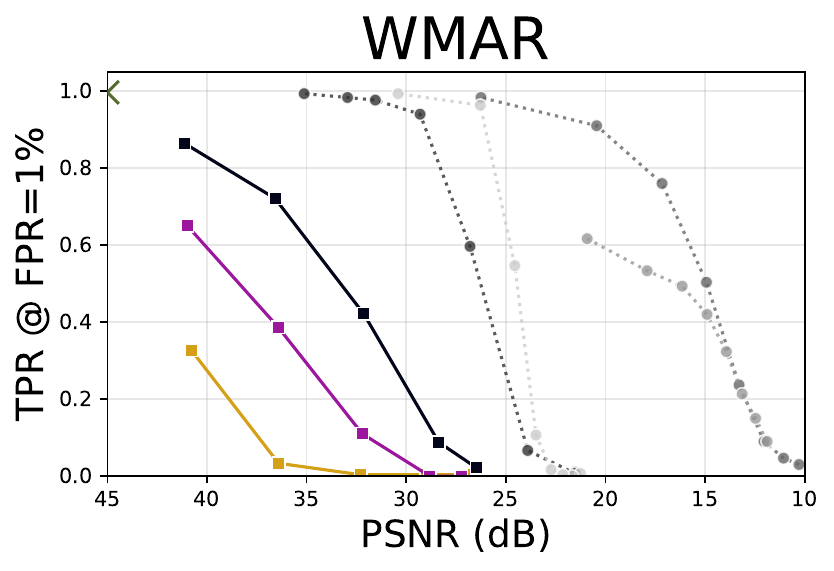}%
            \hspace{2pt}%
            \transferpanel{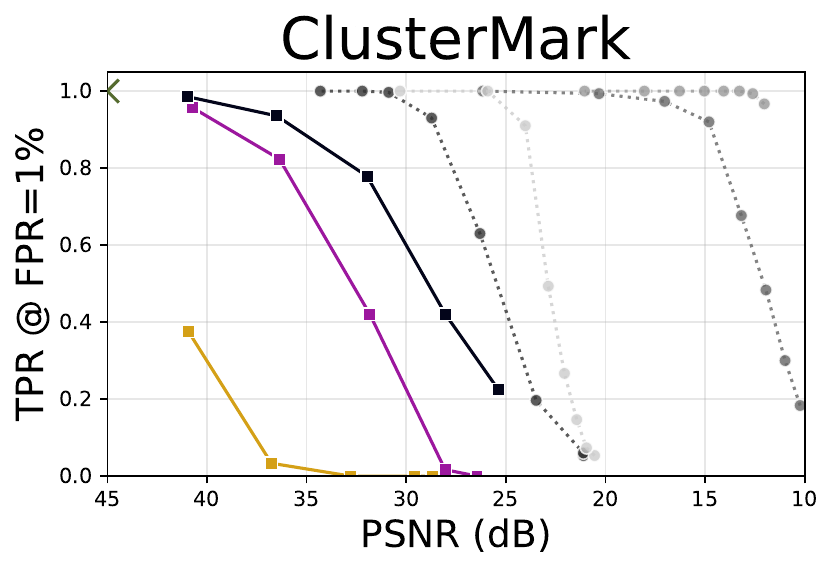}%
            \hspace{2pt}%
            \transferpanel{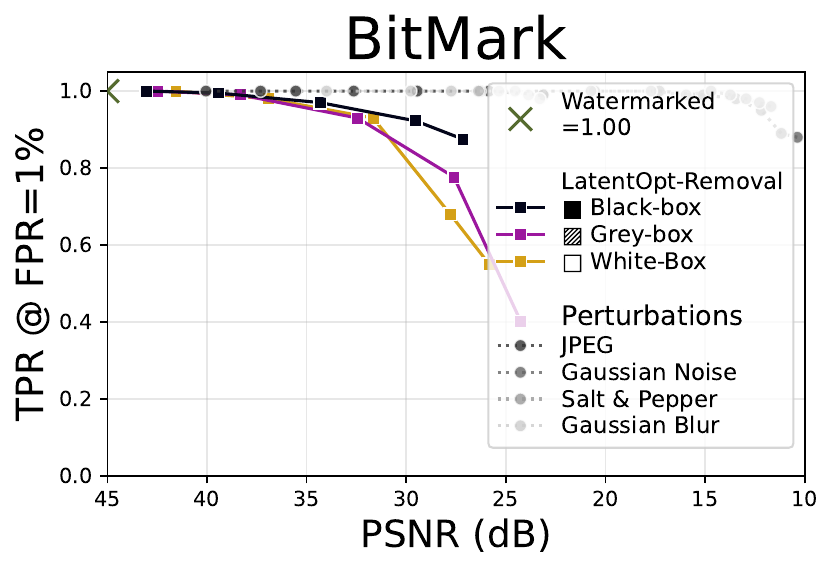}%
            \\[2pt]

            \def\subplotHeight{52pt}%
            \transferpanel{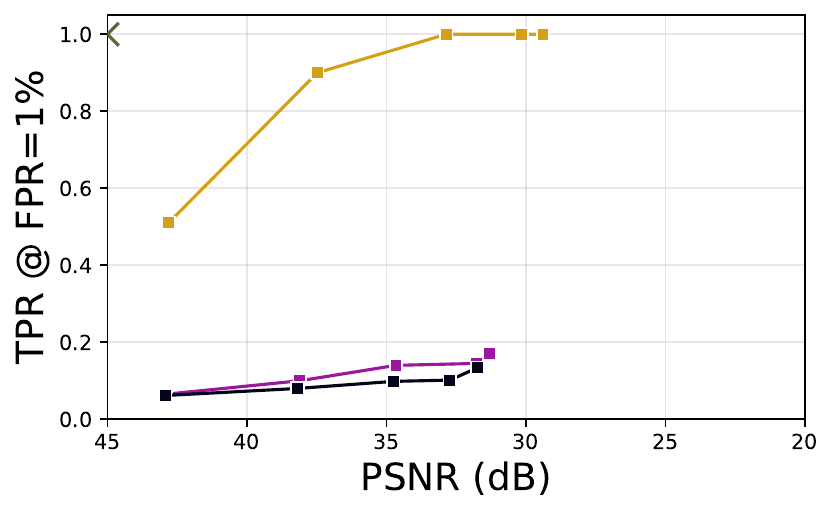}%
            \hspace{2pt}%
            \transferpanel{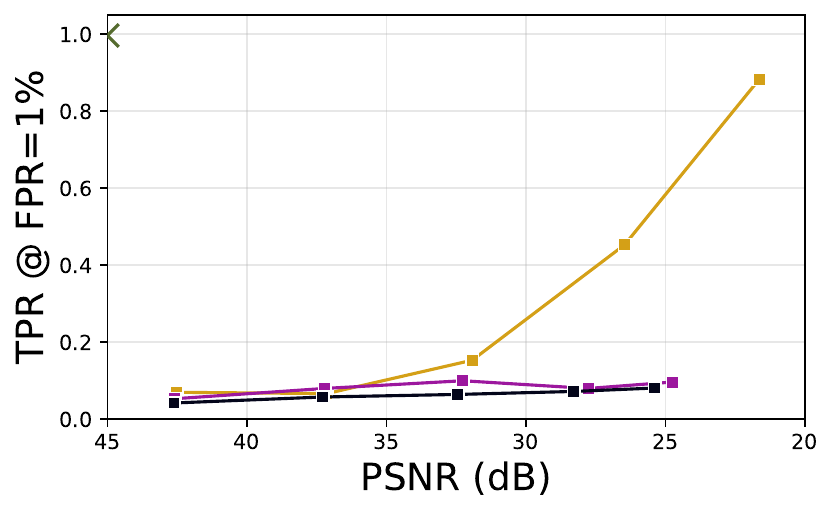}%
            \hspace{2pt}%
            \transferpanel{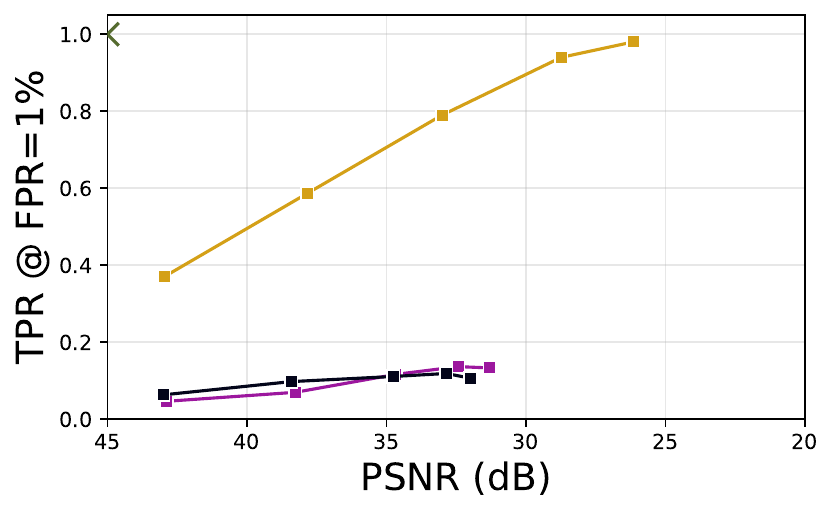}%
            \hspace{2pt}%
            \transferpanel{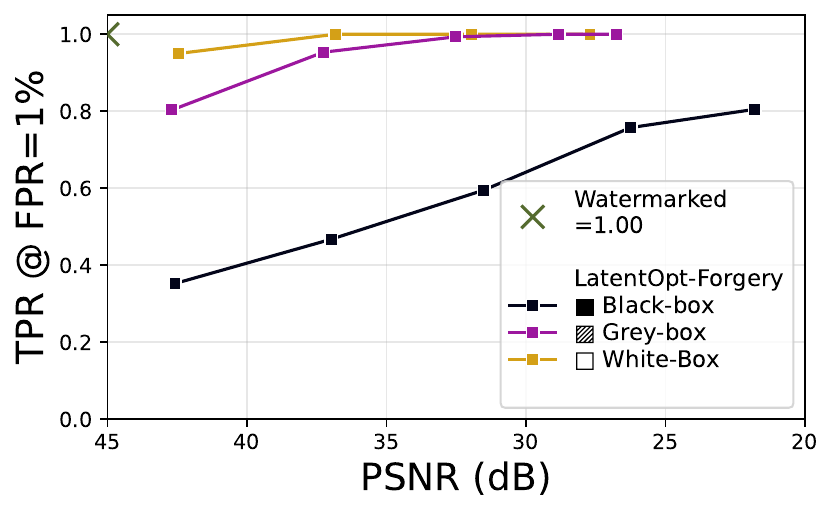}

        \end{tabular}%
    }

    \caption{
    \textbf{Transferability analysis}.
    \advoptname-Removal (top) and \advoptname-Forgery (bottom)
    results for budgets
    $c \in \{\frac{2}{255}, \frac{4}{255}, \frac{8}{255},
    \frac{16}{255}, \frac{32}{255}\}$.
    The top row also shows perturbation baselines in varying strengths.
    Levels of encoder access (black, grey, white) correlate with
    attack success, showing \emph{encoder transferability}.
    }
    \label{fig:encopt_combined_grid}
\end{figure}

\section{Transferability Between Encoders}
\label{sec:pertubation_budget_effect}
We study the trade-off between attack success (TPR@FPR=1\%) and quality degradation (PSNR) of our \advoptname attacks in different box-settings and with varying perturbation budgets $c$, and compare them to different naive image perturbations (noise, blur, etc.) of varying strength (see~\cref{fig:encopt_combined_grid}).

We observe that (1) using gradients informed by a VQ-VAE encoder in order to move an image's latents enables more efficient removal attacks than naive noise or compression perturbations, (2) both removal and forgery efficiency improves with closer alignment between the proxy encoder and the encoder deployed on the verifier's side, i.e., using an unrelated proxy encoder (black-box) is outperformed by using the same architecture (grey-box), which is in turn outperformed by using identical weights (white-box).
This indicates a \emph{transferability property} of adversarial optimizations for attacking autoregressive watermarks, similar to the properties shown in regard to attacks on semantic watermarking for diffusion models~\citep{Muller2025semanticforgery}.

\begin{table}[H]
    \centering
    \caption{Fixed-step \advoptname{}-Removal and -Forgery results for IndexMark, WMAR, ClusterMark, and BitMark after 300 optimization steps.
    \blackbox{}, \greybox{}, and \whitebox{} indicate black-box, grey-box, and white-box access, respectively.}
    \label{supp:tab:fixed_step_results}
    \label{tab:latent-opt-results}
    \resizebox{\textwidth}{!}{%
    \begin{tabular}{
        lll @{\hspace{1.2em}}
        c @{\hspace{0.7em}} S[table-format=1.1e3, exponent-product=\times,tight-spacing=true]cc
        @{\hspace{1.5em}}
        c @{\hspace{0.7em}} S[table-format=1.1e3, exponent-product=\times,tight-spacing=true]cc
        @{\hspace{1.5em}}
        c @{\hspace{0.7em}} S[table-format=1.1e3, exponent-product=\times,tight-spacing=true]cc
        @{\hspace{1.5em}}
        c @{\hspace{0.7em}} S[table-format=1.1e3, exponent-product=\times,tight-spacing=true]cc
    }
        \toprule
        & & & \multicolumn{4}{c}{IndexMark}
              & \multicolumn{4}{c}{WMAR}
              & \multicolumn{4}{c}{ClusterMark}
              & \multicolumn{4}{c}{BitMark} \\
        \cmidrule(l{-0.2em}r){4-7}
        \cmidrule(l{-0.2em}r){8-11}
        \cmidrule(l{-0.2em}r){12-15}
        \cmidrule(l{-0.2em}r){16-19}
        & & & \makecell{TPR} & \multicolumn{1}{c}{P-value} & PSNR & LPIPS
              & \makecell{TPR} & \multicolumn{1}{c}{P-value} & PSNR & LPIPS
              & \makecell{TPR} & \multicolumn{1}{c}{P-value} & PSNR & LPIPS
              & \makecell{TPR} & \multicolumn{1}{c}{P-value} & PSNR & LPIPS \\
        \midrule
        \multirow{3}{*}{\makecell[l]{\advoptname\\(Removal)}}
            & \multicolumn{2}{l}{\blackbox}
            & 0.78 & 1.5e-6 & 31.41 & 0.12
            & 0.55 & 2.7e-3 & 31.33 & 0.13
            & 0.90 & 9.9e-11 & 31.27 & 0.12
            & 0.99 & 7.9e-125 & 31.97 & 0.24 \\
            & \multicolumn{2}{l}{\greybox}
            & 0.39 & 3.5e-2 & 30.88 & 0.15
            & 0.23 & 2.4e-1 & 31.33 & 0.13
            & 0.68 & 1.8e-4 & 31.02 & 0.14
            & 0.97 & 3.5e-57 & 31.20 & 0.28 \\
            & \multicolumn{2}{l}{\whitebox}
            & 0.24 & 7.2e-1 & 31.36 & 0.04
            & 0.04 & 4.4e-1 & 30.92 & 0.14
            & 0.03 & 6.9e-1 & 31.64 & 0.11
            & 0.99 & 5.9e-28 & 30.68 & 0.27 \\
        \midrule
        \multirow{3}{*}{\makecell[l]{\advoptname\\(Forgery)}}
            & \multicolumn{2}{l}{\blackbox}
            & 0.01 & 5.5e-1 & 31.62 & 0.10
            & 0.01 & 6.2e-1 & 31.20 & 0.10
            & 0.02 & 5.1e-1 & 31.63 & 0.09
            & 0.20 & 1.6e-1 & 31.44 & 0.19 \\
            & \multicolumn{2}{l}{\greybox}
            & 0.00 & 5.0e-1 & 32.01 & 0.08
            & 0.01 & 6.2e-1 & 31.25 & 0.11
            & 0.01 & 5.7e-1 & 31.82 & 0.09
            & 0.99 & 3.9e-20 & 32.24 & 0.15 \\
            & \multicolumn{2}{l}{\whitebox}
            & 1.00 & 8.6e-78 & 32.84 & 0.03
            & 0.03 & 5.7e-1 & 30.97 & 0.10
            & 0.74 & 5.4e-10 & 32.45 & 0.07
            & 1.00 & 1.2e-220 & 31.39 & 0.17 \\
        \bottomrule
    \end{tabular}%
    }
\end{table}

\section{Fixed-Step Results for \advoptname Attacks}
\label{sec:supp:fixed_step_results}
We report alternative fixed-step results for \advoptname attacks after 300 optimization steps across all target watermarking schemes in Table~\ref{supp:tab:fixed_step_results} without query access to target detectors. Again, results are aggregated across all deployed target models for each individual scheme, as well proxy models used in the black-box and grey-box settings, in accordance with the reporting format used in~\ref{tab:token-based} and~\ref{tab:bitmark}.

\clearpage

\vspace{5cm}  %
\section{Full Experimental Results}
\label{sec:supp:full_results}

\begin{figure}[!t]
    \centering
    \includegraphics[width=0.95\columnwidth]{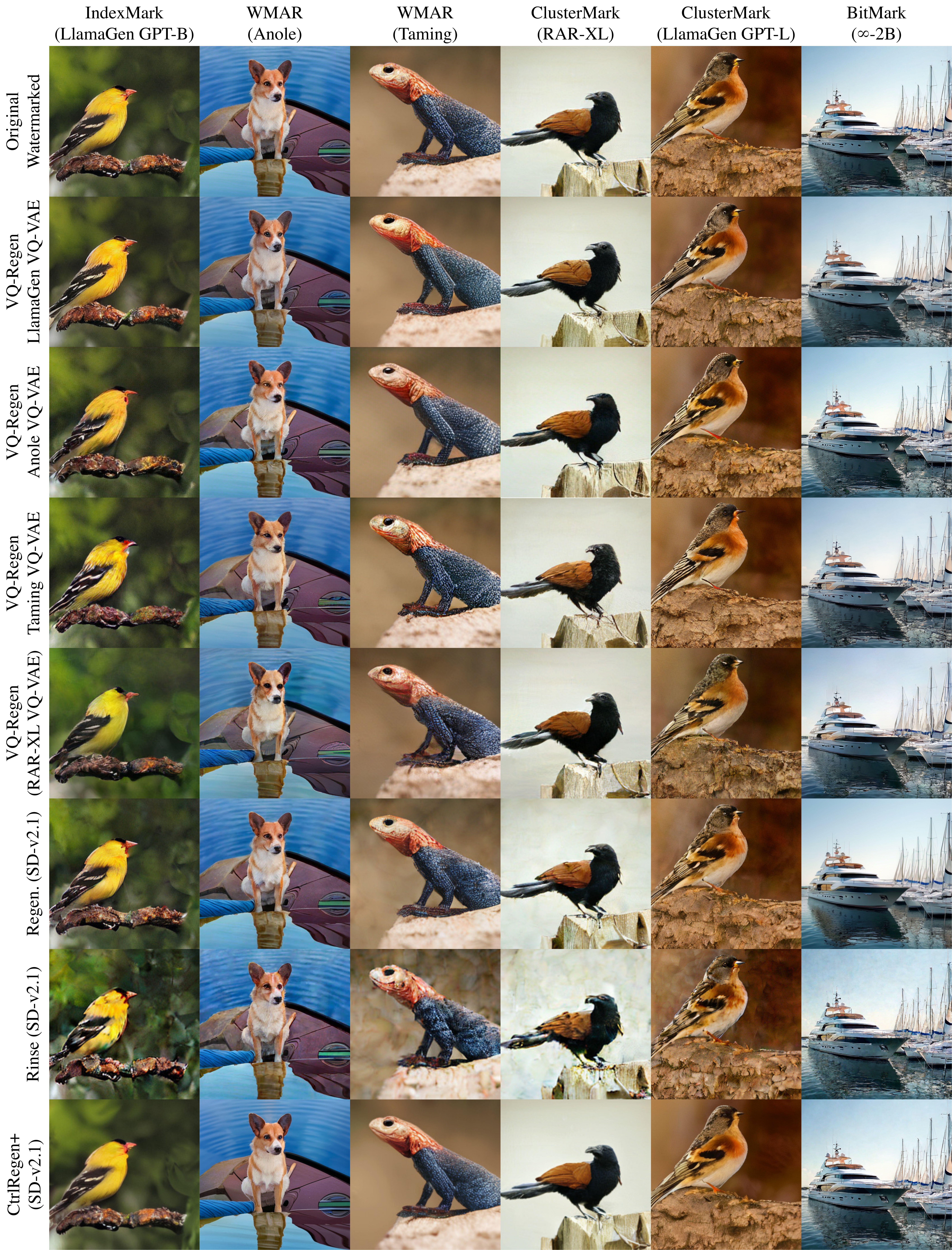}
    \caption{Visual examples of \textbf{VQ-Regen} and \textbf{diffusion regeneration-based} removal attacks.
    }
    \label{fig:supp:more_visual_examples_regen}
\end{figure}

\begin{figure}[!t]
    \centering
    \includegraphics[width=1.0\columnwidth]{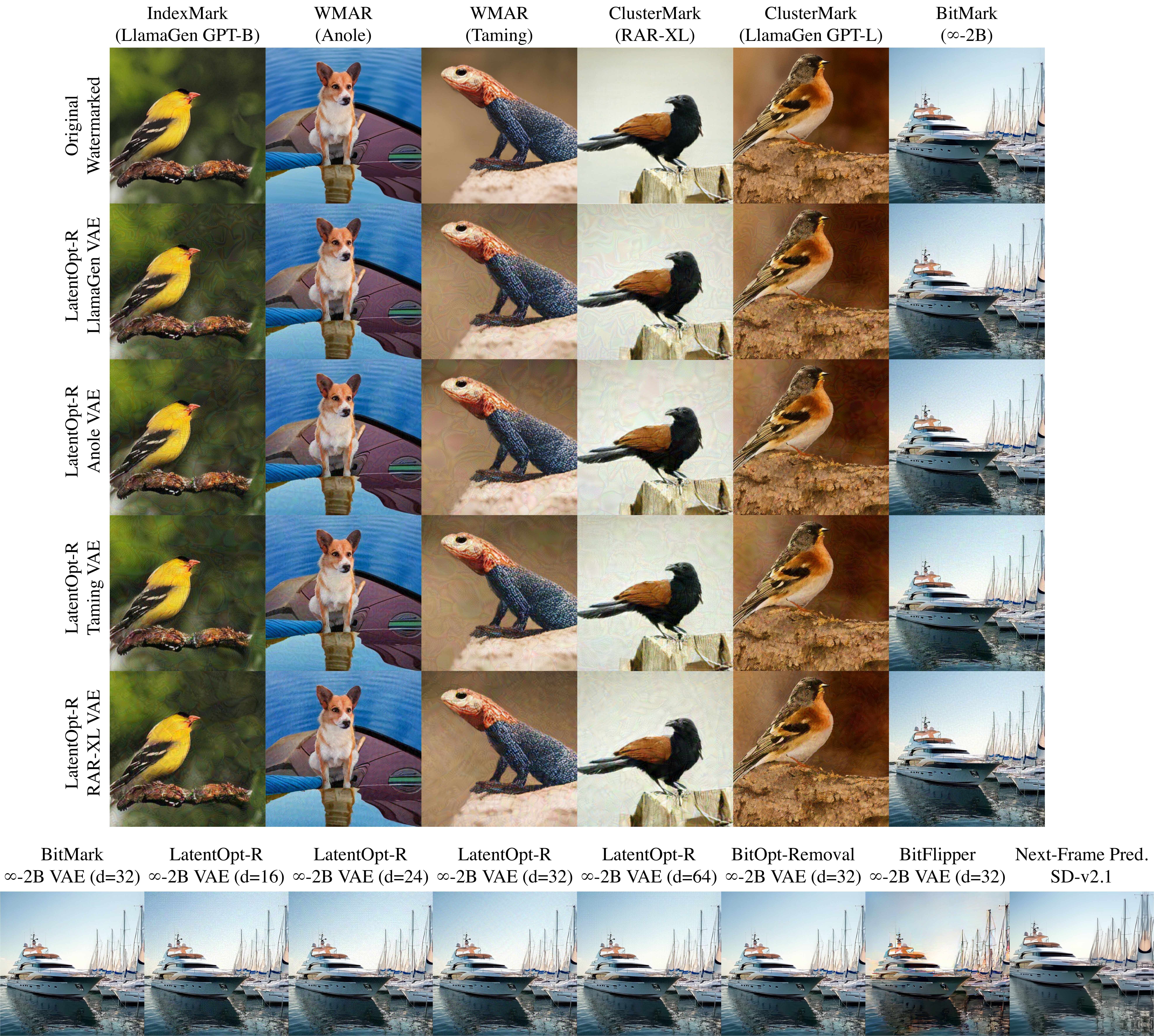}
    \caption{Visual examples of the \textbf{\advoptname-Removal} for all deployed schemes and models, as well as \textbf{\bitadvoptname-Removal}, \textbf{BitFlipper}, and \textbf{Next-Frame Prediction} attacks for BitMark.
    }
    \label{fig:supp:more_visual_examples_advopt}
\end{figure}

\begin{figure}[!t]
    \centering
    \includegraphics[width=1.0\columnwidth]{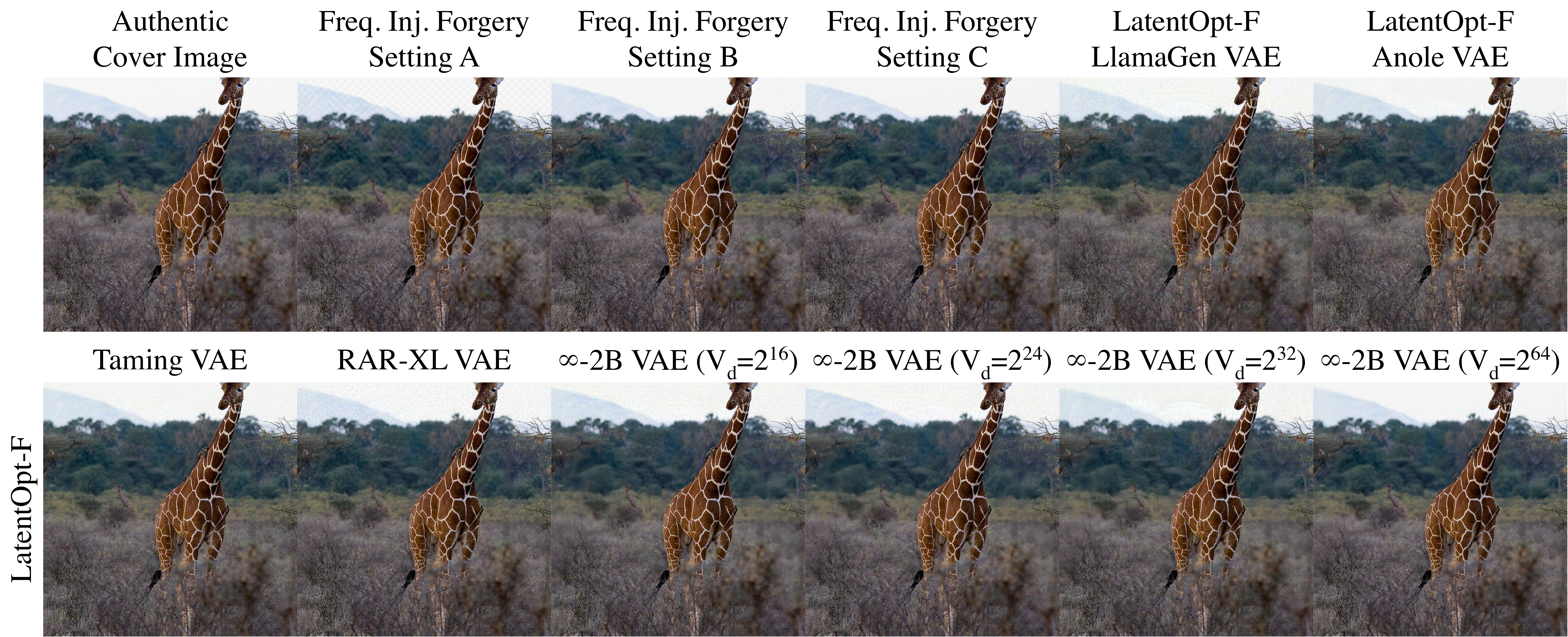}
    \caption{Visual examples of the \textbf{Frequency Injection Forgery} and \textbf{\advoptname-Forgery} at step 300) attacks on BitMark.
    }
    \label{fig:supp:more_visual_examples_forgery}
\end{figure}

\cref{fig:supp:more_visual_examples_regen,fig:supp:reprompt_example,fig:supp:more_visual_examples_advopt,fig:supp:more_visual_examples_forgery} show more visual examples for every attack.
Tables~\ref{tab:supp:full_results:IndexMark:GPT-B} to~\ref{tab:supp:full_results:BitMark:1M} show a detailed breakdown of all experimental results for all verifiers, their deployed models, and every attack setting, including individual proxy models used as mean TPR@FPR=1\%,
median p-value, as well as mean with standard deviation for both PSNR and LPIPS.
We also report confidence intervals (95\% Wilson score CIs for the binary TPR measure and student-t CIs for PSNR and LPIPS).
Again, removal attack aim for lower TPR, z-score, and higher p-value, while forgery attacks aim for the opposite. All attacks aim for PSNR$\uparrow$ and LPIPS$\downarrow$.

\paragraph{Experiment Runtime and Compute Resources.}
\label{sec:supp:compute}
All experiments were run over the course of two months on a server with 2 $\times$ 32-core AMD EPYC 7452, $\times$ 4 NVIDIA A40 GPUs (48 GB of VRAM), running Ubuntu 20.04.6 LTS. This includes generating 1,000 images for each deployed target model, for every target watermarking scheme, with the exception of BitMark for Infinity, where we generated 5,000 images for averaging the watermarking signal. Another 1,000 unwatermarked images were generated with the Infinity model to obtain different ratios of watermarked and unwatermarked images for fine-tuning models in the radioactivity experiments. With each of the five radioactive models, another 1,000 images were then generated. For all these settings, this amounts to roughly 20,000 generated images.

We provide average ranges runtimes of our attacks. In general, reproducing LatentOpt-Removal and Forgery with the chosen setting of 300 steps across all 96 attack settings for 100 attack examples is expensive and made up the majority of compute used.

\emph{LatentOpt-Removal and Forgery} with 300 steps takes between \textasciitilde4.3 seconds on average per sample in a batch for small images on the lower end (e.g. using the LlamaGen encoder with a batch size of 16 to attack images of size $256\times256$ generated by ClusterMark deployed with LlamaGen GPT-B), and \textasciitilde125 seconds on average per sample in a batch for large images on the higher end (e.g. using the Infinity encoder with a batch size of 2 to attack images of size $1024\times1024$ generated by Infinty). Note that while we ran the attack for 300 steps, runtime can in practice be lowered by early stopping.

\emph{BitOpt-Removal} with 100 steps takes between \textasciitilde68.7 seconds on average per sample with batch size 1. Again, the runtime can in practice be lowered by early stopping.

\emph{VQ-Regen} takes between \textasciitilde0.017 seconds on average per sample in a batch for small images on the lower end (e.g. using the LlamaGen encoder with a batch size of 32 to attack images of size $256\times256$ generated by ClusterMark deployed with LlamaGen GPT-B), and \textasciitilde0.34 seconds on average per sample in a batch for large images on the higher end (e.g. using the LlamaGen encoder with a batch size of 4 to attack images of size $1024\times1024$ generated by Infinty).

\emph{Next-Frame Predicction} takes \textasciitilde11.8 seconds on average per sample in a batch for images generated with Bitmark and Infinity of size $1024\times1024$ and batch size 1.

\emph{Reprompt-Forgery} takes \textasciitilde73.5 seconds on average per sample in a batch for images generated with Bitmark and Infinity of size $1024\times1024$ and batch size 2.

\emph{Frequency Injection Forgery} takes \textasciitilde0.17 seconds on average per sample in a batch for images generated with Bitmark and Infinity of size $1024\times1024$ and batch size 8.

\emph{Diffusion regeneration (Regen.)} takes between \textasciitilde0.09 seconds on average per sample in a batch for small images on the lower end (e.g. using the LlamaGen encoder to attack images $256\times256$ generated by ClusterMark deployed with LlamaGen GPT-B with a batch size of 32), and \textasciitilde0.92 seconds on average per sample in a batch for large images on the higher end (attacking images of size $1024\times1024$ generated by Infinty with a batch size of 4).

\emph{Rinse} takes between \textasciitilde0.5 seconds on average per sample in a batch for small images on the lower end (e.g. using the LlamaGen encoder to attack images $256\times256$ generated by ClusterMark deployed with LlamaGen GPT-B with a batch size of 32), and \textasciitilde6.32 seconds on average per sample in a batch for large images on the higher end (attacking images of size $1024\times1024$ generated by Infinty with a batch size of 4).

\emph{CtrlRegen} takes between 1 to 2 seconds on average per sample in a batch, depending on the size of attacked images.

\begin{figure}[!t]
    \centering
    \includegraphics[width=0.8\columnwidth]{figures/reprompt.pdf}
    \caption{Visual example the \textbf{Reprompt-Forgery} attack. While it can regenerate a watermarked image into a potentially harmful generated image, it does not allow for transferring the watermark into an authentic cover image.
    }
    \label{fig:supp:reprompt_example}
\end{figure}

\clearpage

\clearpage

\newlength{\tokentablewidth}
\setlength{\tokentablewidth}{1.0\linewidth}

\begin{table}[t]
\centering
\caption{
Full experimental results (\textbf{IndexMark} deployed w. \textbf{LlamaGen GPT-B)}
}
\resizebox{\tokentablewidth}{!}{%
  \input{tables/supp/full_results/IndexMark_GPT-B}
}
\label{tab:supp:full_results:IndexMark:GPT-B}
\end{table}

\begin{table}[t]
\centering
\caption{
Full experimental results (\textbf{IndexMark} deployed w. \textbf{LlamaGen GPT-L)}
}
\resizebox{\tokentablewidth}{!}{%
  \input{tables/supp/full_results/IndexMark_GPT-L}
}
\label{tab:supp:full_results:IndexMark:GPT-L}
\end{table}

\begin{table}[t]
\centering
\caption{
Full experimental results (\textbf{WMAR} deployed w. \textbf{Anole)}
}
\resizebox{\tokentablewidth}{!}{%
  \input{tables/supp/full_results/WMAR_chameleon}
}
\label{tab:supp:full_results:WMAR:Anole}
\end{table}

\begin{table}[t]
\centering
\caption{
Full experimental results (\textbf{WMAR} deployed w. \textbf{Taming)}
}
\resizebox{\tokentablewidth}{!}{%
  \input{tables/supp/full_results/WMAR_taming}
}
\label{tab:supp:full_results:WMAR:Taming}
\end{table}

\begin{table}[t]
\centering
\caption{
Full experimental results (\textbf{WMAR} deployed w. \textbf{RAR-XL)}
}
\resizebox{\tokentablewidth}{!}{%
  \input{tables/supp/full_results/WMAR_rar_xl}
}
\label{tab:supp:full_results:WMAR:RAR}
\end{table}

\begin{table}[t]
\centering
\caption{
Full experimental results (\textbf{ClusterMark} deployed w. \textbf{LlamaGen GPT-B)}
}
\resizebox{\tokentablewidth}{!}{%
  \input{tables/supp/full_results/arwm_GPT-B}
}
\label{tab:supp:full_results:ClusterMark:GPT-B}
\end{table}

\begin{table}[t]
\centering
\caption{
Full experimental results (\textbf{ClusterMark} deployed w. \textbf{LlamaGen GPT-L)}
}
\resizebox{\tokentablewidth}{!}{%
  \input{tables/supp/full_results/arwm_GPT-L}
}
\label{tab:supp:full_results:ClusterMark:GPT-L}
\end{table}

\begin{table}[t]
\centering
\caption{
Full experimental results (\textbf{ClusterMark} w. \textbf{RAR-XL)}
}
\resizebox{\tokentablewidth}{!}{%
  \input{tables/supp/full_results/arwm_rar_xl}
}
\label{tab:supp:full_results:ClusterMark:RAR}
\end{table}

\begin{table}[t]
\centering
\caption{
Full experimental results (\textbf{BitMark} deployed w. \textbf{$\infty$-2B ($d=32$))}
}
\resizebox{1.0\linewidth}{!}{%
  \input{tables/supp/full_results/bitmark_1M}
}
\label{tab:supp:full_results:BitMark:1M}
\end{table}

\clearpage
\input{checklist.tex}

\end{document}

%% file: custom_commands.tex
\newcommand{\encoder}{\ensuremath{\mathcal{E}}\xspace}  %
\newcommand{\decoder}{\ensuremath{D}\xspace}  %

\newcommand{\armodel}{\ensuremath{p_{\theta}}\xspace} 
\newcommand{\codebook}{\ensuremath{\mathcal{C}}\xspace}  
\newcommand{\vocabulary}{\ensuremath{\mathcal{V}}\xspace} 

\newcommand{\wmstrength}{\ensuremath{\delta}\xspace}

\newcommand{\softmax}{\ensuremath{\text{softmax}}\xspace}
\newcommand{\argmax}{\ensuremath{\text{argmax}}\xspace}
\newcommand{\argmin}{\ensuremath{\text{argmin}}\xspace}

\newcommand{\advoptname}{LatentOpt\xspace}
\newcommand{\bitadvoptname}{BitOpt\xspace}

\newcommand{\bitmarkonly}{$^{(\text{BitMark-only})}$\xspace}
\newcommand{\blackbox}{\ensuremath{\textcolor{black}{\blacksquare}}\xspace}
\newcommand{\greybox}{\ensuremath{\textcolor{gray}{\blacksquare}}\xspace}
\newcommand{\whitebox}{\ensuremath{\square}\xspace}

\definecolor{mygray}{HTML}{999999}
\definecolor{myred}{HTML}{FB6A4A}
\definecolor{myblue}{HTML}{3182BD}

\newcommand{\pert}{\ensuremath{\Delta x}\xspace}

\newcommand{\R}{\ensuremath{\mathbb{R}}\xspace}

%% file: tables/tokenbased_main.tex
\begin{tabular}{lll @{\hspace{1.2em}} c @{\hspace{0.7em}} S[table-format=1.1e3, exponent-product=\times,tight-spacing=true]cc @{\hspace{1.5em}} c @{\hspace{0.7em}} S[table-format=1.1e3, exponent-product=\times,tight-spacing=true]cc @{\hspace{1.5em}} c @{\hspace{0.7em}} S[table-format=1.1e3, exponent-product=\times,tight-spacing=true]cc}
\toprule
 &  &  & \multicolumn{4}{c}{IndexMark} & \multicolumn{4}{c}{WMAR} & \multicolumn{4}{c}{ClusterMark} \\
\cmidrule(l{-0.2em}r){4-7} \cmidrule(l{-0.2em}r){8-11} \cmidrule(l{-0.2em}r){12-15}
 &  &  & \makecell{TPR} & \multicolumn{1}{c}{P-value} & PSNR & LPIPS & \makecell{TPR} & \multicolumn{1}{c}{P-value} & PSNR & LPIPS & \makecell{TPR} & \multicolumn{1}{c}{P-value} & PSNR & LPIPS \\
\midrule
\multicolumn{3}{l}{WM} & 1.00 & 8.6e-78 & - & - & 0.99 & 1.1e-42 & - & - & 1.00 & 3.4e-81 & - & - \\
\midrule
\vspace{-1.3em}\\
\multicolumn{15}{c}{Removal} \\
\vspace{-1.3em}\\
\midrule
\multicolumn{3}{l}{Geom.} & 0.32 & 7.5e-2 & 16.16 & 0.18 & 0.30 & 9.7e-2 & 16.44 & 0.17 & 0.56 & 4.3e-3 & 16.26 & 0.17 \\
 \multicolumn{3}{l}{Perturb.} & 0.52 & 6.9e-3 & 17.56 & 0.43 & 0.45 & 3.1e-2 & 17.60 & 0.42 & 0.80 & 4.9e-21 & 17.56 & 0.43 \\
\midrule
Regen. & \multicolumn{2}{l}{\blackbox} & 0.52 & 1.0e-2 & 23.63 & 0.11 & 0.34 & 5.7e-2 & 24.30 & 0.11 & 0.85 & 9.9e-6 & 23.51 & 0.12 \\
Rinse & \multicolumn{2}{l}{\blackbox} & 0.03 & 3.8e-1 & 20.63 & 0.28 & 0.02 & 4.0e-1 & 21.26 & 0.28 & 0.26 & 6.2e-2 & 20.33 & 0.30 \\
CtrlRegen+ & \multicolumn{2}{l}{\blackbox} & 0.35 & 3.7e-2 & 24.14 & 0.13 & 0.26 & 7.4e-2 & 24.61 & 0.10 & 0.81 & 3.1e-5 & 24.26 & 0.12 \\
\midrule
\multirow[t]{6}{*}{VQ-Regen} & \multicolumn{2}{l}{\blackbox} & 0.03 & 4.2e-1 & 20.32 & 0.15 & 0.12 & 2.3e-1 & 21.59 & 0.12 & 0.39 & 2.5e-2 & 20.49 & 0.15 \\
 & \multicolumn{2}{l}{\greybox}  & 0.01 & 7.9e-1 & 22.30 & 0.10 & 0.19 & 2.4e-1 & 22.13 & 0.11 & 0.79 & 3.2e-5 & 21.12 & 0.12 \\
 & \multicolumn{2}{l}{\whitebox}  & 0.00 & 1.0e0 & 21.81 & 0.12 & 0.08 & 3.3e-1 & 21.76 & 0.13 & 0.82 & 1.8e-5 & 17.54 & 0.22 \\
\midrule
\multirow[t]{3}{*}{\parbox{1.5cm}{\vspace{1em}\advoptname \\ (Removal)}} & \multicolumn{2}{l}{\blackbox}  & 0.63 & 1.8e-4 & 31.94 & 0.12 & 0.42 & 5.2e-2 & 32.13 & 0.13 & 0.78 & 4.1e-7 & 31.97 & 0.12 \\
 & \multicolumn{2}{l}{\greybox} & 0.10 & 6.2e-1 & 31.67 & 0.15 & 0.11 & 8.4e-1 & 32.21 & 0.13 & 0.42 & 3.0e-2 & 31.82 & 0.14 \\
 & \multicolumn{2}{l}{\whitebox} & 0.00 & 1.0e0 & 31.81 & 0.04 & 0.00 & 9.4e-1 & 32.28 & 0.13 & 0.00 & 9.8e-1 & 32.79 & 0.10 \\
\midrule
\vspace{-1.3em}\\
\multicolumn{15}{c}{Forgery} \\
\vspace{-1.3em}\\
\midrule
\multirow[t]{3}{*}{\parbox{1.5cm}{\vspace{1em}\advoptname \\ (Forgery)}} & \multicolumn{2}{l}{\blackbox}  & 0.10 & 9.8e-2 & 34.76 & 0.07 & 0.06 & 1.6e-1 & 32.44 & 0.10 & 0.11 & 1.0e-1 & 34.75 & 0.06 \\
 & \multicolumn{2}{l}{\greybox}  & 0.14 & 9.5e-2 & 34.65 & 0.06 & 0.10 & 1.6e-1 & 32.26 & 0.10 & 0.12 & 1.1e-1 & 34.65 & 0.06 \\
 & \multicolumn{2}{l}{\whitebox}  & 1.00 & 8.6e-78 & 32.85 & 0.03 & 0.15 & 1.0e-1 & 31.92 & 0.10 & 0.79 & 3.6e-11 & 33.00 & 0.06 \\
\bottomrule
\end{tabular}

%% file: tables/bitmark_main_removal.tex
\begin{tabular}[t]{lllc
S[table-format=1.1e4,tight-spacing=true]
cc}
\toprule
 &  &  & \makecell{TPR} & \multicolumn{1}{c}{P-value}  & PSNR & LPIPS \\
\midrule
Geom. & && 1.00 & 4.2e-56 & 14.48 & 0.32 \\
Perturb. &&& 0.98 & 6.0e-250 & 18.13 & 0.46 \\

\midrule

Regen. & \blackbox & & 1.00 & 4.5e-235 & 27.96 & 0.06 \\
Rinse & \blackbox & & 1.00 & 1.7e-70 & 25.10 & 0.14 \\
CtrlRegen+ & \blackbox & & 0.99 & 2.3e-22 & 23.45 & 0.22 \\

\midrule

VQ-Regen & \blackbox & & 1.00 & 3.3e-83 & 24.21 & 0.10 \\

\midrule

\advoptname-R$^\clubsuit$ & \blackbox & & 0.88 & 2.4e-32 & 32.34 & 0.28 \\
\multirow[t]{3}{*}{\advoptname-R} & \blackbox & & 0.97 & 5.2e-100 & 34.32 & 0.22 \\
 & \greybox & & 0.93 & 2.0e-47 & 32.45 & 0.26 \\
 & \whitebox & & 0.93 & 7.6e-18 & 31.67 & 0.26 \\

\midrule

Next-Frame P. & \blackbox & & 1.00 & 6.9e-226 & 11.31 & 0.53 \\

\midrule

BitFlipper & \whitebox{}$+$ & & 0.68 & 2.3e-8 & 19.22 & 0.24 \\

\midrule

\bitadvoptname-R & \whitebox{}$+$ & & 0.00 & 3.0e-1 & 46.17 & 0.01 \\

\bottomrule
\end{tabular}

%% file: tables/bitmark_main_forgery.tex
\begin{tabular}[t]{lll c S[table-format=1.1e4,tight-spacing=true] cc}
\toprule
 &  &  & \makecell{TPR} & \multicolumn{1}{c}{P-value} & PSNR & LPIPS \\
\midrule
Watermarked & &  & 1.00 & 0.0e0 & - & - \\
\midrule

 \multicolumn{3}{l}{Radio. $\infty$-2B, 25\%} & 0.99 & 1.1e-11 & - & - \\
 \multicolumn{3}{l}{\phantom{Radio. }$\infty$-2B, 50\% } & 1.00 & 2.3e-55 & - & - \\
 \multicolumn{3}{l}{\phantom{Radio. }$\infty$-2B, 100\% } & 1.00 & 8.4e-212 & - & - \\
 \multicolumn{3}{l}{\phantom{Radio. }SD2.1, 50\% } & 0.81 & 2.9e-5 & -  & - \\
 \multicolumn{3}{l}{\phantom{Radio. }SD2.1, 100\% }& 0.98 & 6.5e-14 & - & - \\
\midrule
\advoptname-F$^\clubsuit$ & \blackbox & & 0.88 & 3.7e-5 & 31.59 & 0.18 \\
\multirow[t]{3}{*}{\advoptname-F} & \blackbox & & 0.59 & 3.5e-3 & 31.51 & 0.19 \\
 & \greybox & & 0.99 & 1.7e-23 & 32.51 & 0.15 \\
 & \whitebox & & 1.00 & 3.4e-232 & 31.96 & 0.16 \\
\midrule

Reprompt-F & \blackbox & & 0.98 & 1.9e-70 & 14.09 & 0.51 \\
\midrule

Freq.Inj.-F~(A) & \blackbox & & 0.72 & 2.9e-6 & 37.50 & 0.06 \\

\phantom{Freq.Inj.-F}~(B) & \blackbox & & 0.79 & 7.4e-9 & 35.44 & 0.09 \\

\phantom{Freq.Inj.-F}~(C) & \blackbox & & 0.91 & 3.0e-23 & 29.76 & 0.18 \\

\bottomrule
\end{tabular}

%% file: tables/supp/full_setup.tex
\begin{tabular}{@{}l @{\hspace{1.7em}} l @{\hspace{0.7em}} l @{\hspace{1.7em}} l @{\hspace{0.7em}} l @{\hspace{0.7em}} l @{\hspace{1.7em}} l @{\hspace{0.7em}} l @{\hspace{0.7em}} l @{\hspace{1.7em}} l@{}}

\toprule
Attacker Model& \multicolumn{2}{c}{IndexMark} & \multicolumn{3}{c}{WMAR} & \multicolumn{3}{c}{ClusterMark} & \multicolumn{1}{c}{BitMark} \\
\cmidrule(r){2-3} \cmidrule(r){4-6} \cmidrule(r){7-9} \cmidrule(r){10-10}
 & \multicolumn{2}{c}{LlamaGen} & \multicolumn{3}{c}{ } & \multicolumn{2}{c}{LlamaGen} & \multicolumn{1}{c}{} & \multicolumn{1}{c}{$\infty$-2B} \\
 & \multicolumn{1}{c}{GPT-B} & \multicolumn{1}{c}{GPT-L} & \multicolumn{1}{c}{Anole} & \multicolumn{1}{c}{Taming} & \multicolumn{1}{c}{RAR-XL} & \multicolumn{1}{c}{GPT-B} & \multicolumn{1}{c}{GPT-L} & \multicolumn{1}{c}{RAR-XL} & \multicolumn{1}{c}{($d=32$)} \\
\midrule

\multicolumn{10}{c}{Regen. (60 steps)} \\
\midrule
SD-v2.1 & \blackbox & \blackbox & \blackbox & \blackbox & \blackbox & \blackbox & \blackbox & \blackbox & \blackbox \\
\midrule

\multicolumn{10}{c}{Rinse ($3\times$ Regen. 60 steps)} \\
\midrule
SD-v2.1 & \blackbox & \blackbox & \blackbox & \blackbox & \blackbox & \blackbox & \blackbox & \blackbox & \blackbox \\
\midrule

\multicolumn{10}{c}{CtrlRegen+ (Guidance=2, Strength=0.3)} \\
\midrule
SD-v2.1 & \blackbox & \blackbox & \blackbox & \blackbox & \blackbox & \blackbox & \blackbox & \blackbox & \blackbox \\
\midrule

\multicolumn{10}{c}{VQ-Regen (Substitution Rank $k=2$)} \\
\midrule
LlamaGen & \greybox\whitebox & \greybox\whitebox & \blackbox & \blackbox & \blackbox & \greybox\whitebox & \greybox\whitebox & \blackbox & \blackbox \\
Anole   & \blackbox & \blackbox & \greybox\whitebox & \blackbox & \blackbox & \blackbox & \blackbox & \blackbox & \blackbox \\
Taming  & \blackbox & \blackbox & \blackbox & \greybox\whitebox & \blackbox & \blackbox & \blackbox & \blackbox & \blackbox \\
RAR-XL   & \blackbox & \blackbox & \blackbox & \blackbox & \greybox\whitebox & \blackbox & \blackbox & \greybox\whitebox & \blackbox \\
\midrule

\multicolumn{10}{c}{\advoptname-Removal (300 steps, $c=\frac{8}{255}$, LR=0.001)} \\
\midrule
LlamaGen & \greybox\whitebox & \greybox\whitebox & \blackbox & \blackbox & \blackbox & \greybox\whitebox & \greybox\whitebox & \blackbox & \blackbox \\
Anole   & \blackbox & \blackbox & \greybox\whitebox & \blackbox & \blackbox & \blackbox & \blackbox & \blackbox & \blackbox \\
Taming  & \blackbox & \blackbox & \blackbox & \greybox\whitebox & \blackbox & \blackbox & \blackbox & \blackbox & \blackbox \\
RAR-XL   & \blackbox & \blackbox & \blackbox & \blackbox & \greybox\whitebox & \blackbox & \blackbox & \greybox\whitebox & \blackbox \\
$\infty$-2B ($d=16$)   & - & - & - & - & - & - & - & - & \greybox \\
$\infty$-2B ($d=24$)   & - & - & - & - & - & - & - & - & \greybox \\
$\infty$-2B ($d=32$)   & - & - & - & - & - & - & - & - & \whitebox \\
$\infty$-2B ($d=64$)   & - & - & - & - & - & - & - & - & \greybox \\
\midrule

\multicolumn{10}{c}{\advoptname-Forgery (300 steps, $c=\frac{8}{255}$, LR=0.001)} \\
\midrule
LlamaGen & \greybox\whitebox & \greybox\whitebox & \blackbox \tiny LR=0.01 & \blackbox \tiny LR=0.01 & \blackbox & \greybox\whitebox & \greybox\whitebox & \blackbox & \blackbox \tiny LR=0.01\\
Anole   & \blackbox & \blackbox & \greybox\whitebox \tiny LR=0.01  & \blackbox \tiny LR=0.01 & \blackbox & \blackbox & \blackbox & \blackbox & \blackbox \tiny LR=0.01\\
Taming  & \blackbox & \blackbox & \blackbox \tiny LR=0.01 & \greybox\whitebox  \tiny LR=0.01 & \blackbox & \blackbox & \blackbox & \blackbox & \blackbox \tiny LR=0.01\\
RAR-XL   & \blackbox & \blackbox & \blackbox \tiny LR=0.01 & \blackbox \tiny LR=0.01 & \greybox\whitebox & \blackbox & \blackbox & \greybox\whitebox & \blackbox \tiny LR=0.01\\
$\infty$ ($d=16$)   & - & - & - & - & - & - & - & - & \greybox \\
$\infty$ ($d=24$)   & - & - & - & - & - & - & - & - & \greybox \\
$\infty$ ($d=32$)   & - & - & - & - & - & - & - & - & \whitebox \\
$\infty$ ($d=64$)   & - & - & - & - & - & - & - & - & \greybox \\
\midrule

\multicolumn{10}{c}{BitFlipper ($\phi=2.2$)} \\
\midrule
$\infty$ ($d=32$)   & - & - & - & - & - & - & - & - & \whitebox{}$+$ \\
\midrule

\multicolumn{10}{c}{\bitadvoptname (100 steps, LR=0.0005)} \\
\midrule
$\infty$ ($d=32$)   & - & - & - & - & - & - & - & - & \whitebox{}$+$ \\
\midrule

\multicolumn{10}{c}{Next-Frame Prediction} \\
\midrule
SD-v2.1   & - & - & - & - & - & - & - & - & \blackbox \\
\midrule

\multicolumn{10}{c}{Reprompt-Forgery} \\
\midrule
SD-v2.1 & -  & - & - & - & - & - & - & - & \blackbox \\
\midrule

\multicolumn{10}{c}{Frequency Injection Forgery (Setting A,B,C)} \\
\midrule
- & -  & - & - & - & - & - & - & - & \blackbox \\

\bottomrule
\end{tabular}

%% file: tables/supp/full_results/IndexMark_GPT-B.tex
\begin{tabular}{lll @{\hspace{1.2em}} r @{\hspace{0.7em}} S[table-format=1.1e4, exponent-product=\times,tight-spacing=true] r @{\hspace{0.7em}} r}
\toprule
 &  &  & \multicolumn{1}{l}{TPR} & \multicolumn{1}{c}{P-Value} & \multicolumn{1}{c}{PSNR$\uparrow$} & \multicolumn{1}{c}{LPIPS$\downarrow$} \\
\midrule
Watermarked & &  & $1.00$ $[1.00,\,1.00]$ & 8.6e-78 & - & - \\
\midrule
\multicolumn{7}{c}{Removal} \\
\midrule
Geometric Transf. & & & $0.28$ $[0.27,\,0.30]$ & 1.2e-01 & ${16.154}_{\pm3.363}$ $[16.050,\,16.258]$ & ${0.159}_{\pm0.067}$ $[0.157,\,0.161]$ \\
Perturbations & & & $0.49$ $[0.48,\,0.50]$ & 1.4e-02 & ${17.291}_{\pm8.352}$ $[17.095,\,17.486]$ & ${0.422}_{\pm0.378}$ $[0.413,\,0.431]$ \\
\midrule
Regen. & & \blackbox & $0.33$ $[0.30,\,0.35]$ & 4.6e-02 & ${22.504}_{\pm3.064}$ $[22.314,\,22.694]$ & ${0.129}_{\pm0.047}$ $[0.126,\,0.131]$ \\
Rinse & & \blackbox & $0.03$ $[0.02,\,0.04]$ & 4.3e-01 & ${19.290}_{\pm2.360}$ $[19.144,\,19.437]$ & ${0.328}_{\pm0.087}$ $[0.323,\,0.333]$ \\
CtrlRegen{}$+$ & & \blackbox & $0.45$ $[0.41,\,0.48]$ & 1.9e-02 & ${24.331}_{\pm2.517}$ $[24.175,\,24.487]$ & ${0.117}_{\pm0.047}$ $[0.114,\,0.120]$ \\
\midrule
VQ-Regen\hspace{0.3cm} (aggregated) &  & \blackbox & $0.02$ $[0.02,\,0.03]$ & 4.3e-01 & ${19.484}_{\pm3.240}$ $[19.368,\,19.600]$ & ${0.159}_{\pm0.051}$ $[0.158,\,0.161]$ \\
\phantom{VQ-Regen\hspace{0.3cm}} RAR-XL &  & \blackbox & $0.01$ $[0.01,\,0.02]$ & 4.8e-01 & ${18.088}_{\pm2.677}$ $[17.922,\,18.254]$ & ${0.179}_{\pm0.050}$ $[0.175,\,0.182]$ \\
\phantom{VQ-Regen\hspace{0.3cm}} Anole &  & \blackbox & $0.02$ $[0.02,\,0.04]$ & 3.8e-01 & ${20.374}_{\pm3.362}$ $[20.166,\,20.583]$ & ${0.147}_{\pm0.050}$ $[0.144,\,0.150]$ \\
\phantom{VQ-Regen\hspace{0.3cm}} Taming &  & \blackbox & $0.03$ $[0.02,\,0.04]$ & 4.3e-01 & ${19.989}_{\pm3.170}$ $[19.793,\,20.186]$ & ${0.152}_{\pm0.046}$ $[0.149,\,0.155]$ \\
\cmidrule(l{1.95cm}){1-7}
\phantom{VQ-Regen\hspace{0.3cm}} LlamaGen &  & \greybox & $0.02$ $[0.01,\,0.03]$ & 7.1e-01 & ${21.496}_{\pm2.792}$ $[21.323,\,21.669]$ & ${0.106}_{\pm0.033}$ $[0.104,\,0.108]$ \\
\phantom{VQ-Regen\hspace{0.3cm}} LlamaGen &  & \whitebox & $0.01$ $[0.00,\,0.01]$ & 1.0e+00 & ${20.997}_{\pm2.688}$ $[20.830,\,21.164]$ & ${0.122}_{\pm0.037}$ $[0.119,\,0.124]$ \\
\midrule
\advoptname-R (aggregated) &  & \blackbox & $0.62$ $[0.56,\,0.67]$ & 4.4e-04 & ${32.125}_{\pm3.350}$ $[31.744,\,32.505]$ & ${0.098}_{\pm0.047}$ $[0.092,\,0.103]$ \\
\phantom{\advoptname-R} RAR-XL &  & \blackbox & $0.78$ $[0.69,\,0.85]$ & 1.5e-08 & ${31.940}_{\pm1.418}$ $[31.659,\,32.222]$ & ${0.101}_{\pm0.038}$ $[0.094,\,0.109]$ \\
\phantom{\advoptname-R} Anole &  & \blackbox & $0.58$ $[0.48,\,0.67]$ & 5.7e-04 & ${32.919}_{\pm5.496}$ $[31.829,\,34.010]$ & ${0.091}_{\pm0.053}$ $[0.080,\,0.101]$ \\
\phantom{\advoptname-R} Taming &  & \blackbox & $0.50$ $[0.40,\,0.60]$ & 1.2e-02 & ${31.515}_{\pm0.790}$ $[31.358,\,31.672]$ & ${0.101}_{\pm0.051}$ $[0.091,\,0.111]$ \\
\cmidrule(l{1.95cm}){1-7}
\phantom{\advoptname-R} LlamaGen &  & \greybox & $0.10$ $[0.06,\,0.17]$ & 5.7e-01 & ${31.373}_{\pm0.947}$ $[31.185,\,31.560]$ & ${0.124}_{\pm0.060}$ $[0.113,\,0.136]$ \\
\phantom{\advoptname-R} LlamaGen &  & \whitebox & $0.00$ $[0.00,\,0.04]$ & 1.0e+00 & ${31.474}_{\pm0.917}$ $[31.292,\,31.656]$ & ${0.032}_{\pm0.020}$ $[0.028,\,0.035]$ \\
\midrule
\multicolumn{7}{c}{Forgery} \\
\midrule
\advoptname-F (aggregated) &  & \blackbox& $0.10$ $[0.07,\,0.14]$ & 9.5e-02 & ${34.687}_{\pm3.614}$ $[34.277,\,35.098]$ & ${0.062}_{\pm0.038}$ $[0.058,\,0.067]$ \\
\phantom{\advoptname-F} RAR-XL &  & \blackbox & $0.10$ $[0.06,\,0.17]$ & 9.5e-02 & ${34.518}_{\pm3.625}$ $[33.799,\,35.237]$ & ${0.067}_{\pm0.041}$ $[0.059,\,0.075]$ \\
\phantom{\advoptname-F} Anole &  & \blackbox & $0.07$ $[0.03,\,0.14]$ & 9.5e-02 & ${35.175}_{\pm3.776}$ $[34.426,\,35.924]$ & ${0.059}_{\pm0.039}$ $[0.052,\,0.067]$ \\
\phantom{\advoptname-F} Taming &  & \blackbox & $0.14$ $[0.09,\,0.22]$ & 9.5e-02 & ${34.370}_{\pm3.417}$ $[33.692,\,35.048]$ & ${0.061}_{\pm0.036}$ $[0.054,\,0.068]$ \\
\cmidrule(l{1.95cm}){1-7}
\phantom{\advoptname-F} LlamaGen &  & \greybox & $0.14$ $[0.09,\,0.22]$ & 7.5e-02 & ${34.359}_{\pm3.235}$ $[33.717,\,35.001]$ & ${0.057}_{\pm0.034}$ $[0.050,\,0.063]$ \\
\phantom{\advoptname-F} LlamaGen &  & \whitebox & $1.00$ $[0.96,\,1.00]$ & 1.2e-64 & ${32.696}_{\pm1.065}$ $[32.485,\,32.908]$ & ${0.028}_{\pm0.013}$ $[0.025,\,0.030]$ \\
\bottomrule
\end{tabular}

%% file: tables/supp/full_results/IndexMark_GPT-L.tex
\begin{tabular}{lll @{\hspace{1.2em}} r @{\hspace{0.7em}} S[table-format=1.1e4, exponent-product=\times,tight-spacing=true] r @{\hspace{0.7em}} r}
\toprule
 &  &  & \multicolumn{1}{l}{TPR} & \multicolumn{1}{c}{P-Value} & \multicolumn{1}{c}{PSNR$\uparrow$} & \multicolumn{1}{c}{LPIPS$\downarrow$} \\
\midrule
Watermarked & &  & $1.00$ $[1.00,\,1.00]$ & 4.0e-174 & - & - \\
\midrule
\multicolumn{7}{c}{Removal} \\
\midrule
Geometric Transf. & & & $0.36$ $[0.35,\,0.38]$ & 4.4e-02 & ${16.172}_{\pm3.348}$ $[16.068,\,16.275]$ & ${0.196}_{\pm0.080}$ $[0.193,\,0.198]$ \\
Perturbations & & & $0.55$ $[0.54,\,0.56]$ & 1.2e-03 & ${17.822}_{\pm8.897}$ $[17.613,\,18.030]$ & ${0.447}_{\pm0.406}$ $[0.437,\,0.456]$ \\
\midrule
Regen. & & \blackbox & $0.71$ $[0.69,\,0.74]$ & 6.6e-04 & ${24.753}_{\pm3.576}$ $[24.531,\,24.975]$ & ${0.101}_{\pm0.040}$ $[0.098,\,0.103]$ \\
Rinse & & \blackbox & $0.04$ $[0.03,\,0.06]$ & 3.2e-01 & ${21.969}_{\pm2.874}$ $[21.791,\,22.148]$ & ${0.222}_{\pm0.072}$ $[0.218,\,0.227]$ \\
CtrlRegen{}$+$ & & \blackbox & $0.26$ $[0.23,\,0.29]$ & 6.2e-02 & ${23.948}_{\pm2.841}$ $[23.772,\,24.125]$ & ${0.136}_{\pm0.051}$ $[0.133,\,0.139]$ \\
\midrule
VQ-Regen\hspace{0.3cm} (aggregated) &  & \blackbox & $0.04$ $[0.04,\,0.05]$ & 3.9e-01 & ${21.160}_{\pm3.673}$ $[21.028,\,21.291]$ & ${0.147}_{\pm0.051}$ $[0.145,\,0.148]$ \\
\phantom{VQ-Regen\hspace{0.3cm}} RAR-XL &  & \blackbox & $0.02$ $[0.01,\,0.03]$ & 5.2e-01 & ${19.555}_{\pm3.057}$ $[19.365,\,19.745]$ & ${0.168}_{\pm0.051}$ $[0.165,\,0.171]$ \\
\phantom{VQ-Regen\hspace{0.3cm}} Anole &  & \blackbox & $0.07$ $[0.05,\,0.09]$ & 2.7e-01 & ${22.187}_{\pm3.809}$ $[21.951,\,22.424]$ & ${0.134}_{\pm0.047}$ $[0.131,\,0.137]$ \\
\phantom{VQ-Regen\hspace{0.3cm}} Taming &  & \blackbox & $0.04$ $[0.03,\,0.06]$ & 3.5e-01 & ${21.736}_{\pm3.563}$ $[21.515,\,21.957]$ & ${0.139}_{\pm0.046}$ $[0.136,\,0.141]$ \\
\cmidrule(l{1.95cm}){1-7}
\phantom{VQ-Regen\hspace{0.3cm}} LlamaGen &  & \greybox & $0.01$ $[0.01,\,0.02]$ & 8.9e-01 & ${23.108}_{\pm3.102}$ $[22.915,\,23.300]$ & ${0.098}_{\pm0.034}$ $[0.095,\,0.100]$ \\
\phantom{VQ-Regen\hspace{0.3cm}} LlamaGen &  & \whitebox & $0.00$ $[0.00,\,0.01]$ & 1.0e+00 & ${22.616}_{\pm2.922}$ $[22.434,\,22.797]$ & ${0.110}_{\pm0.036}$ $[0.108,\,0.112]$ \\
\midrule
\advoptname-R (aggregated) &  & \blackbox & $0.64$ $[0.58,\,0.69]$ & 3.1e-05 & ${31.753}_{\pm1.137}$ $[31.624,\,31.882]$ & ${0.148}_{\pm0.071}$ $[0.140,\,0.156]$ \\
\phantom{\advoptname-R} RAR-XL &  & \blackbox & $0.83$ $[0.74,\,0.89]$ & 6.4e-11 & ${32.329}_{\pm1.407}$ $[32.050,\,32.608]$ & ${0.138}_{\pm0.052}$ $[0.128,\,0.149]$ \\
\phantom{\advoptname-R} Anole &  & \blackbox & $0.56$ $[0.46,\,0.65]$ & 1.1e-03 & ${31.426}_{\pm0.982}$ $[31.231,\,31.621]$ & ${0.152}_{\pm0.079}$ $[0.136,\,0.167]$ \\
\phantom{\advoptname-R} Taming &  & \blackbox & $0.53$ $[0.43,\,0.62]$ & 3.5e-03 & ${31.505}_{\pm0.676}$ $[31.371,\,31.639]$ & ${0.154}_{\pm0.079}$ $[0.139,\,0.170]$ \\
\cmidrule(l{1.95cm}){1-7}
\phantom{\advoptname-R} LlamaGen &  & \greybox & $0.11$ $[0.06,\,0.19]$ & 7.1e-01 & ${31.966}_{\pm1.288}$ $[31.710,\,32.222]$ & ${0.178}_{\pm0.079}$ $[0.162,\,0.193]$ \\
\phantom{\advoptname-R} LlamaGen &  & \whitebox & $0.00$ $[0.00,\,0.04]$ & 1.0e+00 & ${32.150}_{\pm1.126}$ $[31.926,\,32.373]$ & ${0.054}_{\pm0.034}$ $[0.047,\,0.061]$ \\
\midrule
\multicolumn{7}{c}{Forgery} \\
\midrule
\advoptname-F (aggregated) &  & \blackbox& $0.09$ $[0.07,\,0.13]$ & 1.1e-01 & ${34.823}_{\pm3.577}$ $[34.417,\,35.230]$ & ${0.074}_{\pm0.048}$ $[0.068,\,0.079]$ \\
\phantom{\advoptname-F} RAR-XL &  & \blackbox & $0.08$ $[0.04,\,0.15]$ & 1.1e-01 & ${33.618}_{\pm2.673}$ $[33.088,\,34.148]$ & ${0.088}_{\pm0.044}$ $[0.079,\,0.097]$ \\
\phantom{\advoptname-F} Anole &  & \blackbox & $0.07$ $[0.03,\,0.14]$ & 1.2e-01 & ${36.118}_{\pm3.872}$ $[35.350,\,36.886]$ & ${0.062}_{\pm0.052}$ $[0.052,\,0.072]$ \\
\phantom{\advoptname-F} Taming &  & \blackbox & $0.13$ $[0.08,\,0.21]$ & 1.1e-01 & ${34.735}_{\pm3.652}$ $[34.010,\,35.459]$ & ${0.072}_{\pm0.046}$ $[0.063,\,0.081]$ \\
\cmidrule(l{1.95cm}){1-7}
\phantom{\advoptname-F} LlamaGen &  & \greybox & $0.14$ $[0.09,\,0.22]$ & 1.1e-01 & ${34.951}_{\pm3.514}$ $[34.254,\,35.648]$ & ${0.061}_{\pm0.037}$ $[0.053,\,0.068]$ \\
\phantom{\advoptname-F} LlamaGen &  & \whitebox & $1.00$ $[0.96,\,1.00]$ & 4.2e-153 & ${33.000}_{\pm1.127}$ $[32.777,\,33.224]$ & ${0.033}_{\pm0.014}$ $[0.030,\,0.036]$ \\
\bottomrule
\end{tabular}

%% file: tables/supp/full_results/WMAR_chameleon.tex
\begin{tabular}{lll @{\hspace{1.2em}} r @{\hspace{0.7em}} S[table-format=1.1e4, exponent-product=\times,tight-spacing=true] r @{\hspace{0.7em}} r}
\toprule
 &  &  & \multicolumn{1}{l}{TPR} & \multicolumn{1}{c}{P-Value} & \multicolumn{1}{c}{PSNR$\uparrow$} & \multicolumn{1}{c}{LPIPS$\downarrow$} \\
\midrule
Watermarked & &  & $0.98$ $[0.97,\,0.99]$ & 3.1e-47 & - & - \\
\midrule
\multicolumn{7}{c}{Removal} \\
\midrule
Geometric Transf. & & & $0.13$ $[0.12,\,0.14]$ & 1.9e-01 & ${15.638}_{\pm3.662}$ $[15.525,\,15.752]$ & ${0.217}_{\pm0.101}$ $[0.214,\,0.220]$ \\
Perturbations & & & $0.42$ $[0.41,\,0.43]$ & 4.2e-02 & ${17.442}_{\pm8.533}$ $[17.242,\,17.642]$ & ${0.416}_{\pm0.388}$ $[0.407,\,0.425]$ \\
\midrule
Regen. & & \blackbox & $0.63$ $[0.60,\,0.66]$ & 3.0e-03 & ${26.714}_{\pm3.561}$ $[26.493,\,26.935]$ & ${0.059}_{\pm0.030}$ $[0.057,\,0.061]$ \\
Rinse & & \blackbox & $0.04$ $[0.03,\,0.06]$ & 2.3e-01 & ${24.036}_{\pm2.953}$ $[23.853,\,24.219]$ & ${0.141}_{\pm0.063}$ $[0.137,\,0.144]$ \\
CtrlRegen{}$+$ & & \blackbox & $0.18$ $[0.16,\,0.21]$ & 8.8e-02 & ${25.036}_{\pm3.434}$ $[24.823,\,25.249]$ & ${0.077}_{\pm0.035}$ $[0.075,\,0.079]$ \\
\midrule
VQ-Regen\hspace{0.3cm} (aggregated) &  & \blackbox & $0.21$ $[0.19,\,0.22]$ & 1.4e-01 & ${22.678}_{\pm3.409}$ $[22.556,\,22.800]$ & ${0.094}_{\pm0.043}$ $[0.093,\,0.096]$ \\
\phantom{VQ-Regen\hspace{0.3cm}} LlamaGen &  & \blackbox & $0.53$ $[0.50,\,0.56]$ & 8.0e-03 & ${24.132}_{\pm3.117}$ $[23.939,\,24.326]$ & ${0.071}_{\pm0.030}$ $[0.069,\,0.073]$ \\
\phantom{VQ-Regen\hspace{0.3cm}} RAR-XL &  & \blackbox & $0.01$ $[0.00,\,0.01]$ & 4.6e-01 & ${20.698}_{\pm2.930}$ $[20.516,\,20.880]$ & ${0.122}_{\pm0.045}$ $[0.119,\,0.125]$ \\
\phantom{VQ-Regen\hspace{0.3cm}} Taming &  & \blackbox & $0.09$ $[0.07,\,0.11]$ & 1.4e-01 & ${23.203}_{\pm3.202}$ $[23.005,\,23.402]$ & ${0.090}_{\pm0.036}$ $[0.088,\,0.092]$ \\
\cmidrule(l{1.95cm}){1-7}
\phantom{VQ-Regen\hspace{0.3cm}} Anole &  & \greybox & $0.52$ $[0.49,\,0.55]$ & 1.1e-02 & ${25.553}_{\pm3.456}$ $[25.339,\,25.767]$ & ${0.060}_{\pm0.028}$ $[0.059,\,0.062]$ \\
\phantom{VQ-Regen\hspace{0.3cm}} Anole &  & \whitebox & $0.20$ $[0.18,\,0.23]$ & 6.9e-02 & ${24.358}_{\pm3.738}$ $[24.126,\,24.590]$ & ${0.068}_{\pm0.029}$ $[0.066,\,0.070]$ \\
\midrule
\advoptname-R (aggregated) &  & \blackbox & $0.22$ $[0.17,\,0.27]$ & 3.9e-01 & ${32.636}_{\pm1.445}$ $[32.471,\,32.800]$ & ${0.155}_{\pm0.055}$ $[0.148,\,0.161]$ \\
\phantom{\advoptname-R} LlamaGen &  & \blackbox & $0.03$ $[0.01,\,0.08]$ & 8.0e-01 & ${32.350}_{\pm1.399}$ $[32.073,\,32.628]$ & ${0.178}_{\pm0.057}$ $[0.166,\,0.189]$ \\
\phantom{\advoptname-R} RAR-XL &  & \blackbox & $0.49$ $[0.39,\,0.59]$ & 1.9e-02 & ${33.126}_{\pm1.463}$ $[32.836,\,33.416]$ & ${0.130}_{\pm0.035}$ $[0.123,\,0.137]$ \\
\phantom{\advoptname-R} Taming &  & \blackbox & $0.13$ $[0.08,\,0.21]$ & 3.9e-01 & ${32.431}_{\pm1.357}$ $[32.162,\,32.700]$ & ${0.156}_{\pm0.060}$ $[0.144,\,0.168]$ \\
\cmidrule(l{1.95cm}){1-7}
\phantom{\advoptname-R} Anole &  & \greybox & $0.01$ $[0.00,\,0.05]$ & 9.3e-01 & ${32.435}_{\pm2.413}$ $[31.956,\,32.913]$ & ${0.158}_{\pm0.057}$ $[0.147,\,0.169]$ \\
\phantom{\advoptname-R} Anole &  & \whitebox & $0.01$ $[0.00,\,0.05]$ & 9.1e-01 & ${34.044}_{\pm3.513}$ $[33.347,\,34.741]$ & ${0.117}_{\pm0.058}$ $[0.106,\,0.129]$ \\
\midrule
\multicolumn{7}{c}{Forgery} \\
\midrule
\advoptname-F (aggregated) &  & \blackbox& $0.00$ $[0.00,\,0.01]$ & 3.5e-01 & ${31.341}_{\pm0.338}$ $[31.303,\,31.380]$ & ${0.128}_{\pm0.058}$ $[0.121,\,0.135]$ \\
\phantom{\advoptname-F} LlamaGen &  & \blackbox & $0.00$ $[0.00,\,0.04]$ & 3.4e-01 & ${31.430}_{\pm0.363}$ $[31.358,\,31.502]$ & ${0.120}_{\pm0.055}$ $[0.109,\,0.131]$ \\
\phantom{\advoptname-F} RAR-XL &  & \blackbox & $0.00$ $[0.00,\,0.04]$ & 3.4e-01 & ${31.251}_{\pm0.378}$ $[31.176,\,31.326]$ & ${0.136}_{\pm0.058}$ $[0.125,\,0.148]$ \\
\phantom{\advoptname-F} Taming &  & \blackbox & $0.00$ $[0.00,\,0.04]$ & 3.6e-01 & ${31.343}_{\pm0.235}$ $[31.296,\,31.390]$ & ${0.128}_{\pm0.061}$ $[0.116,\,0.140]$ \\
\cmidrule(l{1.95cm}){1-7}
\phantom{\advoptname-F} Anole &  & \greybox & $0.00$ $[0.00,\,0.04]$ & 4.5e-01 & ${31.271}_{\pm0.233}$ $[31.224,\,31.317]$ & ${0.135}_{\pm0.064}$ $[0.122,\,0.148]$ \\
\phantom{\advoptname-F} Anole &  & \whitebox & $0.07$ $[0.03,\,0.14]$ & 3.0e-01 & ${31.088}_{\pm0.249}$ $[31.038,\,31.137]$ & ${0.109}_{\pm0.050}$ $[0.098,\,0.119]$ \\
\bottomrule
\end{tabular}

%% file: tables/supp/full_results/WMAR_taming.tex
\begin{tabular}{lll @{\hspace{1.2em}} r @{\hspace{0.7em}} S[table-format=1.1e4, exponent-product=\times,tight-spacing=true] r @{\hspace{0.7em}} r}
\toprule
 &  &  & \multicolumn{1}{l}{TPR} & \multicolumn{1}{c}{P-Value} & \multicolumn{1}{c}{PSNR$\uparrow$} & \multicolumn{1}{c}{LPIPS$\downarrow$} \\
\midrule
Watermarked & &  & $0.99$ $[0.99,\,1.00]$ & 1.6e-55 & - & - \\
\midrule
\multicolumn{7}{c}{Removal} \\
\midrule
Geometric Transf. & & & $0.25$ $[0.24,\,0.27]$ & 1.5e-01 & ${17.270}_{\pm4.344}$ $[17.135,\,17.404]$ & ${0.148}_{\pm0.067}$ $[0.146,\,0.150]$ \\
Perturbations & & & $0.40$ $[0.39,\,0.42]$ & 6.2e-02 & ${17.913}_{\pm8.910}$ $[17.704,\,18.122]$ & ${0.407}_{\pm0.369}$ $[0.398,\,0.415]$ \\
\midrule
Regen. & & \blackbox & $0.10$ $[0.08,\,0.12]$ & 2.4e-01 & ${23.538}_{\pm4.059}$ $[23.286,\,23.790]$ & ${0.119}_{\pm0.055}$ $[0.116,\,0.123]$ \\
Rinse & & \blackbox & $0.01$ $[0.01,\,0.02]$ & 5.1e-01 & ${20.364}_{\pm3.223}$ $[20.164,\,20.564]$ & ${0.327}_{\pm0.101}$ $[0.321,\,0.334]$ \\
CtrlRegen{}$+$ & & \blackbox & $0.16$ $[0.14,\,0.19]$ & 1.6e-01 & ${24.404}_{\pm3.539}$ $[24.185,\,24.624]$ & ${0.106}_{\pm0.052}$ $[0.103,\,0.109]$ \\
\midrule
VQ-Regen\hspace{0.3cm} (aggregated) &  & \blackbox & $0.06$ $[0.05,\,0.07]$ & 3.3e-01 & ${21.418}_{\pm4.182}$ $[21.269,\,21.568]$ & ${0.114}_{\pm0.051}$ $[0.112,\,0.115]$ \\
\phantom{VQ-Regen\hspace{0.3cm}} LlamaGen &  & \blackbox & $0.11$ $[0.10,\,0.14]$ & 1.6e-01 & ${22.330}_{\pm4.003}$ $[22.082,\,22.578]$ & ${0.089}_{\pm0.037}$ $[0.087,\,0.092]$ \\
\phantom{VQ-Regen\hspace{0.3cm}} RAR-XL &  & \blackbox & $0.02$ $[0.01,\,0.03]$ & 4.8e-01 & ${19.842}_{\pm3.662}$ $[19.615,\,20.069]$ & ${0.141}_{\pm0.051}$ $[0.138,\,0.144]$ \\
\phantom{VQ-Regen\hspace{0.3cm}} Anole &  & \blackbox & $0.04$ $[0.03,\,0.06]$ & 3.4e-01 & ${22.083}_{\pm4.392}$ $[21.811,\,22.356]$ & ${0.111}_{\pm0.049}$ $[0.108,\,0.114]$ \\
\cmidrule(l{1.95cm}){1-7}
\phantom{VQ-Regen\hspace{0.3cm}} Taming &  & \greybox & $0.02$ $[0.01,\,0.03]$ & 4.4e-01 & ${22.267}_{\pm3.714}$ $[22.037,\,22.497]$ & ${0.104}_{\pm0.035}$ $[0.101,\,0.106]$ \\
\phantom{VQ-Regen\hspace{0.3cm}} Taming &  & \whitebox & $0.02$ $[0.01,\,0.03]$ & 5.2e-01 & ${22.684}_{\pm3.494}$ $[22.467,\,22.900]$ & ${0.108}_{\pm0.038}$ $[0.105,\,0.110]$ \\
\midrule
\advoptname-R (aggregated) &  & \blackbox & $0.33$ $[0.28,\,0.39]$ & 1.9e-01 & ${32.227}_{\pm1.859}$ $[32.016,\,32.439]$ & ${0.122}_{\pm0.060}$ $[0.115,\,0.129]$ \\
\phantom{\advoptname-R} LlamaGen &  & \blackbox & $0.18$ $[0.12,\,0.27]$ & 5.3e-01 & ${31.961}_{\pm1.360}$ $[31.692,\,32.231]$ & ${0.141}_{\pm0.069}$ $[0.127,\,0.155]$ \\
\phantom{\advoptname-R} RAR-XL &  & \blackbox & $0.52$ $[0.42,\,0.62]$ & 4.5e-03 & ${32.765}_{\pm1.741}$ $[32.419,\,33.110]$ & ${0.109}_{\pm0.041}$ $[0.101,\,0.117]$ \\
\phantom{\advoptname-R} Anole &  & \blackbox & $0.30$ $[0.22,\,0.40]$ & 3.2e-01 & ${31.956}_{\pm2.263}$ $[31.507,\,32.405]$ & ${0.117}_{\pm0.063}$ $[0.104,\,0.129]$ \\
\cmidrule(l{1.95cm}){1-7}
\phantom{\advoptname-R} Taming &  & \greybox & $0.01$ $[0.00,\,0.05]$ & 9.1e-01 & ${32.289}_{\pm1.773}$ $[31.937,\,32.641]$ & ${0.121}_{\pm0.066}$ $[0.108,\,0.134]$ \\
\phantom{\advoptname-R} Taming &  & \whitebox & $0.00$ $[0.00,\,0.04]$ & 9.7e-01 & ${31.347}_{\pm0.996}$ $[31.149,\,31.545]$ & ${0.127}_{\pm0.061}$ $[0.115,\,0.139]$ \\
\midrule
\multicolumn{7}{c}{Forgery} \\
\midrule
\advoptname-F (aggregated) &  & \blackbox& $0.12$ $[0.09,\,0.16]$ & 7.5e-02 & ${31.181}_{\pm0.301}$ $[31.147,\,31.215]$ & ${0.106}_{\pm0.047}$ $[0.101,\,0.111]$ \\
\phantom{\advoptname-F} LlamaGen &  & \blackbox & $0.13$ $[0.08,\,0.21]$ & 7.8e-02 & ${31.264}_{\pm0.309}$ $[31.202,\,31.325]$ & ${0.098}_{\pm0.044}$ $[0.089,\,0.106]$ \\
\phantom{\advoptname-F} RAR-XL &  & \blackbox & $0.09$ $[0.05,\,0.16]$ & 8.1e-02 & ${31.101}_{\pm0.333}$ $[31.035,\,31.167]$ & ${0.109}_{\pm0.045}$ $[0.100,\,0.118]$ \\
\phantom{\advoptname-F} Anole &  & \blackbox & $0.14$ $[0.09,\,0.22]$ & 6.2e-02 & ${31.178}_{\pm0.233}$ $[31.132,\,31.225]$ & ${0.111}_{\pm0.050}$ $[0.101,\,0.121]$ \\
\cmidrule(l{1.95cm}){1-7}
\phantom{\advoptname-F} Taming &  & \greybox & $0.17$ $[0.11,\,0.26]$ & 5.7e-02 & ${31.191}_{\pm0.233}$ $[31.145,\,31.237]$ & ${0.102}_{\pm0.045}$ $[0.093,\,0.111]$ \\
\phantom{\advoptname-F} Taming &  & \whitebox & $0.24$ $[0.17,\,0.33]$ & 3.5e-02 & ${31.052}_{\pm0.285}$ $[30.996,\,31.109]$ & ${0.099}_{\pm0.041}$ $[0.091,\,0.108]$ \\
\bottomrule
\end{tabular}

%% file: tables/supp/full_results/WMAR_rar_xl.tex
\begin{tabular}{lll @{\hspace{1.2em}} r @{\hspace{0.7em}} S[table-format=1.1e4, exponent-product=\times,tight-spacing=true] r @{\hspace{0.7em}} r}
\toprule
 &  &  & \multicolumn{1}{l}{TPR} & \multicolumn{1}{c}{P-Value} & \multicolumn{1}{c}{PSNR$\uparrow$} & \multicolumn{1}{c}{LPIPS$\downarrow$} \\
\midrule
Watermarked & &  & $1.00$ $[1.00,\,1.00]$ & 1.3e-35 & - & - \\
\midrule
\multicolumn{7}{c}{Removal} \\
\midrule
Geometric Transf. & & & $0.51$ $[0.50,\,0.53]$ & 1.0e-02 & ${16.426}_{\pm3.282}$ $[16.325,\,16.528]$ & ${0.159}_{\pm0.065}$ $[0.156,\,0.161]$ \\
Perturbations & & & $0.53$ $[0.51,\,0.54]$ & 6.4e-03 & ${17.456}_{\pm8.371}$ $[17.260,\,17.652]$ & ${0.429}_{\pm0.385}$ $[0.420,\,0.438]$ \\
\midrule
Regen. & & \blackbox & $0.28$ $[0.26,\,0.31]$ & 6.2e-02 & ${22.660}_{\pm2.579}$ $[22.500,\,22.820]$ & ${0.137}_{\pm0.038}$ $[0.135,\,0.139]$ \\
Rinse & & \blackbox & $0.01$ $[0.01,\,0.02]$ & 4.5e-01 & ${19.394}_{\pm2.010}$ $[19.269,\,19.518]$ & ${0.360}_{\pm0.072}$ $[0.355,\,0.364]$ \\
CtrlRegen{}$+$ & & \blackbox & $0.44$ $[0.41,\,0.47]$ & 1.8e-02 & ${24.391}_{\pm2.198}$ $[24.254,\,24.527]$ & ${0.125}_{\pm0.040}$ $[0.123,\,0.128]$ \\
\midrule
VQ-Regen\hspace{0.3cm} (aggregated) &  & \blackbox & $0.09$ $[0.08,\,0.10]$ & 2.3e-01 & ${20.672}_{\pm2.769}$ $[20.573,\,20.772]$ & ${0.147}_{\pm0.044}$ $[0.145,\,0.149]$ \\
\phantom{VQ-Regen\hspace{0.3cm}} LlamaGen &  & \blackbox & $0.20$ $[0.17,\,0.22]$ & 7.0e-02 & ${21.013}_{\pm2.616}$ $[20.851,\,21.176]$ & ${0.127}_{\pm0.036}$ $[0.125,\,0.129]$ \\
\phantom{VQ-Regen\hspace{0.3cm}} Anole &  & \blackbox & $0.03$ $[0.02,\,0.05]$ & 3.3e-01 & ${20.687}_{\pm2.888}$ $[20.507,\,20.866]$ & ${0.152}_{\pm0.045}$ $[0.150,\,0.155]$ \\
\phantom{VQ-Regen\hspace{0.3cm}} Taming &  & \blackbox & $0.03$ $[0.02,\,0.05]$ & 3.5e-01 & ${20.317}_{\pm2.755}$ $[20.146,\,20.488]$ & ${0.162}_{\pm0.043}$ $[0.159,\,0.164]$ \\
\cmidrule(l{1.95cm}){1-7}
\phantom{VQ-Regen\hspace{0.3cm}} RAR-XL &  & \greybox & $0.01$ $[0.01,\,0.02]$ & 4.5e-01 & ${18.579}_{\pm2.371}$ $[18.432,\,18.727]$ & ${0.167}_{\pm0.041}$ $[0.165,\,0.170]$ \\
\phantom{VQ-Regen\hspace{0.3cm}} RAR-XL &  & \whitebox & $0.01$ $[0.00,\,0.01]$ & 5.1e-01 & ${18.249}_{\pm2.269}$ $[18.109,\,18.390]$ & ${0.204}_{\pm0.046}$ $[0.201,\,0.207]$ \\
\midrule
\advoptname-R (aggregated) &  & \blackbox & $0.71$ $[0.66,\,0.76]$ & 2.0e-05 & ${31.527}_{\pm0.891}$ $[31.426,\,31.629]$ & ${0.111}_{\pm0.048}$ $[0.106,\,0.117]$ \\
\phantom{\advoptname-R} LlamaGen &  & \blackbox & $0.60$ $[0.50,\,0.69]$ & 1.1e-03 & ${31.508}_{\pm0.934}$ $[31.323,\,31.693]$ & ${0.123}_{\pm0.052}$ $[0.113,\,0.133]$ \\
\phantom{\advoptname-R} Anole &  & \blackbox & $0.78$ $[0.69,\,0.85]$ & 1.7e-06 & ${31.701}_{\pm0.973}$ $[31.508,\,31.894]$ & ${0.104}_{\pm0.044}$ $[0.095,\,0.112]$ \\
\phantom{\advoptname-R} Taming &  & \blackbox & $0.76$ $[0.67,\,0.83]$ & 6.7e-07 & ${31.373}_{\pm0.724}$ $[31.230,\,31.517]$ & ${0.107}_{\pm0.046}$ $[0.097,\,0.116]$ \\
\cmidrule(l{1.95cm}){1-7}
\phantom{\advoptname-R} RAR-XL &  & \greybox & $0.31$ $[0.23,\,0.41]$ & 1.3e-01 & ${31.910}_{\pm1.123}$ $[31.687,\,32.133]$ & ${0.108}_{\pm0.036}$ $[0.101,\,0.115]$ \\
\phantom{\advoptname-R} RAR-XL &  & \whitebox & $0.00$ $[0.00,\,0.04]$ & 9.3e-01 & ${31.450}_{\pm1.378}$ $[31.177,\,31.724]$ & ${0.131}_{\pm0.044}$ $[0.123,\,0.140]$ \\
\midrule
\multicolumn{7}{c}{Forgery} \\
\midrule
\advoptname-F (aggregated) &  & \blackbox& $0.07$ $[0.05,\,0.11]$ & 1.3e-01 & ${34.795}_{\pm3.705}$ $[34.374,\,35.216]$ & ${0.059}_{\pm0.040}$ $[0.054,\,0.063]$ \\
\phantom{\advoptname-F} LlamaGen &  & \blackbox & $0.06$ $[0.03,\,0.12]$ & 1.4e-01 & ${35.101}_{\pm3.816}$ $[34.344,\,35.858]$ & ${0.050}_{\pm0.034}$ $[0.043,\,0.056]$ \\
\phantom{\advoptname-F} Anole &  & \blackbox & $0.08$ $[0.04,\,0.15]$ & 1.2e-01 & ${34.906}_{\pm3.901}$ $[34.132,\,35.680]$ & ${0.065}_{\pm0.046}$ $[0.056,\,0.074]$ \\
\phantom{\advoptname-F} Taming &  & \blackbox & $0.08$ $[0.04,\,0.15]$ & 1.3e-01 & ${34.379}_{\pm3.375}$ $[33.709,\,35.049]$ & ${0.062}_{\pm0.038}$ $[0.054,\,0.069]$ \\
\cmidrule(l{1.95cm}){1-7}
\phantom{\advoptname-F} RAR-XL &  & \greybox & $0.13$ $[0.08,\,0.21]$ & 1.1e-01 & ${34.313}_{\pm3.318}$ $[33.655,\,34.971]$ & ${0.069}_{\pm0.042}$ $[0.061,\,0.077]$ \\
\phantom{\advoptname-F} RAR-XL &  & \whitebox & $0.15$ $[0.09,\,0.23]$ & 8.3e-02 & ${33.627}_{\pm3.360}$ $[32.960,\,34.294]$ & ${0.079}_{\pm0.047}$ $[0.069,\,0.088]$ \\
\bottomrule
\end{tabular}

%% file: tables/supp/full_results/arwm_GPT-B.tex
\begin{tabular}{lll @{\hspace{1.2em}} r @{\hspace{0.7em}} S[table-format=1.1e4, exponent-product=\times,tight-spacing=true] r @{\hspace{0.7em}} r}
\toprule
 &  &  & \multicolumn{1}{l}{TPR} & \multicolumn{1}{c}{P-Value} & \multicolumn{1}{c}{PSNR$\uparrow$} & \multicolumn{1}{c}{LPIPS$\downarrow$} \\
\midrule
Watermarked & &  & $1.00$ $[1.00,\,1.00]$ & 7.2e-58 & - & - \\
\midrule
\multicolumn{7}{c}{Removal} \\
\midrule
Geometric Transf. & & & $0.40$ $[0.38,\,0.41]$ & 3.5e-02 & ${16.437}_{\pm3.392}$ $[16.331,\,16.542]$ & ${0.157}_{\pm0.067}$ $[0.155,\,0.159]$ \\
Perturbations & & & $0.81$ $[0.80,\,0.82]$ & 1.6e-18 & ${17.481}_{\pm8.437}$ $[17.284,\,17.679]$ & ${0.423}_{\pm0.387}$ $[0.414,\,0.432]$ \\
\midrule
Regen. & & \blackbox & $0.81$ $[0.79,\,0.84]$ & 9.5e-05 & ${23.029}_{\pm2.938}$ $[22.846,\,23.211]$ & ${0.122}_{\pm0.040}$ $[0.119,\,0.124]$ \\
Rinse & & \blackbox & $0.17$ $[0.15,\,0.19]$ & 1.0e-01 & ${19.633}_{\pm2.261}$ $[19.493,\,19.773]$ & ${0.320}_{\pm0.078}$ $[0.315,\,0.324]$ \\
CtrlRegen{}$+$ & & \blackbox & $0.78$ $[0.76,\,0.81]$ & 1.3e-04 & ${24.656}_{\pm2.530}$ $[24.499,\,24.813]$ & ${0.111}_{\pm0.041}$ $[0.108,\,0.113]$ \\
\midrule
VQ-Regen\hspace{0.3cm} (aggregated) &  & \blackbox & $0.19$ $[0.18,\,0.21]$ & 1.0e-01 & ${19.854}_{\pm3.102}$ $[19.743,\,19.965]$ & ${0.157}_{\pm0.048}$ $[0.155,\,0.159]$ \\
\phantom{VQ-Regen\hspace{0.3cm}} RAR-XL &  & \blackbox & $0.09$ $[0.07,\,0.11]$ & 2.0e-01 & ${18.358}_{\pm2.552}$ $[18.200,\,18.517]$ & ${0.178}_{\pm0.046}$ $[0.175,\,0.181]$ \\
\phantom{VQ-Regen\hspace{0.3cm}} Anole &  & \blackbox & $0.28$ $[0.25,\,0.30]$ & 6.2e-02 & ${20.787}_{\pm3.150}$ $[20.591,\,20.982]$ & ${0.144}_{\pm0.046}$ $[0.141,\,0.147]$ \\
\phantom{VQ-Regen\hspace{0.3cm}} Taming &  & \blackbox & $0.22$ $[0.19,\,0.24]$ & 8.1e-02 & ${20.416}_{\pm3.004}$ $[20.229,\,20.602]$ & ${0.149}_{\pm0.043}$ $[0.146,\,0.152]$ \\
\cmidrule(l{1.95cm}){1-7}
\phantom{VQ-Regen\hspace{0.3cm}} LlamaGen &  & \greybox & $0.92$ $[0.90,\,0.94]$ & 5.5e-05 & ${21.811}_{\pm2.663}$ $[21.646,\,21.976]$ & ${0.104}_{\pm0.031}$ $[0.102,\,0.106]$ \\
\phantom{VQ-Regen\hspace{0.3cm}} LlamaGen &  & \whitebox & $0.94$ $[0.92,\,0.95]$ & 2.5e-05 & ${18.772}_{\pm1.849}$ $[18.658,\,18.887]$ & ${0.174}_{\pm0.036}$ $[0.172,\,0.176]$ \\
\midrule
\advoptname-R (aggregated) &  & \blackbox & $0.75$ $[0.69,\,0.79]$ & 1.8e-05 & ${31.939}_{\pm2.182}$ $[31.691,\,32.187]$ & ${0.104}_{\pm0.044}$ $[0.099,\,0.109]$ \\
\phantom{\advoptname-R} RAR-XL &  & \blackbox & $0.87$ $[0.79,\,0.92]$ & 2.4e-08 & ${32.119}_{\pm1.131}$ $[31.895,\,32.344]$ & ${0.104}_{\pm0.035}$ $[0.097,\,0.111]$ \\
\phantom{\advoptname-R} Anole &  & \blackbox & $0.71$ $[0.61,\,0.79]$ & 5.5e-05 & ${32.067}_{\pm3.465}$ $[31.379,\,32.754]$ & ${0.103}_{\pm0.049}$ $[0.093,\,0.113]$ \\
\phantom{\advoptname-R} Taming &  & \blackbox & $0.66$ $[0.56,\,0.75]$ & 3.5e-04 & ${31.632}_{\pm0.976}$ $[31.438,\,31.825]$ & ${0.105}_{\pm0.046}$ $[0.096,\,0.114]$ \\
\cmidrule(l{1.95cm}){1-7}
\phantom{\advoptname-R} LlamaGen &  & \greybox & $0.31$ $[0.23,\,0.41]$ & 1.2e-01 & ${31.649}_{\pm0.790}$ $[31.492,\,31.805]$ & ${0.131}_{\pm0.056}$ $[0.120,\,0.142]$ \\
\phantom{\advoptname-R} LlamaGen &  & \whitebox & $0.00$ $[0.00,\,0.04]$ & 9.9e-01 & ${33.038}_{\pm2.045}$ $[32.632,\,33.444]$ & ${0.072}_{\pm0.040}$ $[0.064,\,0.080]$ \\
\midrule
\multicolumn{7}{c}{Forgery} \\
\midrule
\advoptname-F (aggregated) &  & \blackbox& $0.12$ $[0.09,\,0.16]$ & 1.0e-01 & ${34.754}_{\pm3.467}$ $[34.360,\,35.148]$ & ${0.061}_{\pm0.040}$ $[0.056,\,0.065]$ \\
\phantom{\advoptname-F} RAR-XL &  & \blackbox & $0.11$ $[0.06,\,0.19]$ & 9.2e-02 & ${34.517}_{\pm3.399}$ $[33.843,\,35.192]$ & ${0.068}_{\pm0.044}$ $[0.059,\,0.076]$ \\
\phantom{\advoptname-F} Anole &  & \blackbox & $0.11$ $[0.06,\,0.19]$ & 1.0e-01 & ${35.169}_{\pm3.751}$ $[34.424,\,35.913]$ & ${0.058}_{\pm0.039}$ $[0.050,\,0.065]$ \\
\phantom{\advoptname-F} Taming &  & \blackbox & $0.13$ $[0.08,\,0.21]$ & 1.0e-01 & ${34.576}_{\pm3.226}$ $[33.936,\,35.216]$ & ${0.058}_{\pm0.036}$ $[0.050,\,0.065]$ \\
\cmidrule(l{1.95cm}){1-7}
\phantom{\advoptname-F} LlamaGen &  & \greybox & $0.09$ $[0.05,\,0.16]$ & 1.6e-01 & ${34.941}_{\pm3.792}$ $[34.189,\,35.693]$ & ${0.050}_{\pm0.033}$ $[0.043,\,0.056]$ \\
\phantom{\advoptname-F} LlamaGen &  & \whitebox & $0.97$ $[0.92,\,0.99]$ & 8.4e-15 & ${32.844}_{\pm0.920}$ $[32.662,\,33.027]$ & ${0.051}_{\pm0.021}$ $[0.047,\,0.055]$ \\
\bottomrule
\end{tabular}

%% file: tables/supp/full_results/arwm_GPT-L.tex
\begin{tabular}{lll @{\hspace{1.2em}} r @{\hspace{0.7em}} S[table-format=1.1e4, exponent-product=\times,tight-spacing=true] r @{\hspace{0.7em}} r}
\toprule
 &  &  & \multicolumn{1}{l}{TPR} & \multicolumn{1}{c}{P-Value} & \multicolumn{1}{c}{PSNR$\uparrow$} & \multicolumn{1}{c}{LPIPS$\downarrow$} \\
\midrule
Watermarked & &  & $1.00$ $[0.99,\,1.00]$ & 3.0e-135 & - & - \\
\midrule
\multicolumn{7}{c}{Removal} \\
\midrule
Geometric Transf. & & & $0.58$ $[0.56,\,0.60]$ & 3.3e-03 & ${16.465}_{\pm3.335}$ $[16.362,\,16.569]$ & ${0.196}_{\pm0.080}$ $[0.193,\,0.198]$ \\
Perturbations & & & $0.84$ $[0.83,\,0.84]$ & 3.4e-38 & ${17.696}_{\pm8.805}$ $[17.490,\,17.902]$ & ${0.457}_{\pm0.415}$ $[0.447,\,0.467]$ \\
\midrule
Regen. & & \blackbox & $0.97$ $[0.96,\,0.98]$ & 1.9e-10 & ${25.340}_{\pm3.351}$ $[25.132,\,25.548]$ & ${0.097}_{\pm0.036}$ $[0.094,\,0.099]$ \\
Rinse & & \blackbox & $0.48$ $[0.45,\,0.51]$ & 9.4e-03 & ${22.364}_{\pm2.689}$ $[22.197,\,22.530]$ & ${0.214}_{\pm0.064}$ $[0.210,\,0.218]$ \\
CtrlRegen{}$+$ & & \blackbox & $0.80$ $[0.77,\,0.82]$ & 3.1e-05 & ${24.196}_{\pm2.650}$ $[24.031,\,24.360]$ & ${0.132}_{\pm0.046}$ $[0.130,\,0.135]$ \\
\midrule
VQ-Regen\hspace{0.3cm} (aggregated) &  & \blackbox & $0.43$ $[0.42,\,0.45]$ & 1.5e-02 & ${21.510}_{\pm3.501}$ $[21.385,\,21.636]$ & ${0.149}_{\pm0.049}$ $[0.147,\,0.151]$ \\
\phantom{VQ-Regen\hspace{0.3cm}} RAR-XL &  & \blackbox & $0.17$ $[0.14,\,0.19]$ & 1.1e-01 & ${19.751}_{\pm2.829}$ $[19.575,\,19.926]$ & ${0.172}_{\pm0.049}$ $[0.169,\,0.175]$ \\
\phantom{VQ-Regen\hspace{0.3cm}} Anole &  & \blackbox & $0.63$ $[0.60,\,0.66]$ & 2.5e-03 & ${22.626}_{\pm3.574}$ $[22.404,\,22.848]$ & ${0.135}_{\pm0.045}$ $[0.132,\,0.138]$ \\
\phantom{VQ-Regen\hspace{0.3cm}} Taming &  & \blackbox & $0.51$ $[0.48,\,0.54]$ & 7.3e-03 & ${22.154}_{\pm3.356}$ $[21.946,\,22.362]$ & ${0.140}_{\pm0.044}$ $[0.137,\,0.143]$ \\
\cmidrule(l{1.95cm}){1-7}
\phantom{VQ-Regen\hspace{0.3cm}} LlamaGen &  & \greybox & $0.99$ $[0.98,\,0.99]$ & 1.8e-09 & ${23.360}_{\pm2.868}$ $[23.183,\,23.538]$ & ${0.098}_{\pm0.032}$ $[0.096,\,0.100]$ \\
\phantom{VQ-Regen\hspace{0.3cm}} LlamaGen &  & \whitebox & $0.99$ $[0.98,\,0.99]$ & 1.9e-10 & ${19.631}_{\pm1.744}$ $[19.523,\,19.740]$ & ${0.181}_{\pm0.035}$ $[0.179,\,0.183]$ \\
\midrule
\advoptname-R (aggregated) &  & \blackbox & $0.87$ $[0.83,\,0.90]$ & 1.1e-10 & ${31.882}_{\pm1.123}$ $[31.754,\,32.010]$ & ${0.144}_{\pm0.062}$ $[0.137,\,0.151]$ \\
\phantom{\advoptname-R} RAR-XL &  & \blackbox & $0.95$ $[0.89,\,0.98]$ & 1.3e-17 & ${32.364}_{\pm1.468}$ $[32.072,\,32.655]$ & ${0.136}_{\pm0.047}$ $[0.127,\,0.146]$ \\
\phantom{\advoptname-R} Anole &  & \blackbox & $0.82$ $[0.73,\,0.88]$ & 2.4e-09 & ${31.551}_{\pm0.871}$ $[31.378,\,31.724]$ & ${0.147}_{\pm0.067}$ $[0.133,\,0.160]$ \\
\phantom{\advoptname-R} Taming &  & \blackbox & $0.84$ $[0.76,\,0.90]$ & 4.6e-08 & ${31.731}_{\pm0.726}$ $[31.587,\,31.875]$ & ${0.149}_{\pm0.070}$ $[0.135,\,0.163]$ \\
\cmidrule(l{1.95cm}){1-7}
\phantom{\advoptname-R} LlamaGen &  & \greybox & $0.45$ $[0.36,\,0.55]$ & 2.2e-02 & ${31.810}_{\pm0.811}$ $[31.650,\,31.971]$ & ${0.178}_{\pm0.080}$ $[0.162,\,0.194]$ \\
\phantom{\advoptname-R} LlamaGen &  & \whitebox & $0.00$ $[0.00,\,0.04]$ & 1.0e+00 & ${33.771}_{\pm2.582}$ $[33.259,\,34.283]$ & ${0.092}_{\pm0.060}$ $[0.080,\,0.104]$ \\
\midrule
\multicolumn{7}{c}{Forgery} \\
\midrule
\advoptname-F (aggregated) &  & \blackbox& $0.17$ $[0.13,\,0.21]$ & 7.9e-02 & ${34.685}_{\pm3.416}$ $[34.297,\,35.073]$ & ${0.073}_{\pm0.045}$ $[0.068,\,0.078]$ \\
\phantom{\advoptname-F} RAR-XL &  & \blackbox & $0.14$ $[0.09,\,0.22]$ & 6.6e-02 & ${34.427}_{\pm3.446}$ $[33.743,\,35.111]$ & ${0.078}_{\pm0.043}$ $[0.070,\,0.087]$ \\
\phantom{\advoptname-F} Anole &  & \blackbox & $0.17$ $[0.11,\,0.26]$ & 9.4e-02 & ${35.109}_{\pm3.669}$ $[34.381,\,35.837]$ & ${0.070}_{\pm0.047}$ $[0.061,\,0.079]$ \\
\phantom{\advoptname-F} Taming &  & \blackbox & $0.19$ $[0.13,\,0.28]$ & 7.2e-02 & ${34.519}_{\pm3.102}$ $[33.903,\,35.134]$ & ${0.071}_{\pm0.046}$ $[0.062,\,0.080]$ \\
\cmidrule(l{1.95cm}){1-7}
\phantom{\advoptname-F} LlamaGen &  & \greybox & $0.18$ $[0.12,\,0.27]$ & 7.2e-02 & ${34.408}_{\pm3.286}$ $[33.756,\,35.060]$ & ${0.066}_{\pm0.035}$ $[0.059,\,0.073]$ \\
\phantom{\advoptname-F} LlamaGen &  & \whitebox & $1.00$ $[0.96,\,1.00]$ & 5.4e-46 & ${33.010}_{\pm1.018}$ $[32.808,\,33.212]$ & ${0.060}_{\pm0.023}$ $[0.055,\,0.064]$ \\
\bottomrule
\end{tabular}

%% file: tables/supp/full_results/arwm_rar_xl.tex
\begin{tabular}{lll @{\hspace{1.2em}} r @{\hspace{0.7em}} S[table-format=1.1e4, exponent-product=\times,tight-spacing=true] r @{\hspace{0.7em}} r}
\toprule
 &  &  & \multicolumn{1}{l}{TPR} & \multicolumn{1}{c}{P-Value} & \multicolumn{1}{c}{PSNR$\uparrow$} & \multicolumn{1}{c}{LPIPS$\downarrow$} \\
\midrule
Watermarked & &  & $1.00$ $[1.00,\,1.00]$ & 2.5e-82 & - & - \\
\midrule
\multicolumn{7}{c}{Removal} \\
\midrule
Geometric Transf. & & & $0.70$ $[0.69,\,0.72]$ & 2.7e-04 & ${15.882}_{\pm3.177}$ $[15.783,\,15.980]$ & ${0.164}_{\pm0.064}$ $[0.162,\,0.166]$ \\
Perturbations & & & $0.77$ $[0.76,\,0.78]$ & 3.0e-15 & ${17.516}_{\pm8.409}$ $[17.319,\,17.713]$ & ${0.425}_{\pm0.381}$ $[0.416,\,0.434]$ \\
\midrule
Regen. & & \blackbox & $0.76$ $[0.73,\,0.79]$ & 2.7e-04 & ${22.151}_{\pm2.644}$ $[21.987,\,22.315]$ & ${0.144}_{\pm0.039}$ $[0.141,\,0.146]$ \\
Rinse & & \blackbox & $0.14$ $[0.12,\,0.16]$ & 1.6e-01 & ${19.006}_{\pm2.026}$ $[18.880,\,19.132]$ & ${0.368}_{\pm0.073}$ $[0.364,\,0.373]$ \\
CtrlRegen{}$+$ & & \blackbox & $0.86$ $[0.84,\,0.88]$ & 5.4e-06 & ${23.929}_{\pm2.217}$ $[23.792,\,24.067]$ & ${0.130}_{\pm0.040}$ $[0.127,\,0.132]$ \\
\midrule
VQ-Regen\hspace{0.3cm} (aggregated) &  & \blackbox & $0.53$ $[0.52,\,0.55]$ & 8.9e-03 & ${20.103}_{\pm2.821}$ $[20.002,\,20.204]$ & ${0.155}_{\pm0.045}$ $[0.153,\,0.156]$ \\
\phantom{VQ-Regen\hspace{0.3cm}} LlamaGen &  & \blackbox & $0.88$ $[0.86,\,0.90]$ & 9.5e-05 & ${20.487}_{\pm2.631}$ $[20.324,\,20.651]$ & ${0.133}_{\pm0.037}$ $[0.131,\,0.136]$ \\
\phantom{VQ-Regen\hspace{0.3cm}} Anole &  & \blackbox & $0.38$ $[0.35,\,0.41]$ & 2.5e-02 & ${20.082}_{\pm2.959}$ $[19.898,\,20.265]$ & ${0.161}_{\pm0.046}$ $[0.158,\,0.164]$ \\
\phantom{VQ-Regen\hspace{0.3cm}} Taming &  & \blackbox & $0.34$ $[0.31,\,0.37]$ & 3.5e-02 & ${19.741}_{\pm2.816}$ $[19.566,\,19.916]$ & ${0.169}_{\pm0.044}$ $[0.167,\,0.172]$ \\
\cmidrule(l{1.95cm}){1-7}
\phantom{VQ-Regen\hspace{0.3cm}} RAR-XL &  & \greybox & $0.47$ $[0.44,\,0.50]$ & 1.3e-02 & ${18.189}_{\pm2.398}$ $[18.041,\,18.338]$ & ${0.168}_{\pm0.040}$ $[0.166,\,0.171]$ \\
\phantom{VQ-Regen\hspace{0.3cm}} RAR-XL &  & \whitebox & $0.53$ $[0.50,\,0.56]$ & 8.9e-03 & ${14.231}_{\pm1.932}$ $[14.111,\,14.351]$ & ${0.302}_{\pm0.068}$ $[0.298,\,0.306]$ \\
\midrule
\advoptname-R (aggregated) &  & \blackbox & $0.72$ $[0.67,\,0.77]$ & 7.6e-06 & ${32.086}_{\pm3.055}$ $[31.739,\,32.433]$ & ${0.107}_{\pm0.051}$ $[0.101,\,0.113]$ \\
\phantom{\advoptname-R} LlamaGen &  & \blackbox & $0.61$ $[0.51,\,0.70]$ & 5.7e-04 & ${31.737}_{\pm0.800}$ $[31.578,\,31.896]$ & ${0.119}_{\pm0.051}$ $[0.109,\,0.130]$ \\
\phantom{\advoptname-R} Anole &  & \blackbox & $0.75$ $[0.66,\,0.82]$ & 2.0e-07 & ${32.907}_{\pm5.091}$ $[31.897,\,33.917]$ & ${0.099}_{\pm0.054}$ $[0.088,\,0.110]$ \\
\phantom{\advoptname-R} Taming &  & \blackbox & $0.80$ $[0.71,\,0.87]$ & 3.0e-07 & ${31.613}_{\pm0.772}$ $[31.460,\,31.767]$ & ${0.104}_{\pm0.048}$ $[0.094,\,0.113]$ \\
\cmidrule(l{1.95cm}){1-7}
\phantom{\advoptname-R} RAR-XL &  & \greybox & $0.50$ $[0.40,\,0.60]$ & 1.1e-02 & ${32.013}_{\pm1.068}$ $[31.801,\,32.224]$ & ${0.104}_{\pm0.033}$ $[0.097,\,0.110]$ \\
\phantom{\advoptname-R} RAR-XL &  & \whitebox & $0.00$ $[0.00,\,0.04]$ & 8.7e-01 & ${31.570}_{\pm1.581}$ $[31.256,\,31.884]$ & ${0.132}_{\pm0.051}$ $[0.122,\,0.142]$ \\
\midrule
\multicolumn{7}{c}{Forgery} \\
\midrule
\advoptname-F (aggregated) &  & \blackbox& $0.05$ $[0.03,\,0.08]$ & 1.6e-01 & ${34.796}_{\pm3.453}$ $[34.404,\,35.189]$ & ${0.058}_{\pm0.038}$ $[0.054,\,0.062]$ \\
\phantom{\advoptname-F} LlamaGen &  & \blackbox & $0.05$ $[0.02,\,0.11]$ & 1.3e-01 & ${34.239}_{\pm3.011}$ $[33.642,\,34.836]$ & ${0.057}_{\pm0.032}$ $[0.051,\,0.063]$ \\
\phantom{\advoptname-F} Anole &  & \blackbox & $0.05$ $[0.02,\,0.11]$ & 1.6e-01 & ${35.500}_{\pm3.724}$ $[34.761,\,36.238]$ & ${0.058}_{\pm0.044}$ $[0.049,\,0.067]$ \\
\phantom{\advoptname-F} Taming &  & \blackbox & $0.05$ $[0.02,\,0.11]$ & 1.6e-01 & ${34.650}_{\pm3.499}$ $[33.956,\,35.344]$ & ${0.059}_{\pm0.037}$ $[0.052,\,0.067]$ \\
\cmidrule(l{1.95cm}){1-7}
\phantom{\advoptname-F} RAR-XL &  & \greybox & $0.08$ $[0.04,\,0.15]$ & 1.6e-01 & ${34.610}_{\pm3.564}$ $[33.903,\,35.318]$ & ${0.064}_{\pm0.039}$ $[0.056,\,0.072]$ \\
\phantom{\advoptname-F} RAR-XL &  & \whitebox & $0.40$ $[0.31,\,0.50]$ & 3.5e-02 & ${33.156}_{\pm2.847}$ $[32.591,\,33.721]$ & ${0.079}_{\pm0.039}$ $[0.071,\,0.087]$ \\
\bottomrule
\end{tabular}

%% file: tables/supp/full_results/bitmark_1M.tex
\begin{tabular}{lll @{\hspace{1.2em}} r @{\hspace{0.7em}} S[table-format=1.1e4, exponent-product=\times,tight-spacing=true] r @{\hspace{0.7em}} r}
\toprule
 &  &  & \multicolumn{1}{l}{TPR} & \multicolumn{1}{c}{P-Value} & \multicolumn{1}{c}{PSNR$\uparrow$} & \multicolumn{1}{c}{LPIPS$\downarrow$} \\
\midrule
Watermarked & & & $1.00$ $[1.00,\,1.00]$ & 0.0e+00 & - & - \\
Radioactive\hspace{0.2cm} $\infty$-2B, 10\% & & & $0.74$ $[0.71,\,0.77]$ & 2.4e-04 & - & - \\
\phantom{Radioactive\hspace{0.2cm}} $\infty$-2B, 25\% & & & $0.99$ $[0.98,\,1.00]$ & 1.1e-11 & - & - \\
\phantom{Radioactive\hspace{0.2cm}} $\infty$-2B, 50\% & & & $1.00$ $[1.00,\,1.00]$ & 2.3e-55 & - & - \\
\phantom{Radioactive\hspace{0.2cm}} $\infty$-2B, 100\% & & & $1.00$ $[1.00,\,1.00]$ & 8.4e-212 & - & - \\
\phantom{Radioactive\hspace{0.2cm}} SD2.1, 50\% & & & $0.81$ $[0.79,\,0.84]$ & 2.9e-05 & - & - \\
\phantom{Radioactive\hspace{0.2cm}} SD2.1, 100\% & & & $0.98$ $[0.97,\,0.99]$ & 6.5e-14 & - & - \\
\midrule
\multicolumn{7}{c}{Removal} \\
\midrule
Geometric Transf. & & & $1.00$ $[1.00,\,1.00]$ & 4.2e-56 & ${14.484}_{\pm2.709}$ $[14.400,\,14.568]$ & ${0.316}_{\pm0.115}$ $[0.313,\,0.320]$ \\
Perturbations & & & $0.98$ $[0.98,\,0.98]$ & 6.0e-250 & ${18.130}_{\pm8.861}$ $[17.922,\,18.337]$ & ${0.462}_{\pm0.442}$ $[0.451,\,0.472]$ \\
\midrule
Regen. & & \blackbox & $1.00$ $[1.00,\,1.00]$ & 4.5e-235 & ${27.956}_{\pm2.982}$ $[27.771,\,28.141]$ & ${0.064}_{\pm0.026}$ $[0.062,\,0.065]$ \\
Rinse & & \blackbox & $1.00$ $[1.00,\,1.00]$ & 1.7e-70 & ${25.102}_{\pm2.474}$ $[24.948,\,25.255]$ & ${0.140}_{\pm0.043}$ $[0.137,\,0.143]$ \\
CtrlRegen{}$+$ & & \blackbox & $0.99$ $[0.98,\,1.00]$ & 2.3e-22 & ${23.451}_{\pm2.353}$ $[23.305,\,23.597]$ & ${0.218}_{\pm0.077}$ $[0.213,\,0.222]$ \\
\midrule
VQ-Regen\hspace{0.3cm} (aggregated) & & \blackbox & $1.00$ $[1.00,\,1.00]$ & 3.3e-83 & ${24.208}_{\pm3.084}$ $[24.112,\,24.303]$ & ${0.102}_{\pm0.042}$ $[0.100,\,0.103]$ \\
\phantom{VQ-Regen\hspace{0.3cm}} LlamaGen & & \blackbox & $1.00$ $[1.00,\,1.00]$ & 6.5e-139 & ${25.179}_{\pm2.522}$ $[25.022,\,25.335]$ & ${0.080}_{\pm0.031}$ $[0.078,\,0.082]$ \\
\phantom{VQ-Regen\hspace{0.3cm}} RAR-XL & & \blackbox & $1.00$ $[0.99,\,1.00]$ & 4.1e-28 & ${21.911}_{\pm2.487}$ $[21.757,\,22.066]$ & ${0.131}_{\pm0.044}$ $[0.129,\,0.134]$ \\
\phantom{VQ-Regen\hspace{0.3cm}} Anole & & \blackbox & $1.00$ $[1.00,\,1.00]$ & 3.6e-93 & ${25.261}_{\pm3.281}$ $[25.058,\,25.465]$ & ${0.092}_{\pm0.036}$ $[0.090,\,0.094]$ \\
\phantom{VQ-Regen\hspace{0.3cm}} Taming & & \blackbox & $1.00$ $[1.00,\,1.00]$ & 3.0e-79 & ${24.480}_{\pm2.712}$ $[24.312,\,24.648]$ & ${0.103}_{\pm0.039}$ $[0.101,\,0.105]$ \\
\midrule
\advoptname-R (aggregated) & & \blackbox & $0.97$ $[0.95,\,0.98]$ & 5.2e-100 & ${34.315}_{\pm4.238}$ $[33.899,\,34.732]$ & ${0.223}_{\pm0.106}$ $[0.213,\,0.234]$ \\
\phantom{\advoptname-R} LlamaGen & & \blackbox & $0.88$ $[0.80,\,0.93]$ & 2.4e-32 & ${32.338}_{\pm1.672}$ $[32.006,\,32.670]$ & ${0.278}_{\pm0.074}$ $[0.263,\,0.293]$ \\
\phantom{\advoptname-R} RAR-XL & & \blackbox & $1.00$ $[0.96,\,1.00]$ & 2.4e-180 & ${36.475}_{\pm4.723}$ $[35.538,\,37.413]$ & ${0.158}_{\pm0.094}$ $[0.140,\,0.177]$ \\
\phantom{\advoptname-R} Anole & & \blackbox & $1.00$ $[0.96,\,1.00]$ & 3.4e-112 & ${34.503}_{\pm4.534}$ $[33.603,\,35.402]$ & ${0.223}_{\pm0.109}$ $[0.201,\,0.244]$ \\
\phantom{\advoptname-R} Taming & & \blackbox & $1.00$ $[0.96,\,1.00]$ & 8.5e-109 & ${33.945}_{\pm4.231}$ $[33.105,\,34.784]$ & ${0.234}_{\pm0.106}$ $[0.213,\,0.255]$ \\
\midrule
\advoptname-R (aggregated) & & \greybox & $0.93$ $[0.90,\,0.95]$ & 2.0e-47 & ${32.446}_{\pm3.512}$ $[32.047,\,32.845]$ & ${0.261}_{\pm0.087}$ $[0.251,\,0.271]$ \\
\phantom{\advoptname-R} $\infty$-2B ($d=16$) & & \greybox & $0.86$ $[0.78,\,0.91]$ & 6.2e-26 & ${31.230}_{\pm1.334}$ $[30.965,\,31.494]$ & ${0.290}_{\pm0.064}$ $[0.277,\,0.303]$ \\
\phantom{\advoptname-R} $\infty$-2B ($d=24$) & & \greybox & $0.93$ $[0.86,\,0.97]$ & 2.8e-28 & ${31.289}_{\pm1.289}$ $[31.033,\,31.545]$ & ${0.282}_{\pm0.062}$ $[0.270,\,0.294]$ \\
\phantom{\advoptname-R} $\infty$-2B ($d=64$) & & \greybox & $1.00$ $[0.96,\,1.00]$ & 2.0e-135 & ${34.819}_{\pm5.028}$ $[33.821,\,35.816]$ & ${0.210}_{\pm0.104}$ $[0.190,\,0.231]$ \\
\midrule
\advoptname-R $\infty$-2B ($d=32$) & & \whitebox & $0.93$ $[0.86,\,0.97]$ & 7.6e-18 & ${31.665}_{\pm1.399}$ $[31.387,\,31.943]$ & ${0.264}_{\pm0.054}$ $[0.253,\,0.275]$ \\
\midrule
Next-Frame P. & & \blackbox & $1.00$ $[0.99,\,1.00]$ & 6.9e-226 & ${11.305}_{\pm1.832}$ $[11.192,\,11.419]$ & ${0.530}_{\pm0.085}$ $[0.524,\,0.535]$ \\
\midrule
BitFlipper & & \whitebox{}$+$ & $0.68$ $[0.65,\,0.71]$ & 2.3e-08 & ${19.222}_{\pm2.230}$ $[19.084,\,19.361]$ & ${0.236}_{\pm0.082}$ $[0.231,\,0.241]$ \\
\bitadvoptname $\infty$-2B ($d=32$) & & \whitebox{}$+$ & $0.00$ $[0.00,\,0.04]$ & 3.0e-01 & ${46.172}_{\pm2.898}$ $[45.597,\,46.747]$ & ${0.009}_{\pm0.008}$ $[0.007,\,0.010]$ \\
\midrule
\multicolumn{7}{c}{Forgery} \\
\midrule
\advoptname-F (aggregated) & & \blackbox & $0.59$ $[0.55,\,0.64]$ & 3.5e-03 & ${31.513}_{\pm0.294}$ $[31.484,\,31.542]$ & ${0.187}_{\pm0.075}$ $[0.180,\,0.194]$ \\
\phantom{\advoptname-F} LlamaGen & & \blackbox & $0.88$ $[0.80,\,0.93]$ & 3.7e-05 & ${31.591}_{\pm0.320}$ $[31.527,\,31.654]$ & ${0.177}_{\pm0.072}$ $[0.163,\,0.192]$ \\
\phantom{\advoptname-F} RAR-XL & & \blackbox & $0.62$ $[0.52,\,0.71]$ & 2.2e-03 & ${31.456}_{\pm0.371}$ $[31.382,\,31.529]$ & ${0.191}_{\pm0.072}$ $[0.176,\,0.205]$ \\
\phantom{\advoptname-F} Anole & & \blackbox & $0.35$ $[0.26,\,0.45]$ & 2.2e-02 & ${31.467}_{\pm0.220}$ $[31.424,\,31.511]$ & ${0.194}_{\pm0.080}$ $[0.178,\,0.210]$ \\
\phantom{\advoptname-F} Taming & & \blackbox & $0.53$ $[0.43,\,0.62]$ & 7.3e-03 & ${31.537}_{\pm0.220}$ $[31.493,\,31.581]$ & ${0.186}_{\pm0.077}$ $[0.171,\,0.202]$ \\
\midrule
\advoptname-F (aggregated) & & \greybox & $0.99$ $[0.98,\,1.00]$ & 1.7e-23 & ${32.512}_{\pm1.124}$ $[32.385,\,32.640]$ & ${0.149}_{\pm0.063}$ $[0.141,\,0.156]$ \\
\phantom{\advoptname-F} $\infty$-2B ($d=16$) & & \greybox & $0.99$ $[0.95,\,1.00]$ & 2.3e-15 & ${31.987}_{\pm0.598}$ $[31.868,\,32.105]$ & ${0.167}_{\pm0.064}$ $[0.154,\,0.180]$ \\
\phantom{\advoptname-F} $\infty$-2B ($d=24$) & & \greybox & $1.00$ $[0.96,\,1.00]$ & 3.6e-31 & ${31.837}_{\pm0.606}$ $[31.717,\,31.958]$ & ${0.168}_{\pm0.063}$ $[0.156,\,0.181]$ \\
\phantom{\advoptname-F} $\infty$-2B ($d=64$) & & \greybox & $0.99$ $[0.95,\,1.00]$ & 4.1e-37 & ${33.713}_{\pm0.946}$ $[33.525,\,33.901]$ & ${0.111}_{\pm0.043}$ $[0.102,\,0.119]$ \\
\midrule
\advoptname-F $\infty$-2B ($d=32$) & & \whitebox & $1.00$ $[0.96,\,1.00]$ & 3.4e-232 & ${31.956}_{\pm1.294}$ $[31.700,\,32.213]$ & ${0.160}_{\pm0.058}$ $[0.149,\,0.172]$ \\
\midrule
Reprompt-F & & \blackbox & $0.97$ $[0.96,\,0.98]$ & 1.9e-70 & ${14.090}_{\pm1.968}$ $[13.968,\,14.212]$ & ${0.507}_{\pm0.098}$ $[0.501,\,0.513]$ \\
\midrule
Freq. Inj. Forgery\hspace{0.0cm}~Setting A & & \blackbox & $0.72$ $[0.69,\,0.75]$ & 2.9e-06 & ${37.504}_{\pm0.961}$ $[37.444,\,37.563]$ & ${0.058}_{\pm0.039}$ $[0.055,\,0.060]$ \\
\phantom{Freq. Inj. Forgery\hspace{0.0cm}}~Setting B & & \blackbox & $0.79$ $[0.76,\,0.81]$ & 7.4e-09 & ${35.436}_{\pm0.995}$ $[35.375,\,35.498]$ & ${0.094}_{\pm0.058}$ $[0.091,\,0.098]$ \\
\phantom{Freq. Inj. Forgery\hspace{0.0cm}}~Setting C & & \blackbox & $0.91$ $[0.89,\,0.92]$ & 3.0e-23 & ${29.755}_{\pm0.541}$ $[29.722,\,29.789]$ & ${0.178}_{\pm0.098}$ $[0.172,\,0.184]$ \\
\bottomrule
\end{tabular}

%% file: checklist.tex
\section*{NeurIPS Paper Checklist}

\begin{enumerate}

\item {\bf Claims}
    \item[] Question: Do the main claims made in the abstract and introduction accurately reflect the paper's contributions and scope?
    \item[] Answer: \answerYes{} %
    \item[] Justification: The claims in the abstract and Section~\ref{sec:introduction} are substantiated with the experiments and the results presented in Section~\ref{sec:evaluation}, with some limitations listed in \cref{sec:limitations}. 

\item {\bf Limitations}
    \item[] Question: Does the paper discuss the limitations of the work performed by the authors?
    \item[] Answer: \answerYes{} %
    \item[] Justification: Limitations are discussed in \cref{sec:limitations}.

\item {\bf Theory assumptions and proofs}
    \item[] Question: For each theoretical result, does the paper provide the full set of assumptions and a complete (and correct) proof?
    \item[] Answer: \answerNA{} %
    \item[] Justification: The paper does not prove any new theoretical results.

    \item {\bf Experimental result reproducibility}
    \item[] Question: Does the paper fully disclose all the information needed to reproduce the main experimental results of the paper to the extent that it affects the main claims and/or conclusions of the paper (regardless of whether the code and data are provided or not)?
    \item[] Answer: \answerYes{} %
    \item[] Justification: All details are given in Section~\ref{sec:evaluation} and Appendix~\ref{sec:supp:experimental_settings}.

\item {\bf Open access to data and code}
    \item[] Question: Does the paper provide open access to the data and code, with sufficient instructions to faithfully reproduce the main experimental results, as described in supplemental material?
    \item[] Answer: \answerYes{} %
    \item[] Justification: We provide code to launch our proposed attacks. All implementation details are provided in Appendix~\ref{sec:supp:experimental_settings}. The image data covering all the experiments amounts to several hundred gigabytes of data and cannot be feasibly shared.

\item {\bf Experimental setting/details}
    \item[] Question: Does the paper specify all the training and test details (e.g., data splits, hyperparameters, how they were chosen, type of optimizer) necessary to understand the results?
    \item[] Answer: \answerYes{} %
    \item[] Justification: Please refer to Setion~\ref{sec:evaluation} for high level settings, as well as Appendix~\ref{sec:supp:experimental_settings} for a more detailed breakdown of all settings.

\item {\bf Experiment statistical significance}
    \item[] Question: Does the paper report error bars suitably and correctly defined or other appropriate information about the statistical significance of the experiments?
    \item[] Answer: \answerYes{} %
    \item[] Justification: Tables~\ref{tab:supp:full_results:IndexMark:GPT-B} to~\ref{tab:supp:full_results:BitMark:1M} show a full breakdown of the main results for each attacker and target model combination. Quality metrics (PSNR, LPIPS) are reported as means with standard deviations. Detection accuracy metrics are reported as both threshold-related mean TPR@FPR=1\%, as well as threshold-agnostic median p-values. We also report confidence intervals (95\% Wilson score CIs for the binary TPR measure and student-t CIs for PSNR and LPIPS).
    For Bitmark, mean z-scores and their standard deviations are reported as well (Fig.~\ref{fig:main_boxplot}). We used 1000 attack instances for each experiment to ensure that the results are stable and not sensitive to random seeds. The exception is \advoptname-Removal and \advoptname-Forgery, as well as \bitadvoptname-Removal. Here, we used 100 attack instances due to a large optimization overhead.

\item {\bf Experiments compute resources}
    \item[] Question: For each experiment, does the paper provide sufficient information on the computer resources (type of compute workers, memory, time of execution) needed to reproduce the experiments?
    \item[] Answer: \answerYes{} %
    \item[] Justification: We provide a breakdown of runtimes for the experiments, as well as the compute resources used and the time span over which experiments were conducted in Appendix~\ref{sec:supp:compute}.
    
\item {\bf Code of ethics}
    \item[] Question: Does the research conducted in the paper conform, in every respect, with the NeurIPS Code of Ethics \url{https://neurips.cc/public/EthicsGuidelines}?
    \item[] Answer: \answerYes{} %
    \item[] Justification: We have reviewed the NeurIPS Code of Ethics. This study does not involve
human subjects or inappropriate use of data. We believe there will be no harmful consequences.

\item {\bf Broader impacts}
    \item[] Question: Does the paper discuss both potential positive societal impacts and negative societal impacts of the work performed?
    \item[] Answer: \answerYes{} %
    \item[] Justification: Please refer to Section~\ref{sec:introduction} and~\ref{sec:conclusion}.
    
\item {\bf Safeguards}
    \item[] Question: Does the paper describe safeguards that have been put in place for responsible release of data or models that have a high risk for misuse (e.g., pre-trained language models, image generators, or scraped datasets)?
    \item[] Answer: \answerNA{} %
    \item[] Justification: The paper does not pose such risk.

\item {\bf Licenses for existing assets}
    \item[] Question: Are the creators or original owners of assets (e.g., code, data, models), used in the paper, properly credited and are the license and terms of use explicitly mentioned and properly respected?
    \item[] Answer: \answerYes{} %
    \item[] Justification: All datasets used (COCO, ImageNet) are properly cited in the main text. COCO annotations are released under CC-BY 4.0, while the images fall under various Flickr Creative Commons licenses. ImageNet is used under its non-commercial research access terms. The URLs for third-party code repositories are provided in the Appendix with the correct commit ID used. Each repository contains a corresponding license. For the third-party code included in our repository, we have retained the original license files (e.g., MIT License) and explicitly pointed to these licenses and their terms of use within our provided README.md. An image in Figure~\ref{fig:frontpage} is attributed to the originator and linked to the license in place.

\item {\bf New assets}
    \item[] Question: Are new assets introduced in the paper well documented and is the documentation provided alongside the assets?
    \item[] Answer: \answerYes{} %
    \item[] Justification: We link to a codebase containing the experiment code to run our attacks, with documentation provided alongside it.

\item {\bf Crowdsourcing and research with human subjects}
    \item[] Question: For crowdsourcing experiments and research with human subjects, does the paper include the full text of instructions given to participants and screenshots, if applicable, as well as details about compensation (if any)? 
    \item[] Answer: \answerNA{} %
    \item[] Justification: The paper does not involve crowdsourcing or research with human subjects.

\item {\bf Institutional review board (IRB) approvals or equivalent for research with human subjects}
    \item[] Question: Does the paper describe potential risks incurred by study participants, whether such risks were disclosed to the subjects, and whether Institutional Review Board (IRB) approvals (or an equivalent approval/review based on the requirements of your country or institution) were obtained?
    \item[] Answer: \answerNA{} %
    \item[] Justification: The paper does not involve crowdsourcing or research with human subjects.

\item {\bf Declaration of LLM usage}
    \item[] Question: Does the paper describe the usage of LLMs if it is an important, original, or non-standard component of the core methods in this research? Note that if the LLM is used only for writing, editing, or formatting purposes and does \emph{not} impact the core methodology, scientific rigor, or originality of the research, declaration is not required.
    \item[] Answer: \answerNA{} %
    \item[] Justification: The core method development in this research does not involve LLMs as any
important, original, or non-standard components.

\end{enumerate}